\documentclass[letterpaper, 10 pt, conference]{ieeeconf}  

\IEEEoverridecommandlockouts                              

\usepackage{graphicx} 
\usepackage{epsfig} 
\usepackage{mathptmx} 
\usepackage{times} 
\usepackage{amsmath} 
\usepackage{amssymb}  
\usepackage{multirow}
\usepackage{makecell}
\usepackage{xcolor}
\usepackage{float}

\usepackage{comment}

\usepackage{algorithm}
\usepackage{algpseudocode}
\usepackage{subfigure} 
\usepackage{hyperref}

\newcommand{\equationref}[1]{\hyperref[#1]{Eq.~\ref*{#1}}}
\newcommand{\figref}[1]{\hyperref[#1]{Fig.~\ref*{#1}}}
\newcommand{\tabref}[1]{\hyperref[#1]{Table~\ref*{#1}}}
\newcommand{\secref}[1]{\hyperref[#1]{Section~\ref*{#1}}}
\newcommand{\algoref}[1]{\hyperref[#1]{Algorithm~\ref*{#1}}}

\usepackage{tikz}

\usepackage{xcolor}
\usepackage{hyperref}
\hypersetup{
    colorlinks=true,
    urlcolor=blue
}

\usepackage[capitalize]{cleveref}
\crefname{section}{Section}{Sections}
\crefname{table}{Table}{Tables}
\crefname{figure}{Figure}{Figures}

\usepackage{subcaption}
 \usepackage{booktabs}
\usepackage{subfigure}

\title{\LARGE \bf

Domain Randomization for Object Detection in Manufacturing Applications using Synthetic Data: A Comprehensive Study

}

\author{
Xiaomeng Zhu$^{1,3}$, Jacob Henningsson$^{1}$, Duruo Li$^{1}$, Pär Mårtensson$^{1}$,\\ Lars Hanson$^{2}$, Mårten Björkman$^{3}$, Atsuto Maki$^{3}$%
\thanks{$^{1}$Scania CV AB, Södertälje, Sweden. {\href{mailto:xiaomeng.zhu@scania.com}{xiaomeng.zhu@scania.com}}}%
\thanks{$^{2}$Skövde University, Skövde, Sweden.} 
\thanks{$^{3}$KTH Royal Institute of Technology, Stockholm, Sweden. {\{xiazhu, atsuto, celle\}@kth.se}}%
}

\begin{document}

\maketitle
\thispagestyle{empty}
\pagestyle{empty}

\begin{abstract}

This paper addresses key aspects of domain randomization in generating synthetic data for manufacturing object detection applications. 
To this end, we present a comprehensive data generation pipeline that reflects different factors: object characteristics, background, illumination, camera settings, and post-processing. 
We also introduce the Synthetic Industrial Parts Object Detection dataset (SIP15-OD) consisting of 15 objects from three industrial use cases under varying environments as a test bed for the study, while also employing an industrial dataset publicly available for robotic applications. In our experiments, we present more abundant results and insights into the feasibility as well as challenges of sim-to-real object detection. In particular,  we identified material properties, rendering methods, post-processing, and distractors as important factors. Our method, leveraging these, achieves top performance on the public dataset with Yolov8 models trained exclusively on synthetic data; mAP@50 scores of 96.4\% for the robotics dataset, and 94.1\%, 99.5\%, and 95.3\% across three of the SIP15-OD use cases, respectively. 
The results showcase the effectiveness of the proposed domain randomization, potentially covering the distribution close to real data for the applications.

\end{abstract}

\section{INTRODUCTION}

In advanced manufacturing, object detection plays a critical role in boosting operational efficiency and ensuring product quality. Tasks such as quality inspection and robotic picking rely on accurate object detection systems \cite{zhu2023towards, ahmad2022deep, vijayakumar2024yolo}. However, deploying deep learning models in manufacturing is challenging due to the limited availability of annotated real-world data, which is costly and time-consuming to gather.

An effective solution is to generate synthetic data from Computer-Aided Design (CAD) models, which are commonly used in manufacturing \cite{eversberg2021generating}. This method produces diverse datasets that replicate complex scenarios without the need for costly real-world data collection and annotation. Synthetic data reduces time, eliminates annotation errors, and enables the creation of rare or hard-to-capture events. However, the sim-to-real reality gap remains a major challenge. Models trained on synthetic data often struggle to generalize to real-world data due to differences in data distribution and environmental conditions \cite{zhu2024automated}.

One approach to bridging the reality gap is domain randomization (DR), which introduces random variations into synthetic data, training models to view real-world data as just another variant to enhance their generalization to real-world applications \cite{tobin2017domain}. Initially proposed by Tobin et al. \cite{tobin2017domain}, DR has been adopted in several manufacturing applications \cite{zhu2023towards, eversberg2021generating, cohen2020cad, tang2024two, sampaio2021novel,  horvath2022object, Mayershofer2021}, focusing on industrial products over everyday objects and outdoor environments \cite{Tremblay2018, prakash2019structured}. Although manufacturing environments are more controlled, they still vary in scenarios, backgrounds, occlusions, and lighting, and the objects themselves often feature unique appearances, such as metallic or textureless surfaces. However, most DR research in manufacturing is limited to single use cases with narrow object diversity and restricted environments, hindering generalizability. Therefore, we argue that current methods have yet to fully capture the complexity and variability of real-world manufacturing settings.

This study aims to identify key DR factors and develop a comprehensive synthetic data generation pipeline for manufacturing object detection applications. Previous literature shows that effective domain randomization (DR) typically applies to components such as object characteristics, background, illumination, camera settings, and post-processing. Building on this, we developed a pipeline that incorporates DR across all these dimensions and generates RGB images from CAD models. To evaluate its effectiveness, the pipeline was tested on two datasets considering multiple manufacturing environments and diverse industrial use cases.  The first dataset, from Horvath et al. \cite{horvath2022object}, is tailored for a robotic application featuring 10 industrial objects. The second is the Industrial Parts Object Detection Dataset (SIP15-OD) developed by us, which includes 15 objects from three distinct manufacturing use cases captured under varied environmental conditions.

To our knowledge, this is the first study to account for all DR components in synthetic data generation and evaluate them across multiple datasets for manufacturing object detection. This comprehensive study revealed several critical factors, such as object material properties and rendering methods, that are crucial for sim-to-real transfer in manufacturing object detection but have been underexplored. Leveraging these insights, we trained Yolov8 models only on generated synthetic data and achieved mAP@50 scores of 96.4\%, 94.1\%, 99.5\%, and 95.3\% on the robotics dataset and the three SIP15-OD use cases, respectively.


The contributions of this paper are as follows:
\begin{enumerate}
\item Along with factors like post-processing and distractors identified in previous research, we found additional key DR factors crucial for sim-to-real manufacturing object detection, including rendering methods and object material properties.

\item We presented the SIP15-OD dataset, comprising 321 images and 877 annotated objects across 15 industrial categories from three distinct manufacturing use cases. This dataset captures some complexity of real-world manufacturing settings and can serve as a useful benchmark for object detection. The dataset and code are publicly available at \href{https://github.com/jacobhenningsson95/SynMfg_Code}{\textcolor{blue}{https://github.com/jacobhenningsson95/SynMfg\_Code}}.

\item We developed a synthetic data generation pipeline that enhanced sim-to-real manufacturing object detection, with models trained solely on its synthetic data achieving top performance on a public industrial dataset. Unlike previous studies limited to single settings, we evaluated it across multiple manufacturing scenarios, offering richer experimental results and broader insights for DR research.


\end{enumerate}


\section{Related work}
\label{sec:literature}
\subsection{Domain adaptation and domain randomization}
Domain adaptation (DA) and domain randomization (DR) are two strategies to address the reality gap in machine learning. DA bridges the gap between two domains by aligning their distributions through techniques such as feature space mapping or adversarial training\cite{zhu2024automated,tobin2017domain}. Within the context of synthetic data generation, it is described as creating photo-realistic images mimicking real conditions, which often requires high-fidelity simulations and substantial computational resources \cite{tobin2017domain, horvath2022object}.

DR, on the other hand, introduces random variations into synthetic training data, enabling models to generalize across diverse scenarios, including unseen ones. By randomizing generation settings, this approach reduces dependence on high-fidelity simulations \cite{eversberg2021generating, tobin2017domain, horvath2022object}. Structured Domain Randomization (SDR) \cite{prakash2019structured} refines DR by placing objects in realistic contexts, such as cars on roads, to enable models to learn spatial relationships. While DR typically assumes no access to real data and employs uniform parameter sampling, Guided Domain Randomization (GDR) incorporates task performance or real-data feedback to guide the randomization process, mitigating the risk of unrealistic or infeasible scenarios resulting from excessively broad distributions \cite{Mayershofer2021}.


\subsection{Object detection}

Recent advances in deep learning have led to significant improvements in object detection models, with EfficientDet, Yolov8, and DETR standing out for their strong performance in terms of both accuracy and speed \cite{tan2020efficientdet, Jocher_Ultralytics_YOLO_2023, carion2020end}. Among these, Yolov8 \cite{Jocher_Ultralytics_YOLO_2023} stands out for its high performance and lightweight design, making it suitable for manufacturing applications. For comprehensive overviews of recent developments in object detection, see \cite{vo2022review}, \cite{zaidi2022survey}, and \cite{ahmad2022deep}.


\subsection{Domain randomization in manufacturing applications}

Different studies have explored DR for synthetic data generation in manufacturing object detection. Tobin et al. \cite{tobin2017domain} pioneered DR in robotic applications, using simple geometric objects like cylinders and squares simulated in MuJoCo \cite{todorov2012mujoco} with VGG-16 \cite{simonyan2014very} for object localization. They randomized factors like object position, distractors, colors, camera view, number of lights, and post-processing noise. They rendered synthetic images using rasterization via OpenGL \cite{shreiner2013opengl}, which efficiently converts 3D geometric shapes into 2D images by projecting them onto a view plane. However, their use of basic shapes and non-realistic textures with low-quality renderers limited the generalizability of their approach to more complex scenarios.

Eversberg and Lambrecht \cite{eversberg2021generating} generated synthetic images for turbine blade quality inspection in a shop floor environment using Faster-RCNN \cite{ren2015faster}. They randomized object poses, textures, backgrounds, camera settings, and lighting conditions in Blender \cite{blender}, leveraging physics-based rendering (PBR) with path tracing for highly realistic images. Path tracing simulates light interactions by tracing rays, capturing shadows, reflections, and global illumination \cite{glassner1989introduction}. Their study highlighted the advantages of combining DR with GDR for manufacturing applications, finding that DR was effective for non-object elements like backgrounds and distractors, while GDR with realistic lighting and PBR textures enhanced object-related features. However, the study focused on single-object scenarios, overlooking the complexities of multi-object detection and lacking gravity simulations and post-processing techniques.

Horvath et al. \cite{horvath2022object} applied DR to a robotic application involving 10 industrial parts, with their 3D models and real data publicly available. They used PyBullet \cite{coumans2016pybullet} for gravity simulation, rasterization rendering, and Yolov4 \cite{bochkovskiy2020yolov4} for detection. Their DR setup involved randomizing object pose, texture, background, camera pose, and post-processing. They highlighted the significance of object texture, pose, and post-processing for effective model performance. However, illumination randomization was not included in their study.

\label{sec:method}
\begin{figure*}[t]
\vspace{-0.01\textwidth}
  \centering
  \includegraphics[width=0.93\linewidth]{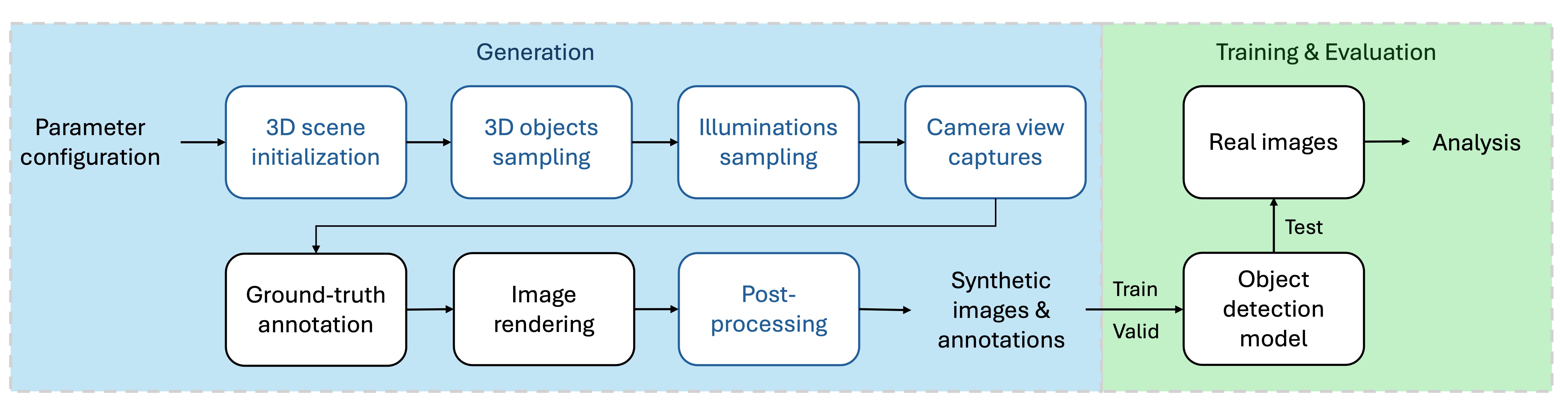}
  \caption{Overview of our data generation pipeline and evaluation method. Processes shown in blue text indicate where domain randomization is applied.}
  \label{fig:flowchart}
  \vspace{-0.03\textwidth}
\end{figure*}

In addition, Sampaio et al. \cite{sampaio2021novel} generated synthetic data for inspecting two motorbike parts, emphasizing the role of distractors and image resolution. Cohen et al. \cite{cohen2020cad} and Tang et al. \cite{tang2024two} focused on post-processing data augmentation rather than domain randomization. Cohen, targeting seven chair parts, highlighted the effect of blur, while Tang, detecting ten engine parts, emphasized object positioning and clutter. All studies conducted ablation experiments, showing that object detection performance improves with more synthetic data, but only up to a certain point, beyond which additional images offer no further accuracy gains.

\section{Method}

\subsection{Domain randomization}
Considering the key factors mentioned in previous work, DR techniques can be summarized in five components: object randomization (quantity, pose, texture), background and distractor randomization, camera randomization (various views), illumination randomization (environmental lights), and post-processing (noise, blur). Our synthetic data generation pipeline applies DR across all these areas, and to our knowledge, we are the first to fully integrate all of them. The key DR parameters in our pipeline are as follows. 

\textbf{3D object sampling} -- \textbf{\textit{Quantity:}} objects are randomly selected from a uniform distribution between zero and the total number of objects, with categories chosen evenly across available categories. \textbf{\textit{Pose:}} Objects are assigned random coordinates and rotation, ensuring no overlap.
\textbf{\textit{Gravity point:}} The lowest points of objects are set to zero to simulate gravity, allowing various orientations to better represent real-world setups where objects do not always lie flat.
\textbf{\textit{Texture:}} Objects are assigned one of three textures, 1) RGB Texture with random RGB values. 2) Image Texture with random images from the Flickr 8K dataset \cite{Hodosh2013Flickr}. 3) PBR Material Texture with predefined materials from the CG Texture dataset \cite{Demes2023Texture}, including properties such as metalness and roughness.

\textbf{Camera view setup} -- \textbf{\textit{Pose:}} Camera is positioned in spherical coordinates (radius \( r \), polar angle \( \theta \), and azimuthal angle \( \phi \)), with the radius determined by the camera sensor and the total object area $ObjectsArea$, which represents the bounding area covering all objects. These coordinates are then transformed into Cartesian coordinates \((x, y, z)\) following
\vspace{-0.5em}
\begin{equation}
  (x, y, z) = (r \cos \phi \sin \theta, r \sin \phi \sin \theta, r \cos \theta)
  \label{eq:camera}
\end{equation} 

By uniformly sampling \( r \), \( \theta \), and \( \phi \) within specified ranges, the camera is randomly positioned on a spherical shell to control the object scale. \textbf{\textit{Focus point:}} Camera aimes at the center of \(ObjectsArea\) with random shifts along the XYZ axes. 
\textbf{\textit{FOV:}} Focal length is randomized to control zoom. 

\textbf{3D scene initialization} -- \textbf{\textit{Background:}} A 3D scene hosts the objects and camera, consisting of a ground plane and four vertical walls connected to its edges. The z-coordinate of the ground plane is set to zero, ensuring all objects are attached to it, simulating point gravity. The scene size is determined by the camera \( r \) and \(ObjectsArea\) to provide sufficient space. For background randomization, a randomly selected image from the BG-20K dataset \cite{bg20kDataset} is applied to the ground plane and walls. \textbf{\textit{Distractors:}} Six base 3D meshes (cubes, spheres, cones, etc.) from Blender \cite{blender} are added as other objects without gravity constraints as distractors. 

\textbf{Illuminations sampling} -- \textbf{\textit{Quantity:}} Area lights are used, with the number determined by the scene size to ensure balanced lighting without over- or under-illumination. \textbf{\textit{Energy \& Color:}} Light intensity is randomly assigned, and the color is randomly selected from RGB values.

\textbf{Post-processing} -- \textbf{\textit{Noise \& Blur:}} Random salt \& pepper noise and Gaussian blur are applied with varying strength and probability.

\subsection{Implementation details}

Blender and its Python API (bpy) \cite{blender_bpy} are used to implement the domain randomization components. \figref{fig:flowchart} shows the flow of our generation pipeline and evaluation method. Random settings are specified in a configuration file, and once selected, the pipeline generates synthetic data accordingly. During generation, segmentation masks and bounding boxes are created based on 2D pixel coordinates in the camera viewport, with masks saved as images and bounding boxes in YOLO format. For rendering, we use either the Cycles engine \cite{blender_cycles} for path tracing or the Eevee engine \cite{blender_evee} for rasterization.

After generating, the synthetic data is split into training and validation sets to train the Yolov8 model for object detection. Yolov8, chosen for its robustness and efficiency, is pre-trained on the COCO dataset. The model is then evaluated on real images using metrics like mean Average Precision (mAP), precision, and recall. These metrics are essential for evaluating object detection models in manufacturing, where precision and reliability are vital for quality inspection and robotic picking. mAP measures accuracy across categories, while precision and recall assess object identification and localization, revealing false positives and negatives.

\section{Dataset}
\label{sec:dataset}

\begin{figure*}[t]
    \centering

    \subfigure[Robotics use case]{
        \includegraphics[width=0.21\textwidth]{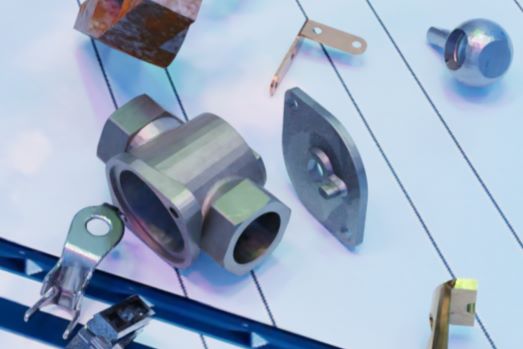}
    }\hspace{-0.5em}
        \vspace{-0.01\textwidth}
    \subfigure[U1]{
        \includegraphics[width=0.21\textwidth]{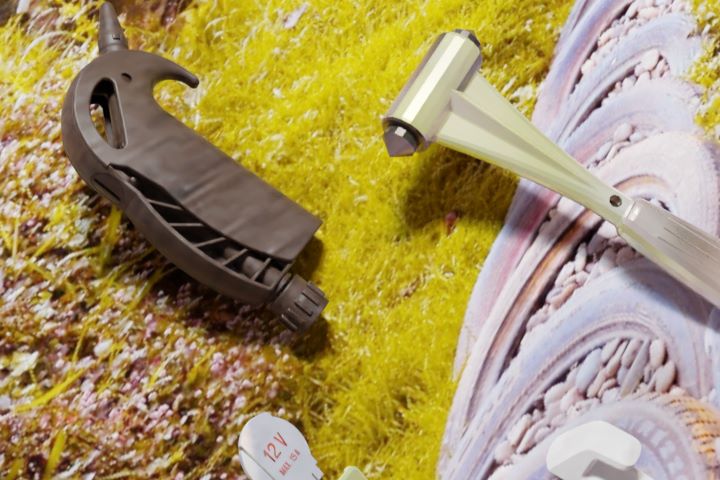}
    }\hspace{-0.5em}
    \subfigure[U2]{
        \includegraphics[width=0.21\textwidth]{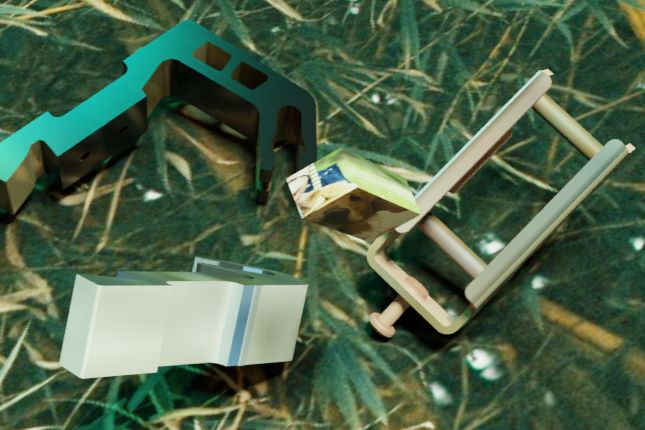}
    }\hspace{-0.5em}
    \subfigure[U3]{
        \includegraphics[width=0.21\textwidth]{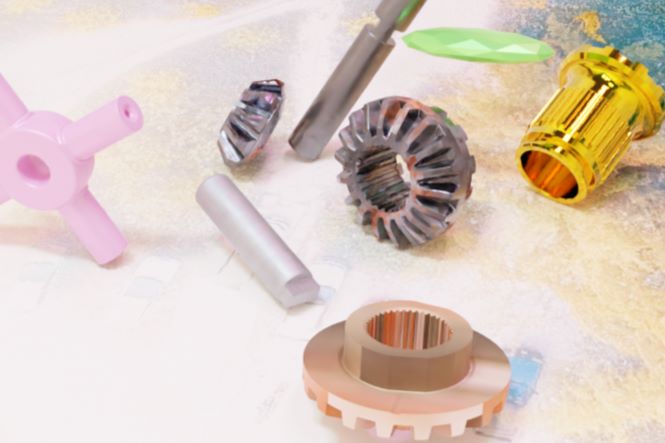}
    }

    \subfigure[Robotics use case \cite{horvath2022object}]{
        \includegraphics[width=0.21\textwidth]{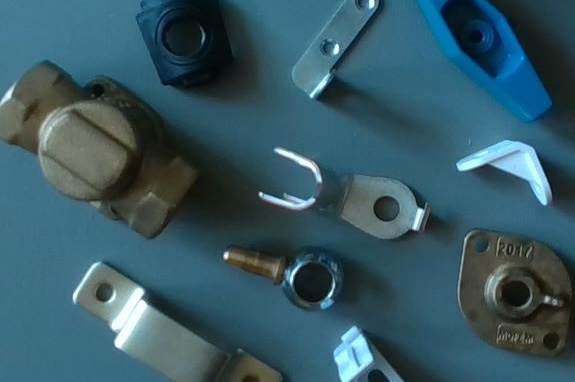}
    }\hspace{-0.5em}
    \subfigure[U1-S1]{
        \includegraphics[width=0.21\textwidth]{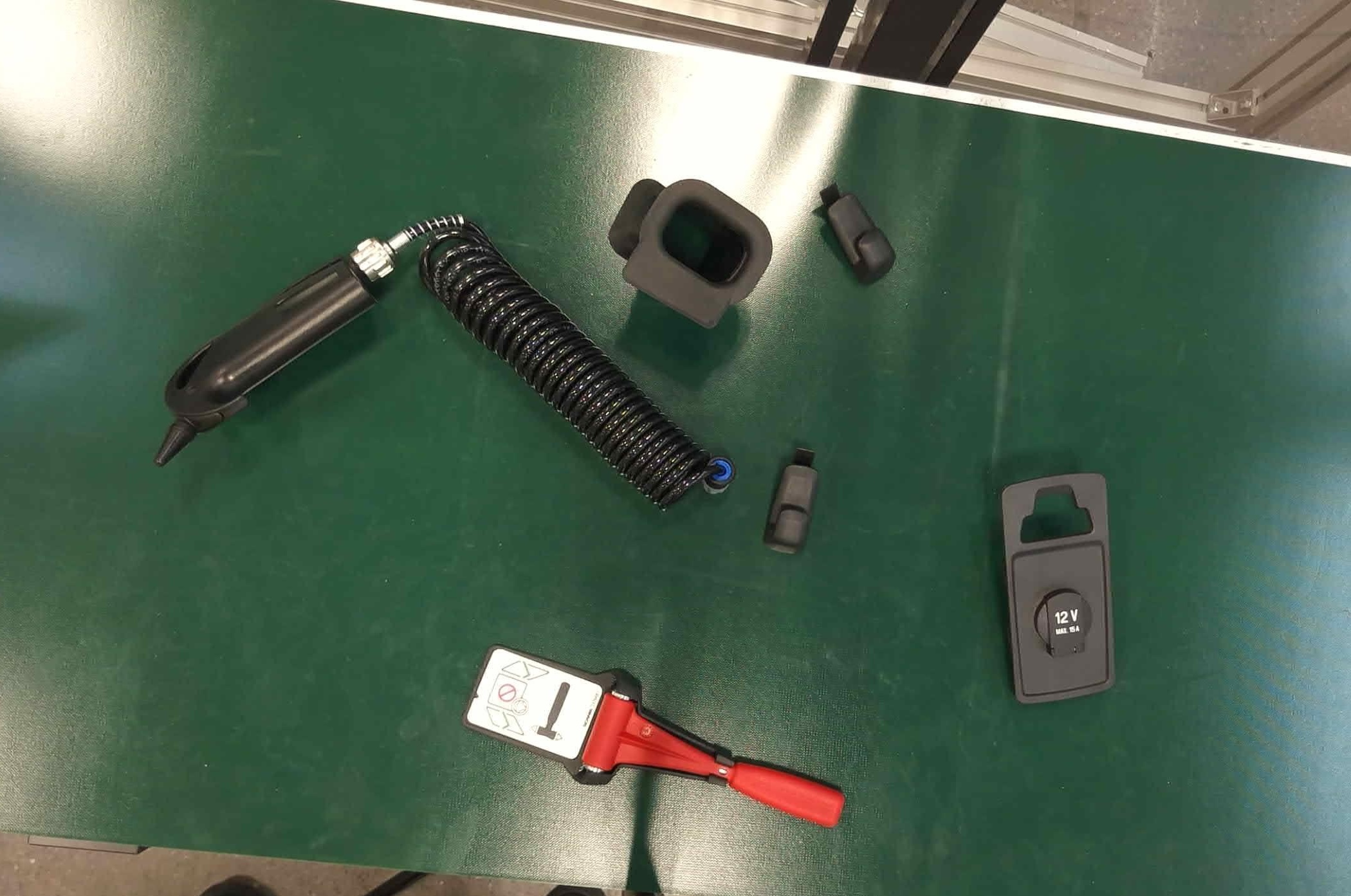}
    }\hspace{-0.5em}
    \subfigure[U1-S2]{
        \includegraphics[width=0.21\textwidth]{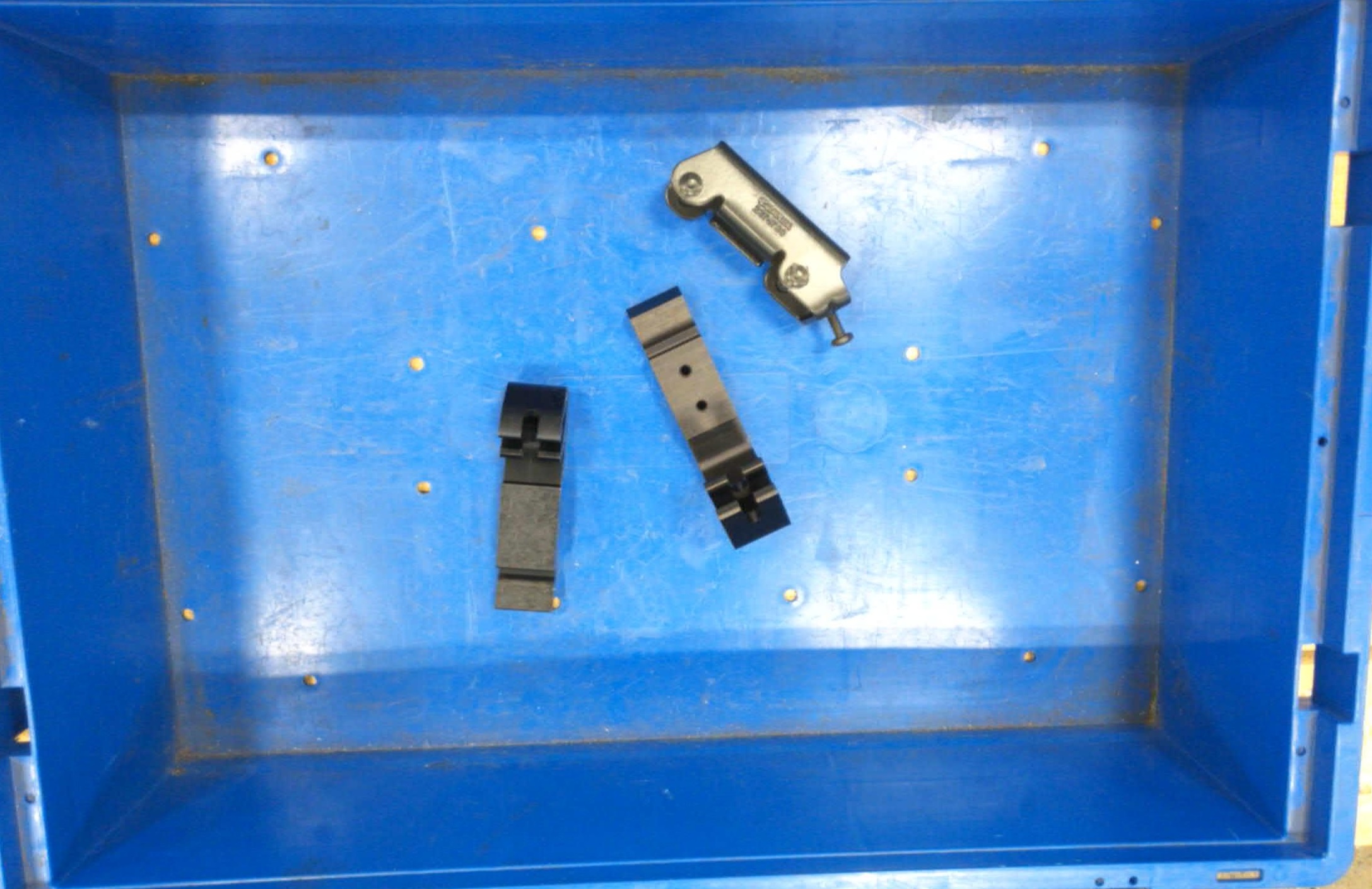}
    }\hspace{-0.5em}
    \subfigure[U3-S3]{
        \includegraphics[width=0.21\textwidth]{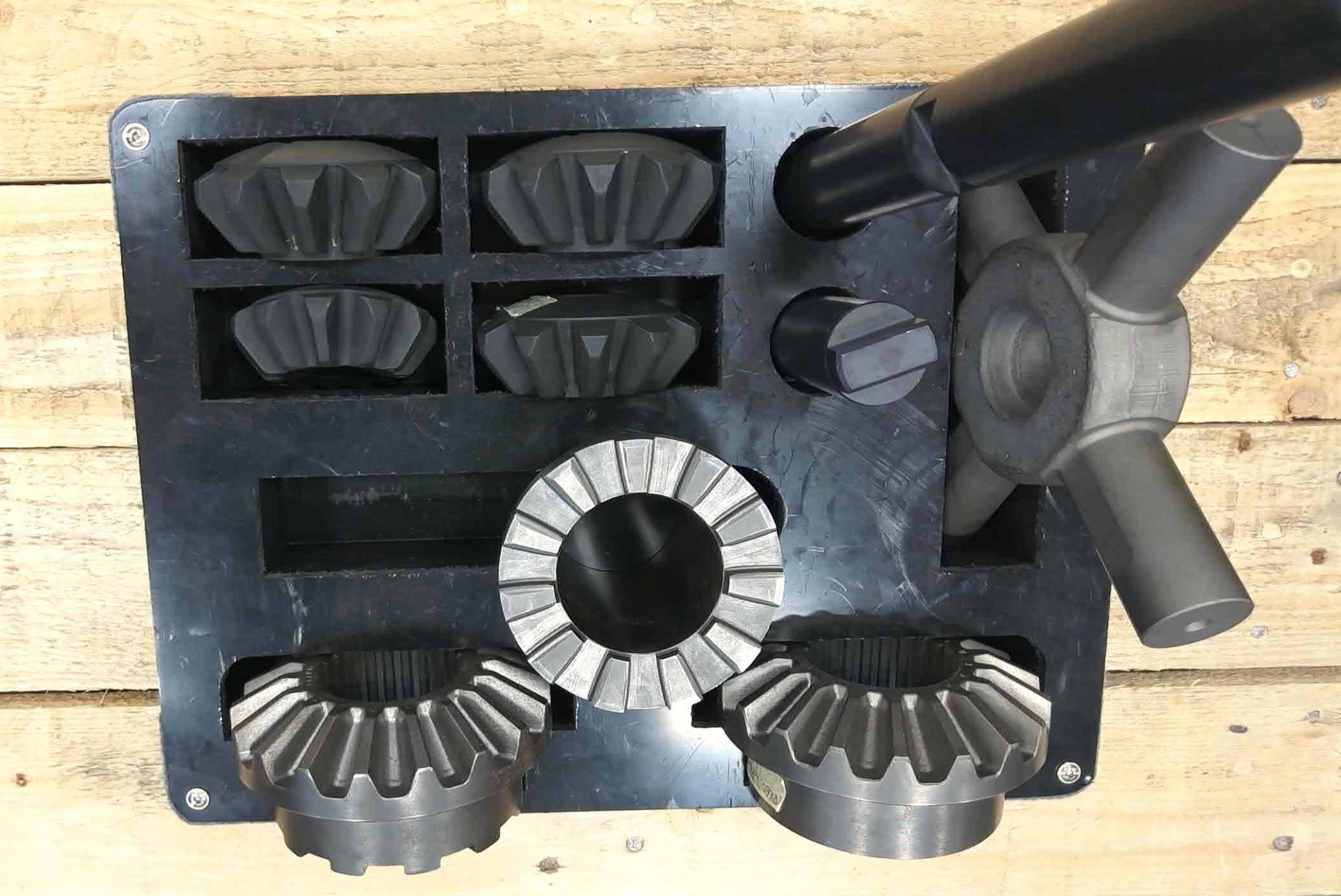}
    }
    \caption{(a-d) Samples of synthetic images. (e-h) Samples of real images. (e) Robotics use cases \cite{horvath2022object}. (f) S1: objects on a conveyor belt (U1 as example). (g) S2: objects in a blue delivery box (U2 as example). (h) S3: objects in their workstation (U3, placed in a black fixture). All use cases have images from S1 and S2; Only U3 has images from S3. Images are cropped for layout, maintaining original aspect ratios.}
    \label{fig:dataset}

\end{figure*}

The pipeline was evaluated on two datasets. The first is a robotic dataset \cite{horvath2022object} includes 10 industrial 3D models and 190 manually annotated real images. Sample real and synthetic images are shown in \figref{fig:dataset} (e) and \figref{fig:albedo} (a). It provides two benchmarks: zero-shot, using only synthetic images for training, and one-shot, combining synthetic images with one real image for training.

\begin{figure}[t]
    \vspace{-0.01\textwidth}
    \centering
    \includegraphics[width=\linewidth]{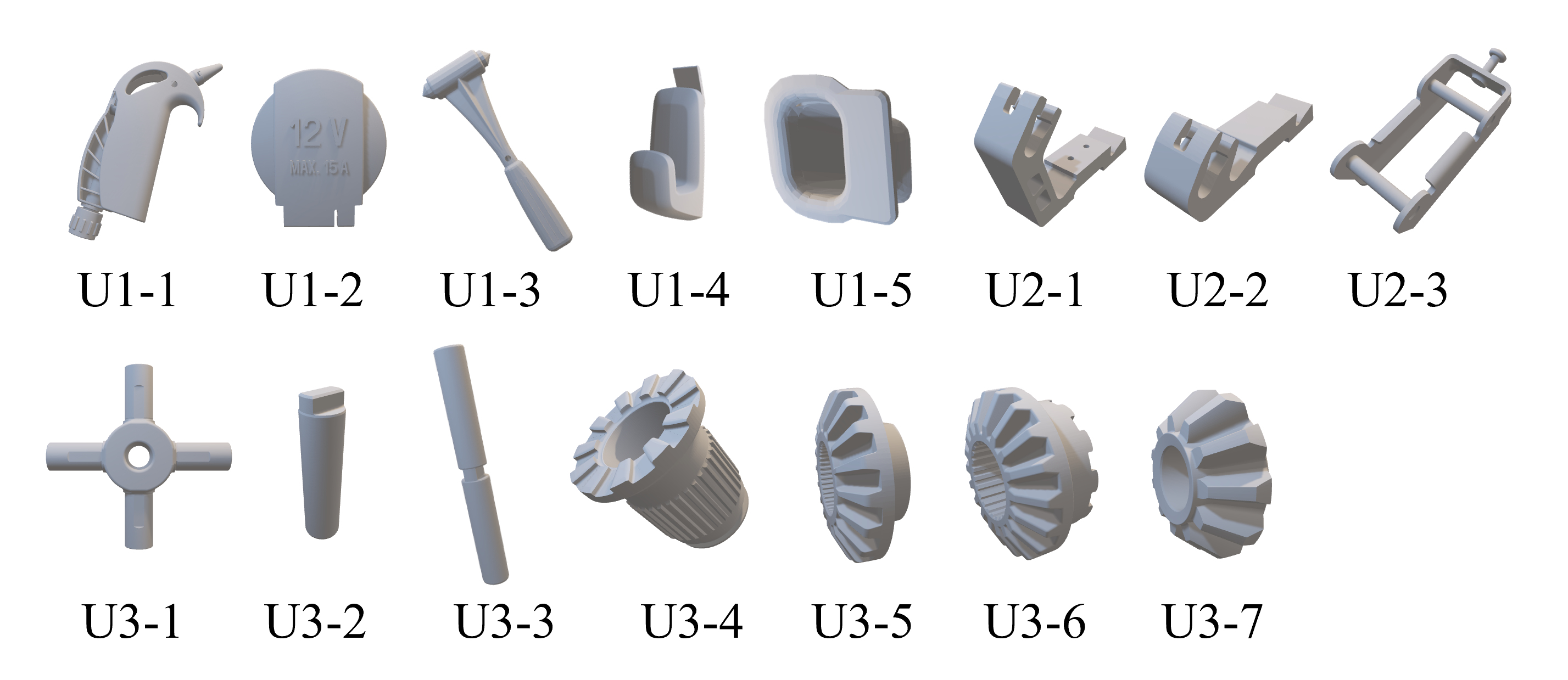}
    \caption{3D models of the SIP15-OD dataset. Their names are as follows in order of their number: Use case 1 (U1): 1. \textit{airgun}, 2. \textit{electricity12v}, 3. \textit{hammer}, 4. \textit{hook}, 5. \textit{plug}; Use case 2 (U2): 1. \textit{fork1}, 2. \textit{fork2}, 3. \textit{fork3}; Use case 3 (U3): 1. \textit{cross}, 2. \textit{pin1}, 3. \textit{pin2}, 4. \textit{couplinghalf}, 5. \textit{gear1}, 6. \textit{gear2}, 7. \textit{pinion}.
}
    \label{fig:3Dmodels}
\end{figure}

The second is the SIP15-OD dataset we developed for manufacturing logistic object detection. It includes 15 3D models from three real-world manufacturing use cases, as shown in \figref{fig:3Dmodels}. Use case 1 (U1) involves five objects from a truck cabin, including black, textureless parts like the \textsl{plug}, \textsl{airgun}, \textsl{electricity12v}, and \textsl{hook}. Use case 2 (U2) involves three similar objects from the roof of a truck, with \textsl{fork1} and \textsl{fork2} being partially identical. Use case 3 (U3) involves seven objects from a truck transmission, all in metal material. Objects like \textsl{cross}, \textsl{pin1}, and \textsl{pin2} have similar side profiles, while \textsl{couplinghalf}, \textsl{gear1}, \textsl{gear2}, and \textsl{pinion} share visual similarities. These objects illustrate common industrial challenges such as textureless or metallic surfaces, geometric complexities, and subtle differences among similar categories. Samples of synthetic images generated from our pipeline for both datasets are displayed in \figref{fig:dataset} (a-d).

 The SIP15-OD dataset includes 321 real images with 877 manually annotated objects, captured in varying scenarios for each use case. Scenario 1 (S1) shows parts on a conveyor belt, Scenario 2 (S2) depicts parts in blue delivery boxes, and Scenario 3 (S3) captures parts in their workstation: a black fixture or a plain surface (wooden table or padded surface) for U3. All use cases have images from S1 and S2, but only U3 includes images from S3. The dataset focuses on manufacturing logistics scenarios, where object detection models are used for verifying correct object placement for quality control or guiding robots in part-picking. Images were captured using an iPhone 12 Pro with a 12 MP camera (4032x3024 resolution) and a Poly Studio P5 webcam (2560x1440 resolution). Samples of real images are shown in \figref{fig:dataset} (e-h), with the number of real images per scenario summarized in \tabref{tab:dataset_overview}.

\begin{table}[t]
\caption{The number of real images in the SIP15-OD dataset.}
\label{tab:dataset_overview}
\centering
\begin{tabular}{>{\centering\arraybackslash}p{0.8cm} 
                >{\centering\arraybackslash}p{1.2cm} 
                >{\centering\arraybackslash}p{0.8cm} 
                >{\centering\arraybackslash}p{0.8cm} 
                >{\centering\arraybackslash}p{0.8cm} 
                >{\centering\arraybackslash}p{0.8cm}}

\hline
& \textbf{Categories} & \textbf{All} & \textbf{S1} & \textbf{S2} & \textbf{S3} \\ 
\hline
U1         & 5    & 70   & 27   & 43   & -  \\ 
U2         & 3    & 91    & 27   & 64   & -    \\ 
U3         & 7    & 160   & 46   & 61   & 53   \\ 
Total      & 15   & 321   & 100  & 168  & 53  \\ 
\hline
\end{tabular}
\end{table}


\section{Experimental results}
\label{sec:results}

All experiments were run on A100 Tensor Core GPUs. Unless otherwise specified, all images were rendered at 720x720 resolution using the Cycles engine in Blender with path tracing. The Yolov8 models were trained with default settings from Ultralytics \cite{Jocher_Ultralytics_YOLO_2023}, using a batch size of 80 and an image size of 720. Rendering times using 10 Blender instances ranged from 1 to 3 seconds per image, depending on the size of 3D meshes and their texture settings. Training a Yolov8 model with 4000 synthetic images for 500 epochs took around four hours on a single GPU. Each experiment was repeated three times, and average results were reported. For all results, an IoU threshold of 0.6 and a confidence threshold of 0.5 were used.

\subsection{Results on the robotic dataset}

\begin{table}[t]
\caption{Overall mAP@50 scores (\%) on the robotic dataset.}
\label{tab:rball}
\centering
\begin{tabular}{>{\centering\arraybackslash}p{0.4cm} 
                >{\centering\arraybackslash}p{0.9cm} 
                >{\centering\arraybackslash}p{0.8cm} 
                >{\centering\arraybackslash}p{0.8cm} 
                >{\centering\arraybackslash}p{1.8cm} 
                >{\centering\arraybackslash}p{1.6cm} 
                >{\centering\arraybackslash}p{0.5cm}}
\hline
& \multicolumn{3}{c}{\textbf{Zero-shot}} & \multicolumn{2}{c}{\textbf{One-shot}} \\
\cline{2-4} \cline{5-6}
\textbf{Size} & \textbf{Horvath \cite{horvath2022object}} & \textbf{Ours Yolov4} & \textbf{Ours Yolov8} & \textbf{Horvath \cite{horvath2022object} (2000 syn)} & \textbf{Ours} \\
\hline
4k  & 83.2  & 88.4  & 96.1  & 97.1  & 97.5  \\ 
8k  & 83.6  & 90.4  & 96.4  & -     & -     \\ 
\hline
\end{tabular}
\end{table}

Ten industrial models in \cite{horvath2022object} were used to generate synthetic data with our pipeline. For a fair comparison, the data was trained with both Yolov4 (used in \cite{horvath2022object}) and Yolov8. Optimal results were achieved by applying the PBR metal material texture from the CG Texture dataset \cite{Demes2023Texture}, with other parameters randomized. As shown in \tabref{tab:rball}, our pipeline increased the mAP@50 score by more than 5\% compared to \cite{horvath2022object} with the same training data and model. We achieved 96.4\% mAP@50 in zero-shot with Yolov8, nearly matching the one-shot model in \cite{horvath2022object}. In their one-shot experiments, \cite{horvath2022object} duplicated one real image 2000 times to train alongside synthetic data. However, training with synthetic data and then fine-tuning with one real image performed better in our experiments, achieving 97.5\% mAP@50. Compared to \cite{horvath2022object}, our one-shot results showed slight improvement because the zero-shot results were already high.

\subsection{Results on the SIP15-OD dataset}

\begin{table}[t]
\caption{Overall mAP scores (\%) on the SIP15-OD dataset.}
\label{tab:sip-all}
\centering
\renewcommand{\arraystretch}{1.2} 

\begin{tabular}{>{\centering\arraybackslash}p{0.5cm} 
                >{\centering\arraybackslash}p{0.5cm} 
                >{\centering\arraybackslash}p{0.5cm} 
                >{\centering\arraybackslash}p{0.5cm} 
                >{\centering\arraybackslash}p{0.5cm} 
                >{\centering\arraybackslash}p{0.5cm} 
                >{\centering\arraybackslash}p{0.5cm} 
                >{\centering\arraybackslash}p{0.5cm} 
                >{\centering\arraybackslash}p{0.5cm} }
\hline
& \multicolumn{4}{c}{\textbf{mAP@50}} & \multicolumn{4}{c}{\textbf{mAP@50-95}} \\ 
\cline{2-5}\cline{6-9}
\textbf{Size} & \textbf{All} & \textbf{S1} & \textbf{S2} & \textbf{S3} & \textbf{All} & \textbf{S1} & \textbf{S2} & \textbf{S3} \\ 
\hline
U1  & 94.1 & 95.4  & 93.3  & -  & 81.5 & 81.0  & 81.8  & -  \\ 
U2  & 99.5 & 99.5  & 99.5  & -  & 96.2 & 96.7  & 96.1  & -  \\ 
U3  & 95.3 & 92.5  & 95.3  & 95.7  & 92.0 & 89.2  & 92.6  & 91.9  \\ 
\hline
\end{tabular}
\end{table}

The overall mAP@50 and mAP@50-95 scores of the three use cases are summarized in \tabref{tab:sip-all}, where all models were trained on synthetic images but evaluated on real images. Instead of relying on random DR, we systematically identified key factors for object detection in manufacturing and used GDR to selectively adjust parameters, preventing unrealistic scenarios. The best-performing configurations were: In U1, as shown in \figref{fig:dataset} (g), only one side of the objects is visible in their workstation, so we limited the rotation of 3D objects to 30 degrees in the x and y directions, generating 7,500 images in total. In U2, all parameters were randomized, generating 18,000 images. In U3, objects used metal material textures with other parameters randomized, generating 10,500 images. As shown in \tabref{tab:sip-all}, all use cases achieved over 90\% mAP@50 across all scenarios, demonstrating the effectiveness of our pipeline. Additionally, all mAP@50-95 scores remained above 80\%, indicating the robustness of the pipeline across IoU thresholds.

\section{Discussions}
\label{sec:results}
We conducted ablation studies to identify key factors and provide insights into the feasibility and challenges of sim-to-real manufacturing object detection with DR.

\subsection{Key factors for DR}

\begin{figure}[t]
    \centering
    \subfigure[]{
        \includegraphics[width=0.13\textwidth]{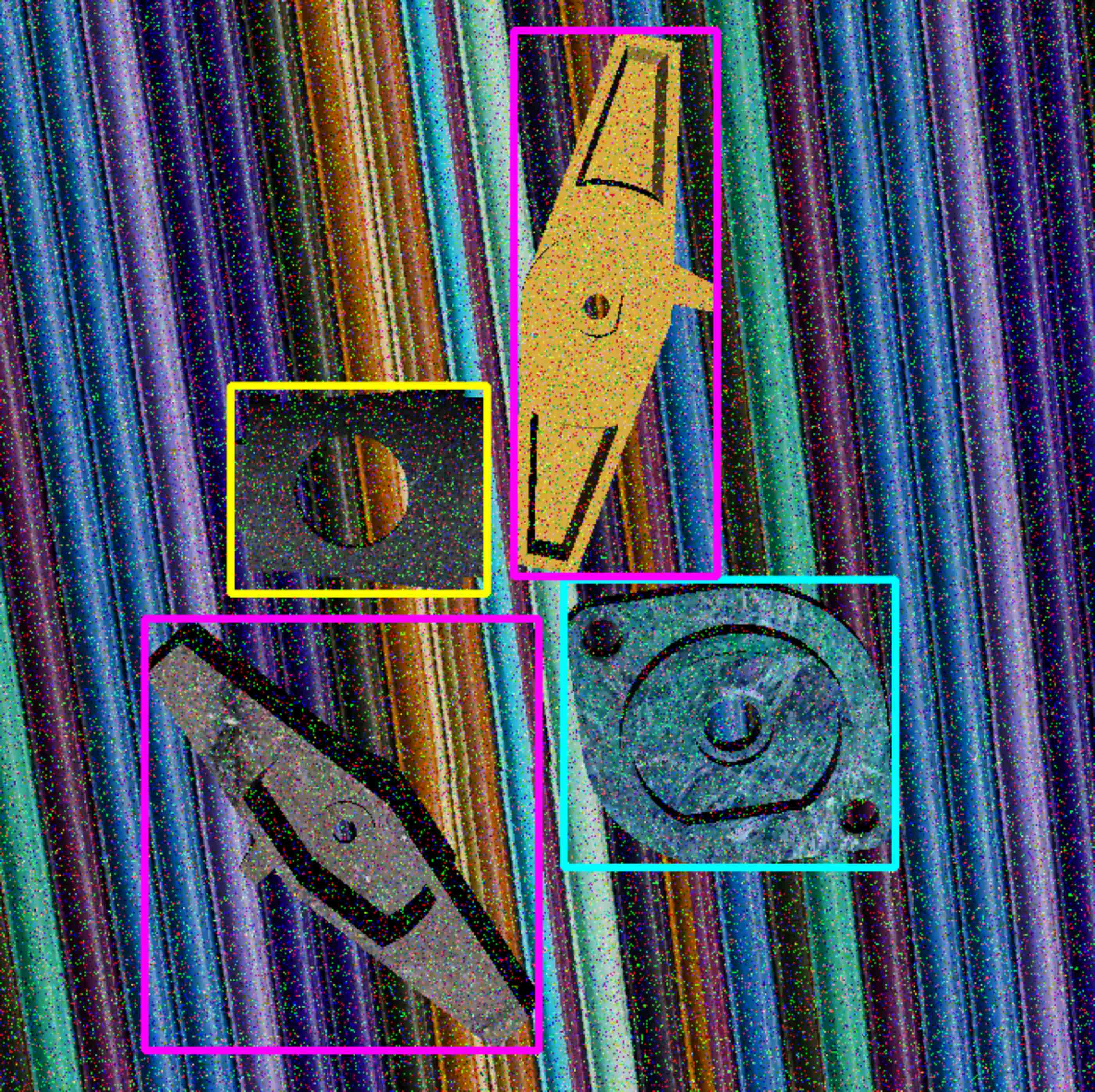}
        \label{fig:albedo-b}
    }\hspace{-0.5em}
    \subfigure[]{
        \includegraphics[width=0.13\textwidth]{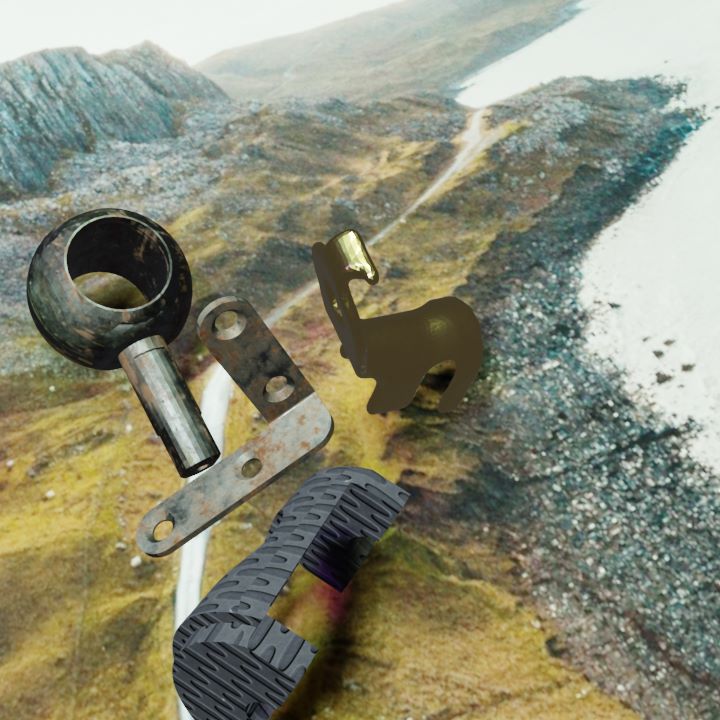}
        \label{fig:albedo-c}
    }\hspace{-0.5em}
    \subfigure[]{
        \includegraphics[width=0.13\textwidth]{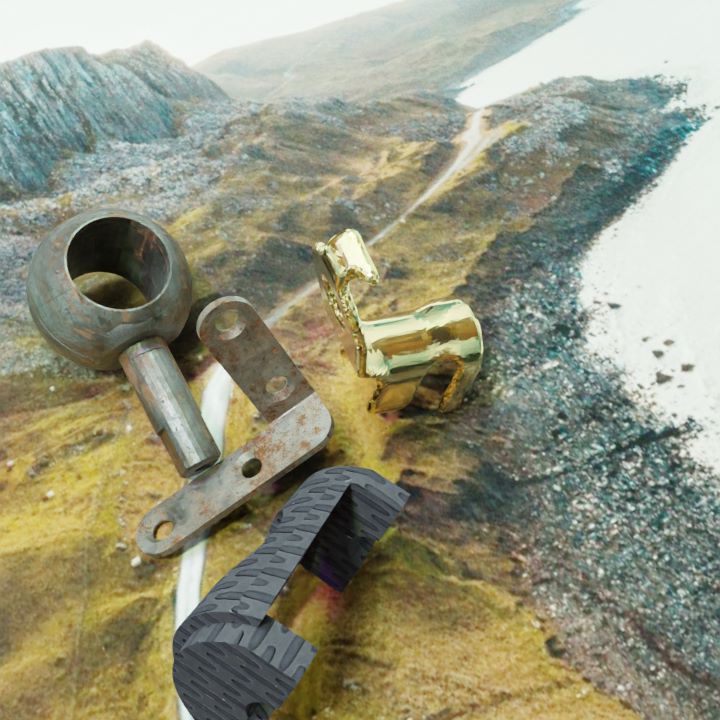}
        \label{fig:albedo-d}
    }
    \vspace{-0.01\textwidth}
    \caption{Comparison of synthetic data rendered with different methods in the robotic use case. (a) Horvath et al. \cite{horvath2022object}, using random material images textures and rasterization (OpenGL). (b) Ours, using PBR metal textures and rasterization (Blender). (c) Ours, using PBR metal textures and path tracing (Blender).}
    \label{fig:albedo}
    \vspace{-0.02\textwidth}
\end{figure}

\textbf{a. Object material properties and rendering methods.} Our results surpass those in \cite{horvath2022object}, likely due to the use of PBR metal material textures, varied illumination, and path tracing rendering. In contrast, \cite{horvath2022object} employs material image textures and rasterization rendering. Our approach produces more realistic surface reflectance and appearance, as path tracing simulates light interactions more accurately by modeling how light bounces off surfaces, providing a more lifelike representation than rasterization. To illustrate these differences, \figref{fig:albedo} shows sample images with different textures and rendering methods from the robotic dataset. (c) generated with PBR materials and path tracing show clear visual improvements over (a) and (b), especially in handling reflective surfaces, reducing the visual gap between synthetic and real metal objects.


\begin{figure*}[t]
  \centering
  \includegraphics[height=0.2\linewidth ]{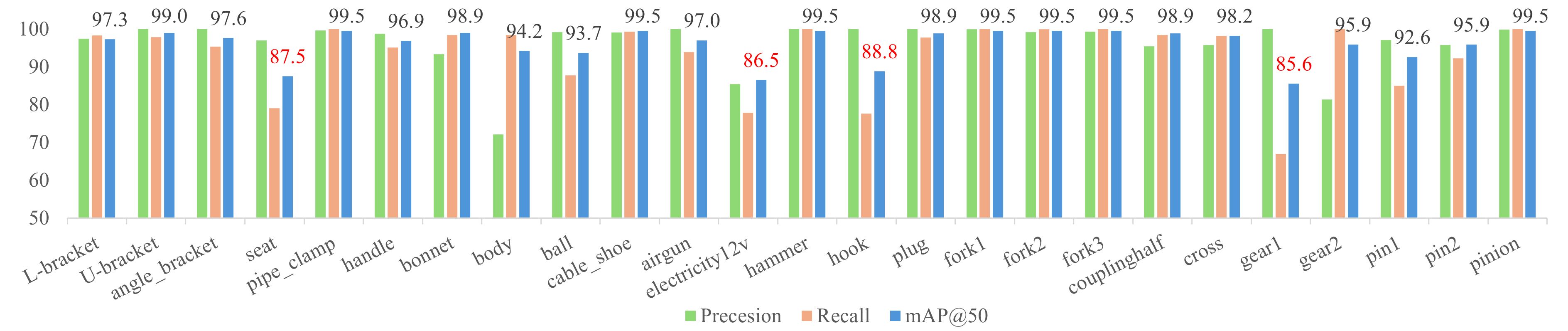}
  \caption{Class-wise results (\%) for each use case. All mAP@50 scores are shown in numbers, with scores below 90\% highlighted in red.}
  \label{fig:classwise-all}
\end{figure*}

\begin{table}[t]
\caption{mAP@50 scores (\%) comparison for textures and rendering methods. }
 \vspace{-0.01\textwidth}
\label{tab:albedo}
\centering
\begin{tabular}{>{\centering\arraybackslash}p{1.2cm} 
                >{\centering\arraybackslash}p{1.3cm} 
                >{\centering\arraybackslash}p{1.3cm} 
                >{\centering\arraybackslash}p{1.3cm} 
                >{\centering\arraybackslash}p{1.3cm}}
\hline
& \multicolumn{2}{c}{\textbf{All textures}} & \multicolumn{2}{c}{\textbf{Realistic textures}} \\
\cline{2-3} \cline{4-5}
\textbf{} & \textbf{R} & \textbf{P} & \textbf{R} & \textbf{P} \\
\hline
Robotics  & 90.2 & 91.9 & 89.6 & \textbf{96.4} \\
U1        & 92.0 & \textbf{94.1} & 92.4 & 94.0  \\
U2        & \textbf{99.5} & \textbf{99.5} & \textbf{99.5} & \textbf{99.5} \\
U3        & 94.2 & 94.6 & 94.4 & \textbf{95.3} \\
\hline
\end{tabular}
\begin{flushleft}
"All textures" includes random textures (RGB, image textures, and PBR materials), while "Realistic textures" refers to PBR metal textures for Robotics and U3, PBR metal and plastic textures for U1, and PBR plastic textures for U2. "R" represents rasterization, and "P" represents path tracing rendering. The highest scores of each use case are marked in bold.
\end{flushleft}
\end{table}

Furthermore, \tabref{tab:albedo} compares performance across different use cases using random textures versus realistic material textures, and rendering with rasterization (Eevee engine) versus path tracing (Cycles engine). As shown in \tabref{tab:albedo}, path tracing consistently outperforms rasterization, highlighting its importance. However, the choice of textures depends on the use case. For objects with unique materials, such as those in robotics and U3 (primarily metal), realistic PBR textures improved performance by more accurately representing real-world surfaces. For textureless objects, like those in U1 and U2 (often black and without texture), random and realistic textures yielded similar results.

\textbf{b. Post-processing and distractors.} Post-processing and distractors generally improved model performance, as summarized in \tabref{tab:various_conditions}. Noise and blur were applied with a 10\% probability, and removing them caused performance drops in Robotics, U1, and U3, underscoring their importance. This suggests that even with extensive DR, post-processing further boosts model generalization and robustness.

Distractors, which add irrelevant objects to images, help the model handle cluttered environments, particularly improving performance in the Robotics and U1. In U2 and U3, performance remained similar regardless of post-processing and distractors, indicating their impact varies by scenario. While the improvements may not always be significant, they are generally beneficial. Based on existing research and our findings, we recommend always including post-processing and distractors when generating synthetic images.

\begin{table}[t]
\caption{mAP@50 score (\%) comparison for pp and d.}
\label{tab:various_conditions}
\centering
\begin{tabular}{>{\centering\arraybackslash}p{1.2cm} 
                >{\centering\arraybackslash}p{1.2cm} 
                >{\centering\arraybackslash}p{1.4cm} 
                >{\centering\arraybackslash}p{1.4cm}}
\hline
& \textbf{Zero-shot} & \textbf{Without pp} & \textbf{Without d} \\
\hline
Robotics  & \textbf{96.4} & 90.0  & 88.9  \\
U1        & \textbf{94.1} & 92.0  & 90.1  \\
U2        & \textbf{99.5} & \textbf{99.5} & \textbf{99.5} \\
U3        & \textbf{95.3} & 94.4  & 95.1  \\
\hline
\end{tabular}
\begin{flushleft}
"pp" represents post-processing, and "d" represents distractors. The highest scores of each use case are marked in bold
\end{flushleft}
\end{table}

\subsection{Other insights for DR}

\begin{table}[t]
\caption{Ideal Data size (per category) and training epoch for different use cases. }
\label{tab:datasize}
\centering
\begin{tabular}{>{\centering\arraybackslash}p{1.1cm} 
                >{\centering\arraybackslash}p{1.3cm} 
                >{\centering\arraybackslash}p{1cm} 
                >{\centering\arraybackslash}p{1.3cm} 
                >{\centering\arraybackslash}p{1cm}}
\hline
& \multicolumn{2}{c}{\textbf{All textures}} & \multicolumn{2}{c}{\textbf{Realistic textures}} \\
\cline{2-3} \cline{4-5}
\textbf{} & \textbf{Data sizes} & \textbf{Epochs} & \textbf{Data sizes} & \textbf{Epochs} \\
\hline
Robotics  & 800  & 500  & 800  & 500  \\
U1        & 1.5k & 2.5k & 1.5k & 1k   \\
U2        & 6k   & 1k   & 3k   & 500  \\
U3        & 3k   & 1k   & 1.5k & 1k   \\
\hline
\end{tabular}
\end{table}

\textbf{Data sizes and training epochs.} Object detection performance improves with increased data size up to a limit, after which additional synthetic images no longer boost accuracy, consistent with previous studies. A similar effect is observed with training epochs. When training a deep learning model, performance improves with more epochs until overfitting, typically mitigated by a validation set and early stopping. However, in zero-shot training, a synthetic validation set may not reliably identify the best model for real test data due to distribution differences. \tabref{tab:datasize} presents data size limits and ideal training epochs that produced the best results. Training beyond these limits did not improve performance in our experiments. These settings were used to obtain the best results with path tracing in \tabref{tab:albedo}.

Data size limits seem influenced by the complexity of real data backgrounds. In the Robotic dataset with a controlled background (\figref{fig:albedo} (a)), 400 images per category were sufficient, with only a slight improvement (0.3\% mAP@50) up to 800 images. In contrast, SIP15-OD, with more varied real-world data, required at least 1,500 images per category for strong performance. Beyond that, U2 and U3 improved slightly (about 1\%) with 6K and 3K images, respectively. Specific textures helped reduce the required data size by lowering randomization in the data.

Regarding training epochs, all model performance on synthetic validation data plateaued within 500 epochs, but performance on real test data continued to improve and stabilized after 1,000 epochs for the SIP15-OD dataset. Therefore, higher early stopping thresholds can be suggested to allow the model to train longer with synthetic data, which may enhance its performance on real data.

\subsection{Failure cases in DR}

Class-wise results of all use cases in \figref{fig:classwise-all} show low recall and mAP for categories \textsl{seat}, \textsl{electricity12v}, \textsl{hook}, and \textsl{gear1}. Two primary failure cases were identified:

\textbf{a. Missed detection.}  Low performance is mainly due to low recall and false negatives, where objects like \textsl{electricity12v}, \textsl{hook}, and \textsl{pin1} are misclassified as background. This persists even with a lowered confidence threshold, likely due to their simple geometric shapes (e.g., circles, squares, cylinders as shown in \figref{fig:3Dmodels}). During DR, with random textures and backgrounds, the model may rely heavily on shape, and these simple shapes can make it difficult for the model to use them as distinguishing features, causing challenges in learning.

\textbf{b. Wrong detection.} Categories like \textsl{seat} and \textsl{gear1} perform poorly due to confusion with similar objects (e.g., \textsl{seat} detected as \textsl{body}, \textsl{gear1} as \textsl{gear2}). This is caused by their visual similarities. As illustrated in \figref{fig:3Dmodels}, \textsl{gear1} and \textsl{gear2} are partially identical and hard to differentiate, even in real images. In the robotic dataset, \textsl{body} and \textsl{seat} appear similar from certain angles, but they differ in scale in real images, suggesting that one-shot learning may improve the performance of the \textsl{seat} category.

\section{Conclusion and future work}
\label{sec:conclusion}

This paper presents a synthetic data generation pipeline for manufacturing object detection, incorporating domain randomization in object characteristics, background, illumination, camera view, and post-processing. The pipeline was evaluated on a public robotics dataset \cite{horvath2022object} and the SIP15-OD dataset we developed, featuring 15 categories from three manufacturing use cases in varied environments. Through different ablation studies, we identified key factors for effective synthetic data generation, including the use of path tracing, post-processing, and distractors. Material properties also proved critical. For unique materials like metal, structured realistic material textures outperform random textures. Additionally, we provide insights on optimal data sizes, training epochs, and challenges in DR, such as missed and incorrect detections.

The pipeline achieved mAP@50 scores ranging from 94.1\% to 99.5\% across four use cases, with the Yolov8 model trained only on synthetic data. These results demonstrate that the pipeline generates data that closely covers the distribution of real data, showing its robustness for various manufacturing applications like quality inspection and robotic part-picking.

However, categories with visually similar objects and simple shapes exhibited lower performance. Future work should refine the data generation process to better target these categories, improving performance balance and overall detection accuracy.

\section{Acknowledgement}
\label{acknowledgement}

This work was partially supported by the Wallenberg AI, Autonomous Systems and Software Program (WASP) funded by the Knut and Alice Wallenberg Foundation. The computations were enabled by the Berzelius resource provided by the Knut and Alice Wallenberg Foundation at the National Supercomputer Centre. 

We gratefully acknowledge our colleagues at the Production Oskarshamn, Production Zwolle, Transmission Assembly, Engine Assembly, Academy, and Smart Factory Lab
Departments at Scania CV AB for providing the CAD models and use cases. We also extend our thanks to Prof. Joakim Lindblad at the Department of Information Technology, Uppsala University, for his valuable insights and constructive feedback on this study. 

\appendix

\paragraph{Related Work} Different studies have explored domain randomization (DR) for synthetic data generation in various manufacturing object detection applications. The DR techniques used in these studies, including ours, are summarized in \cref{appendix:tab_dr_comparison}.

\begin{figure*}[t]
    \centering
    \subfigure[3D scene.]{%
        \includegraphics[width=0.22\textwidth]{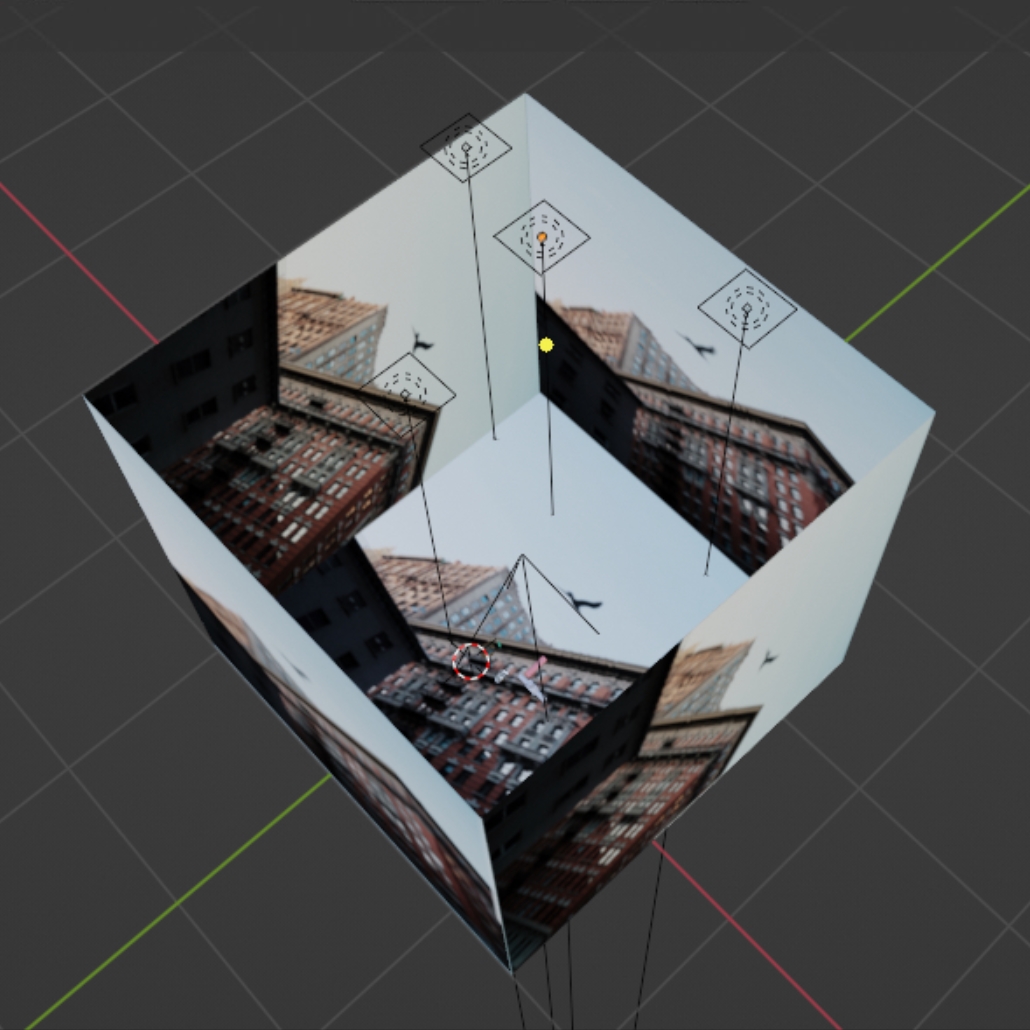}%
        \label{fig:subimage-a}%
    }\hspace{-0.5em}%
    \subfigure[Annotations.]{%
        \includegraphics[width=0.22\textwidth]{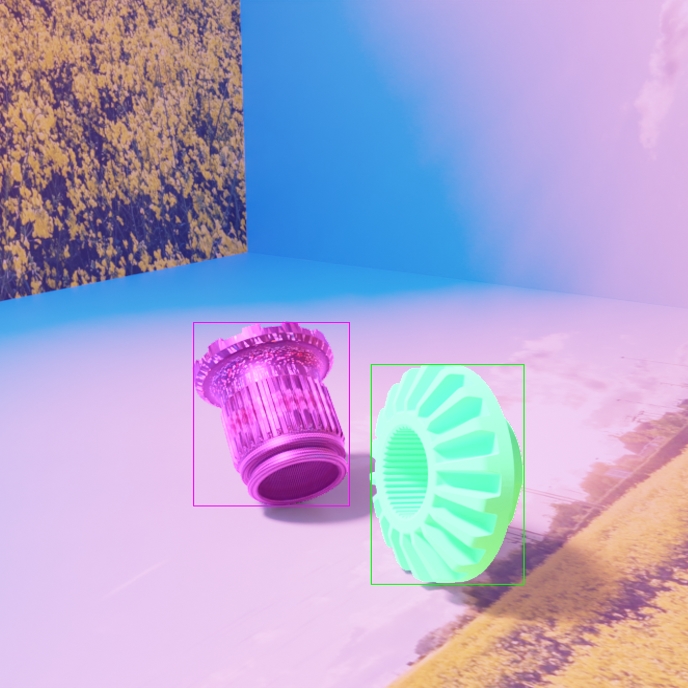}%
        \label{fig:subimage-b}%
    }\\[-0.2em]
    \subfigure[Objects with RGB textures.]{%
        \includegraphics[width=0.22\textwidth]{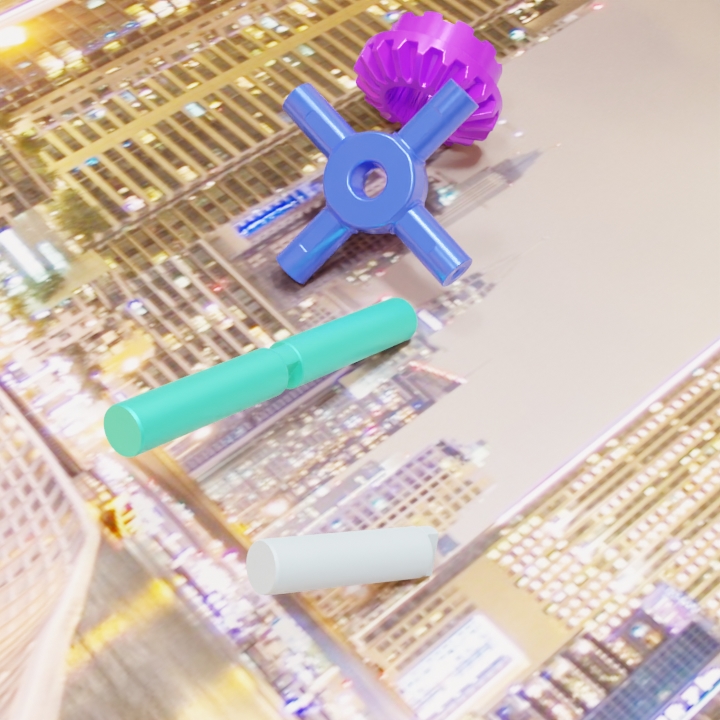}%
        \label{fig:subimage-c}%
    }\hspace{-0.5em}%
    \subfigure[Object with image textures.]{%
        \includegraphics[width=0.22\textwidth]{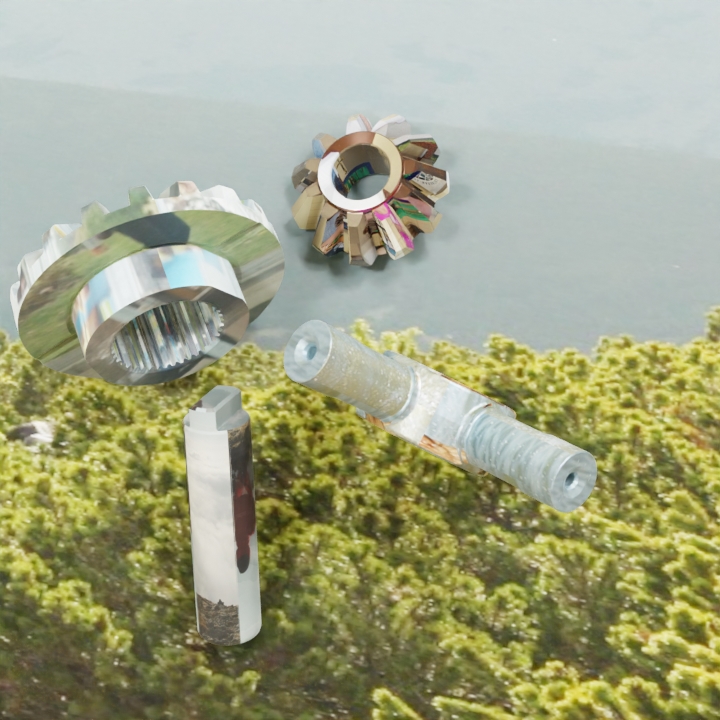}%
        \label{fig:subimage-d}%
    }\hspace{-0.5em}%
    \subfigure[Object with material textures.]{%
        \includegraphics[width=0.22\textwidth]{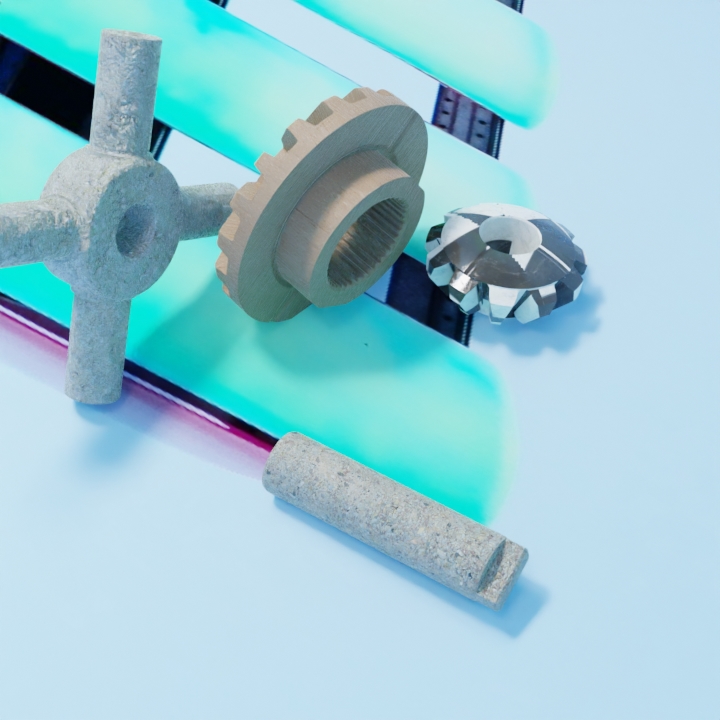}%
        \label{fig:subimage-e}%
    }\hspace{-0.5em}%
    \subfigure[Object with metal textures.]{%
        \includegraphics[width=0.22\textwidth]{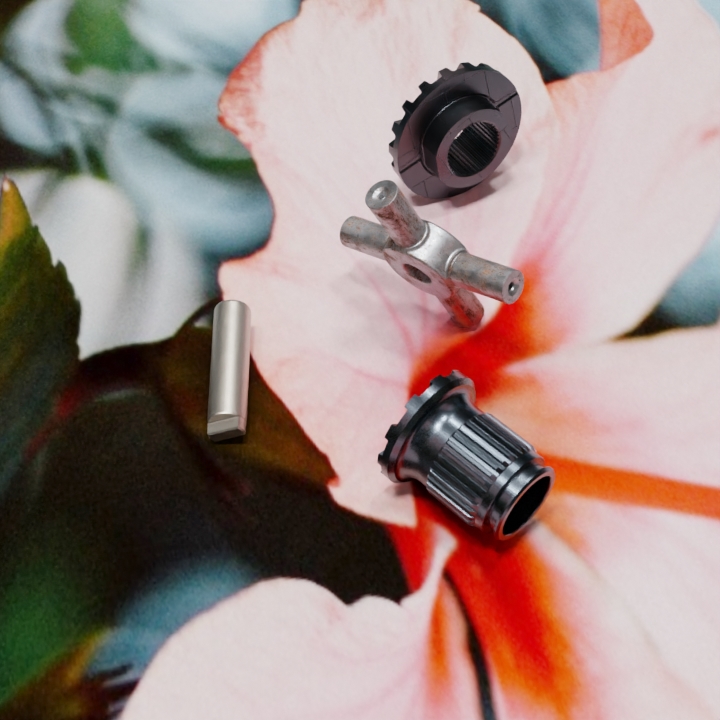}%
        \label{fig:subimage-f}%
    }\\[-0.2em]
    \subfigure[Distractors. ]{%
        \includegraphics[width=0.22\textwidth]{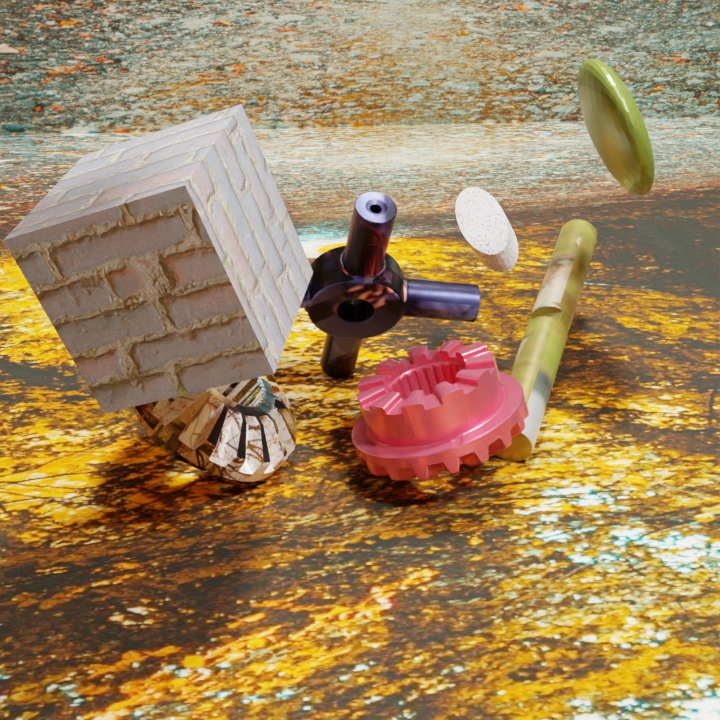}%
        \label{fig:subimage-g}%
    }\hspace{-0.5em}%
    \subfigure[Camera with varied view. ]{%
        \includegraphics[width=0.22\textwidth]{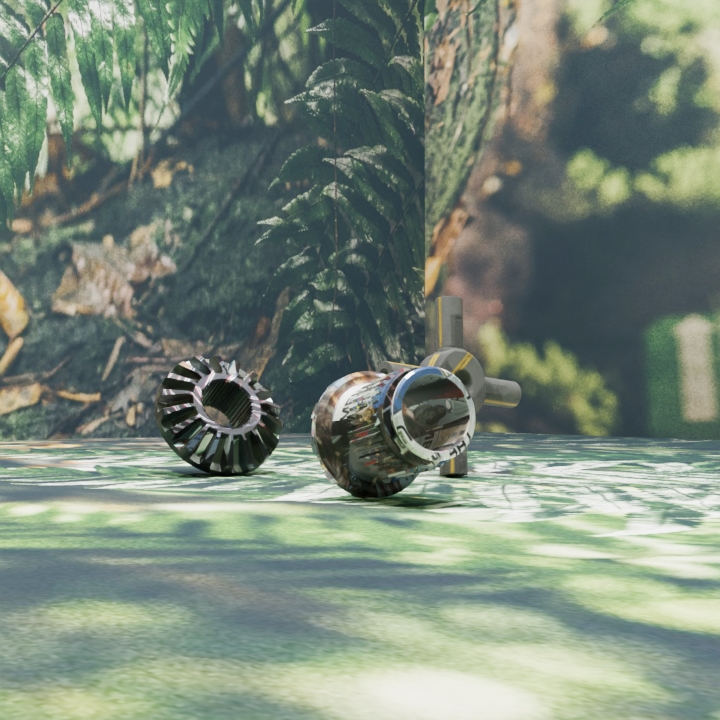}%
        \label{fig:subimage-h}%
    }\hspace{-0.5em}%
    \subfigure[Camera with view shift.]{%
        \includegraphics[width=0.22\textwidth]{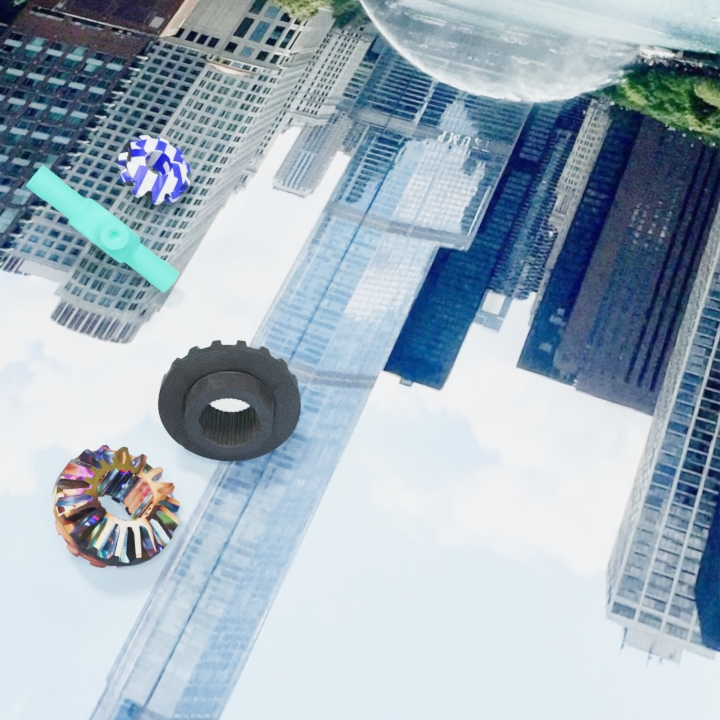}%
        \label{fig:subimage-i}%
    }\hspace{-0.5em}%
    \subfigure[Camera with small FOV.]{%
        \includegraphics[width=0.22\textwidth]{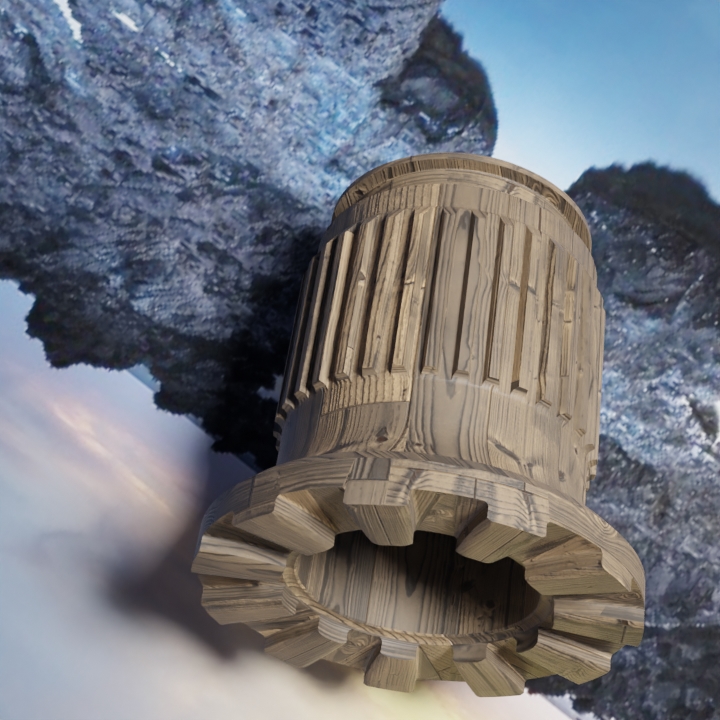}%
        \label{fig:subimage-j}%
    }\\[-0.2em]
    \subfigure[Camera with big FOV.]{%
        \includegraphics[width=0.22\textwidth]{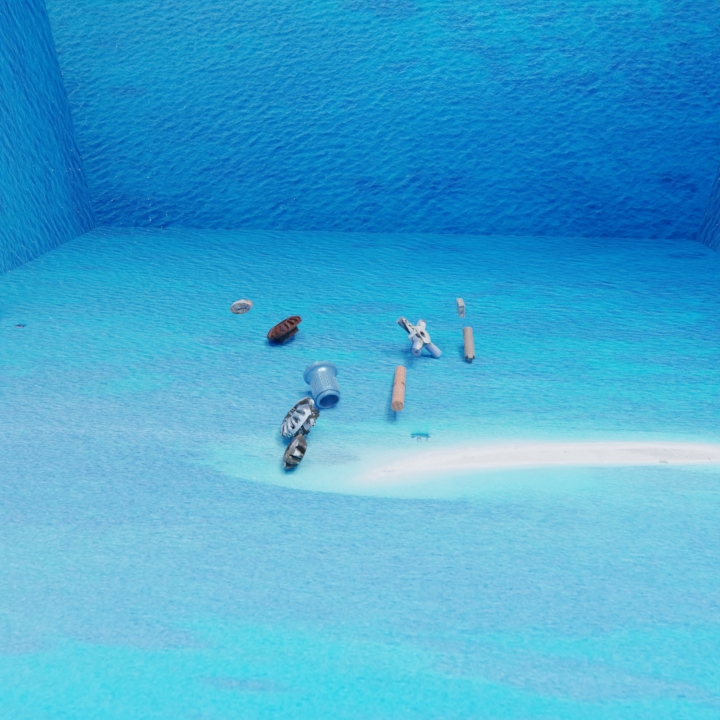}%
        \label{fig:subimage-k}%
    }\hspace{-0.5em}%
    \subfigure[Strong lights.]{%
        \includegraphics[width=0.22\textwidth]{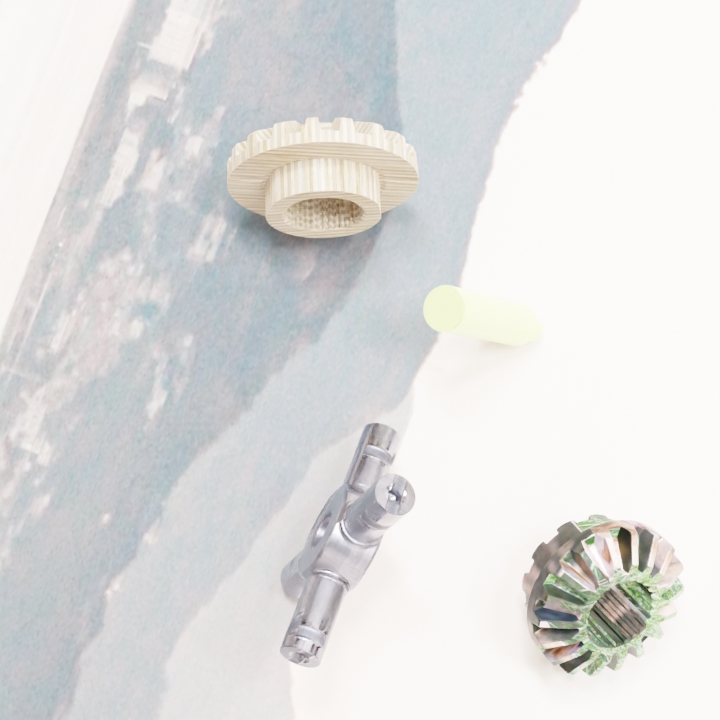}%
        \label{fig:subimage-l}%
    }\hspace{-0.5em}%
    \subfigure[Weak lights. ]{%
        \includegraphics[width=0.22\textwidth]{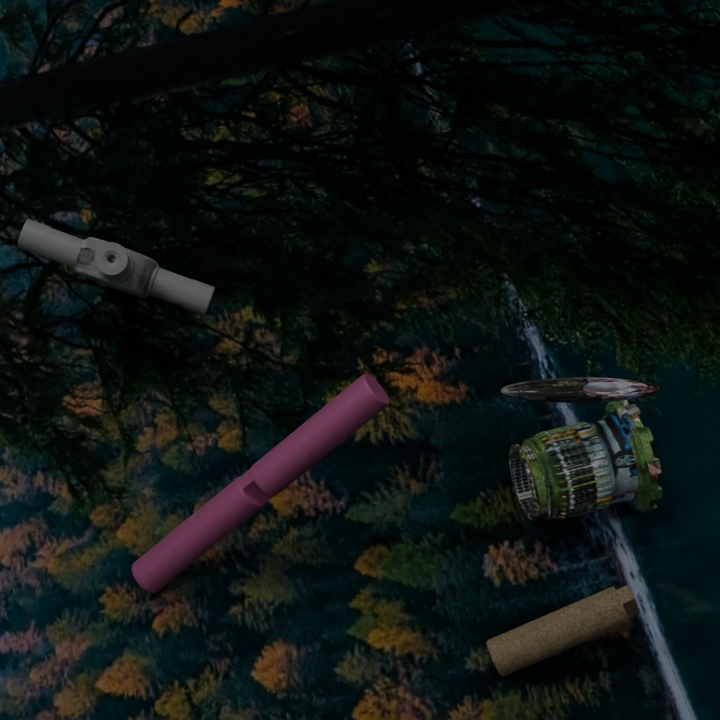}%
        \label{fig:subimage-m}%
    }\hspace{-0.5em}%
    \subfigure[Green lights.]{%
        \includegraphics[width=0.22\textwidth]{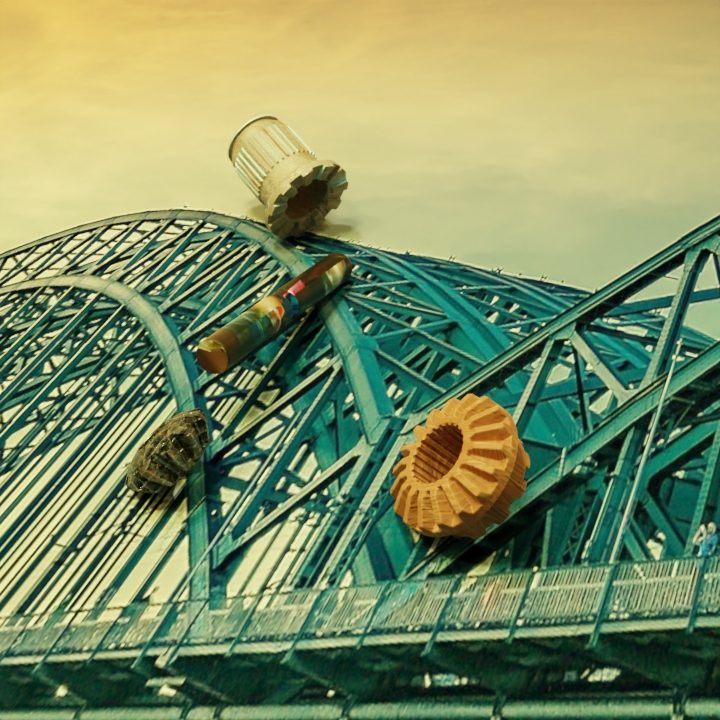}%
        \label{fig:subimage-n}%
    }\\[-0.2em]
    \subfigure[Red lights.]{%
        \includegraphics[width=0.22\textwidth]{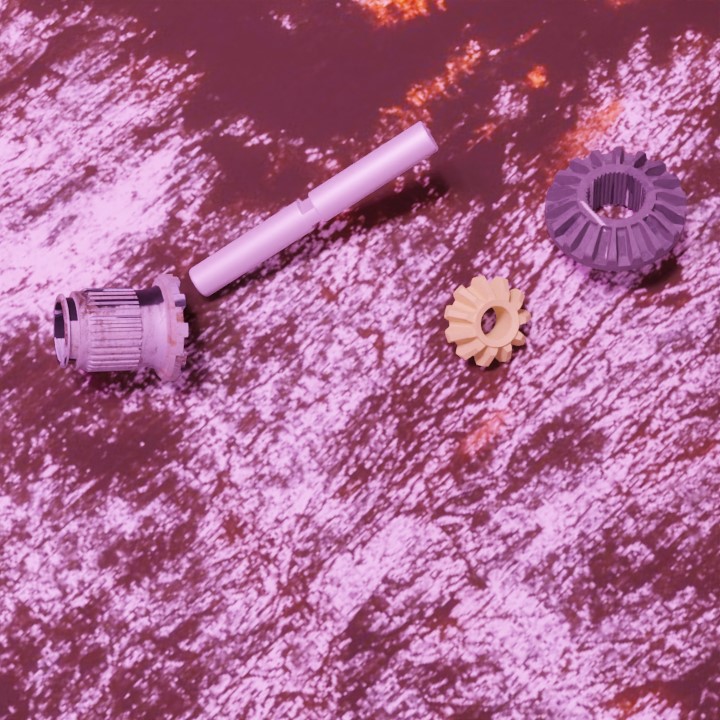}%
        \label{fig:subimage-o}%
    }\hspace{-0.5em}%
    \subfigure[Blue lights. ]{%
        \includegraphics[width=0.22\textwidth]{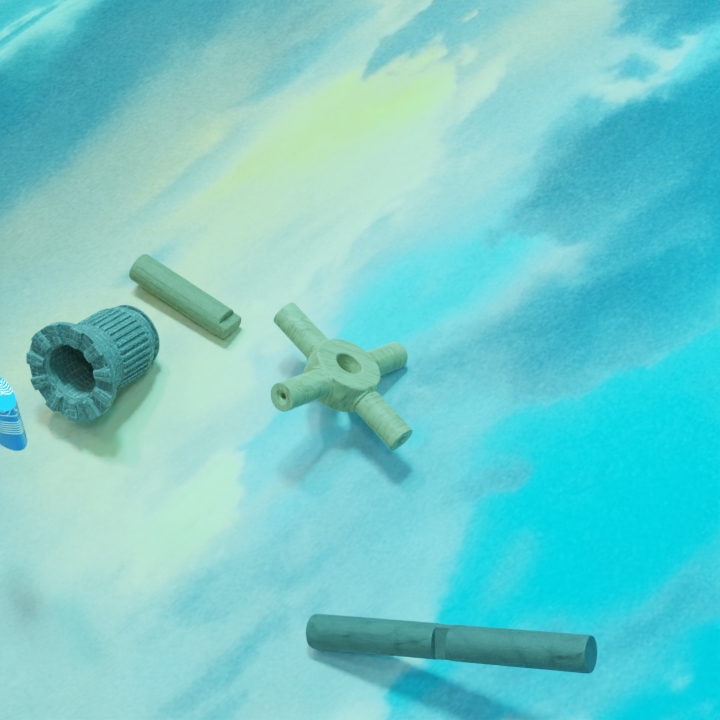}%
        \label{fig:subimage-p}%
    }\hspace{-0.5em}%
    \subfigure[Noise. ]{%
        \includegraphics[width=0.22\textwidth]{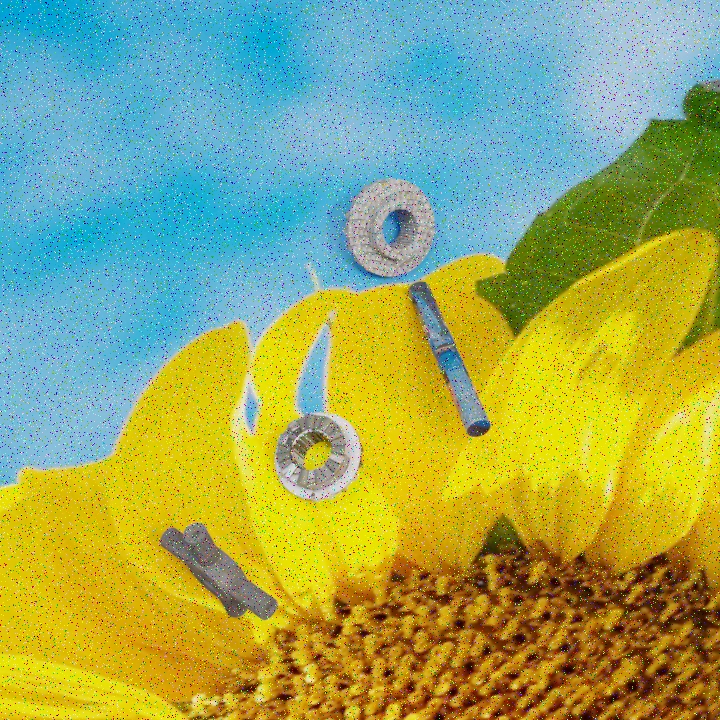}%
        \label{fig:subimage-q}%
    }\hspace{-0.5em}%
    \subfigure[Blur.]{%
        \includegraphics[width=0.22\textwidth]{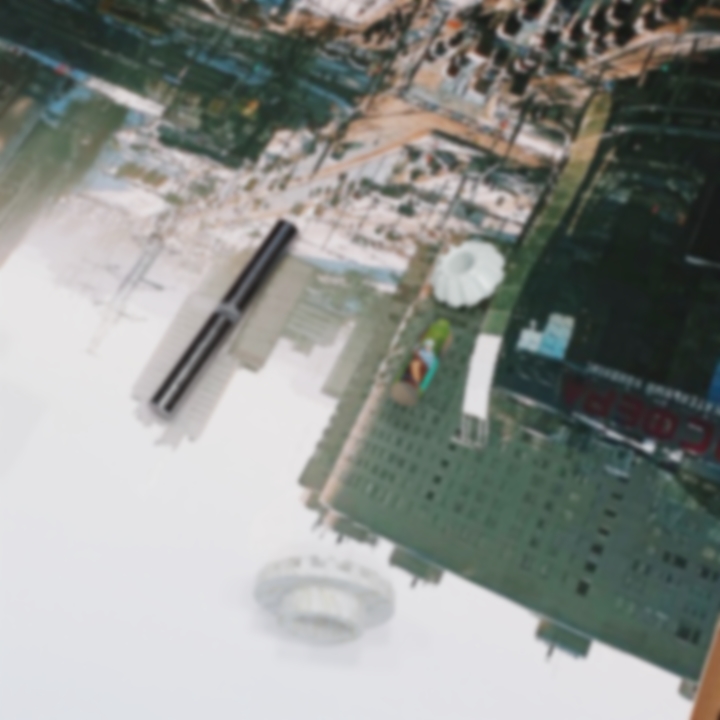}%
        \label{fig:subimage-r}%
    }
    \vspace{0.5em}
    \caption{Samples from the SIP15-OD dataset U3 illustrate the synthetic data generation pipeline. (a) Shows the 3D scene in Blender. (b) Displays annotations with bounding boxes and segmentation masks. Synthetic images demonstrate various domain randomization techniques applied to: (c-f) objects, (g) backgrounds (including random images and distractors), (h-k) camera settings, (l-p) illumination, and (q-r) post-processing.}
    \label{fig:DR_pipeline}
    \vspace{-0.02\textwidth}
\end{figure*}

\begin{figure*}[t]
  \centering
  \subfigure[]{\includegraphics[width=0.245\textwidth]{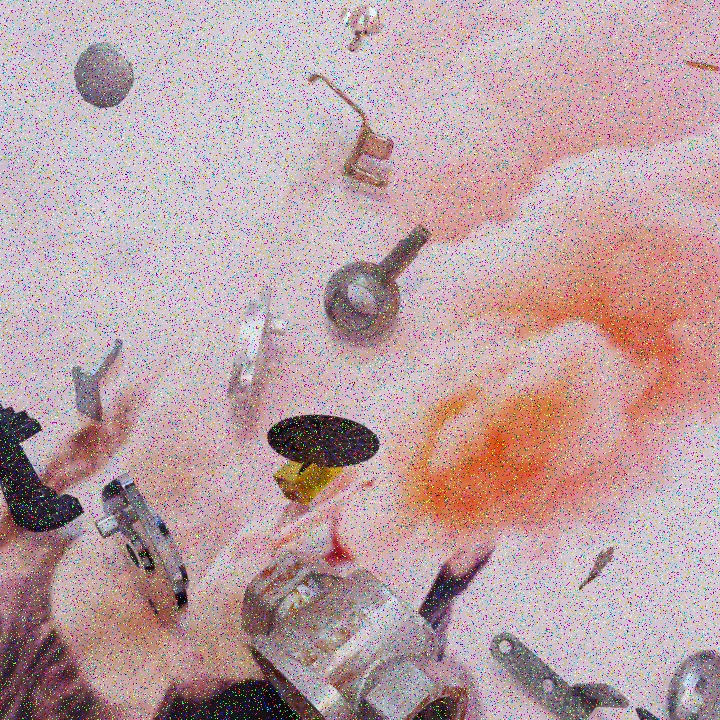}}\hspace{-0.5em}%
  \subfigure[]{\includegraphics[width=0.245\textwidth]{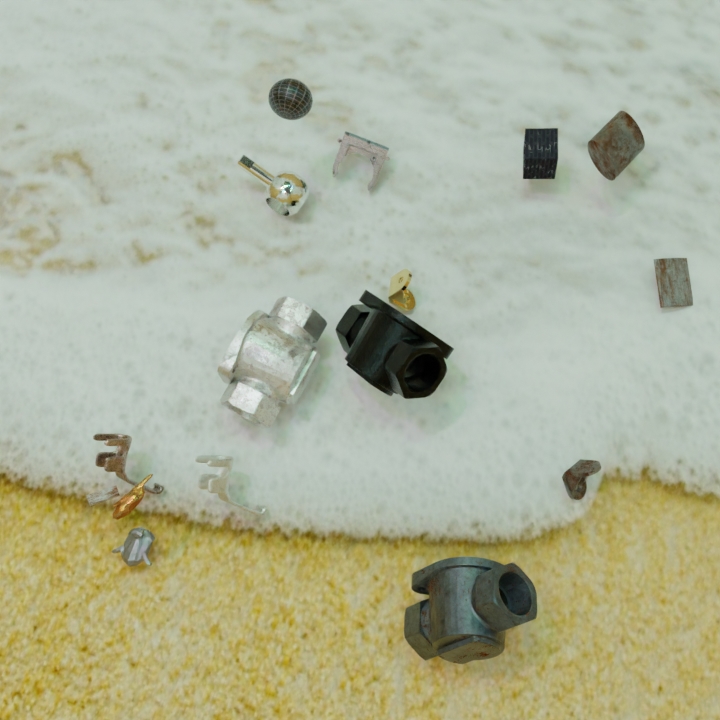}}\hspace{-0.5em}%
  \subfigure[]{\includegraphics[width=0.245\textwidth]{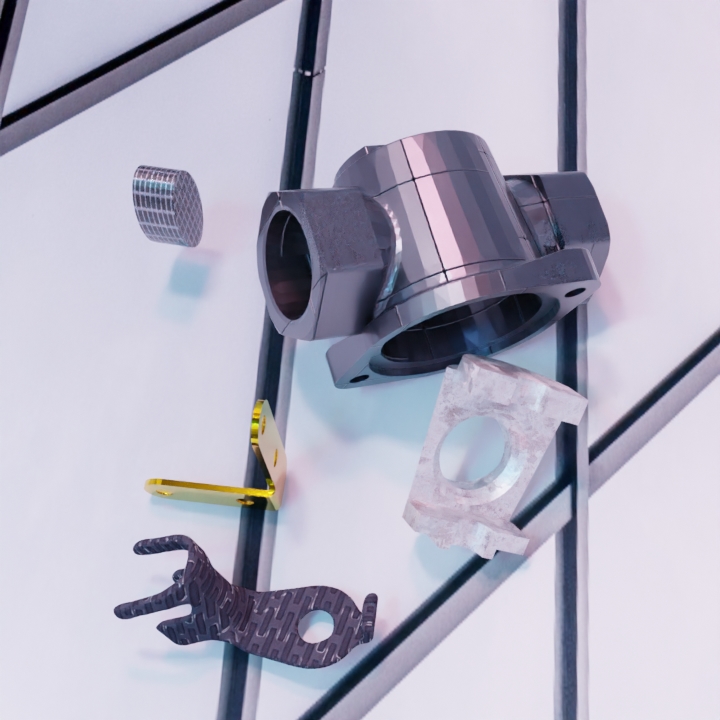}}\hspace{-0.5em}%
  \subfigure[]{\includegraphics[width=0.245\textwidth]{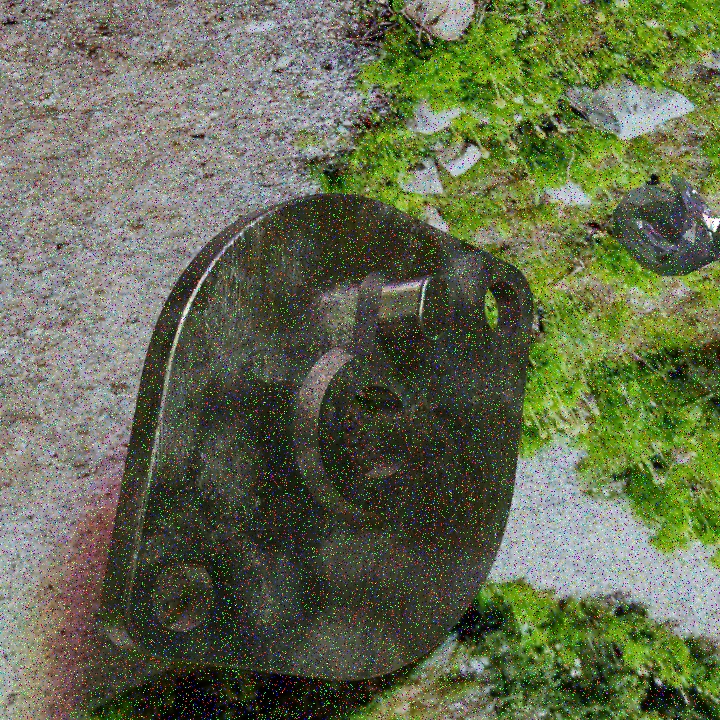}}\\[-0.2em]
  \subfigure[\cite{horvath2022object}]{\includegraphics[width=0.245\textwidth]{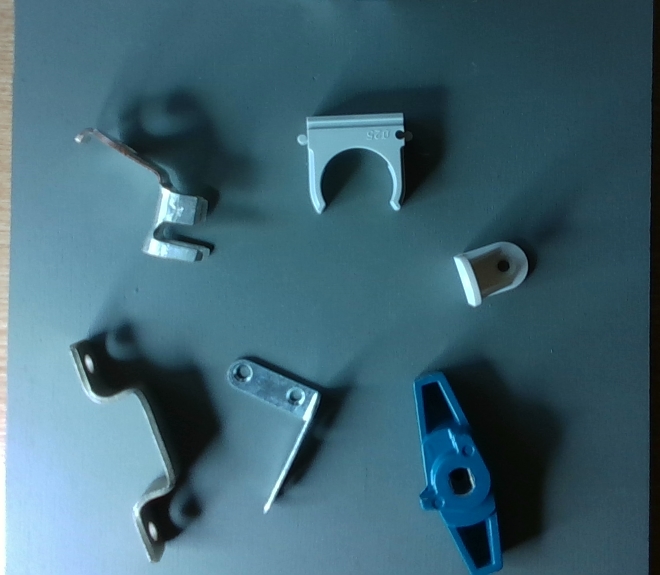}}\hspace{-0.5em}%
  \subfigure[\cite{horvath2022object}]{\includegraphics[width=0.245\textwidth]{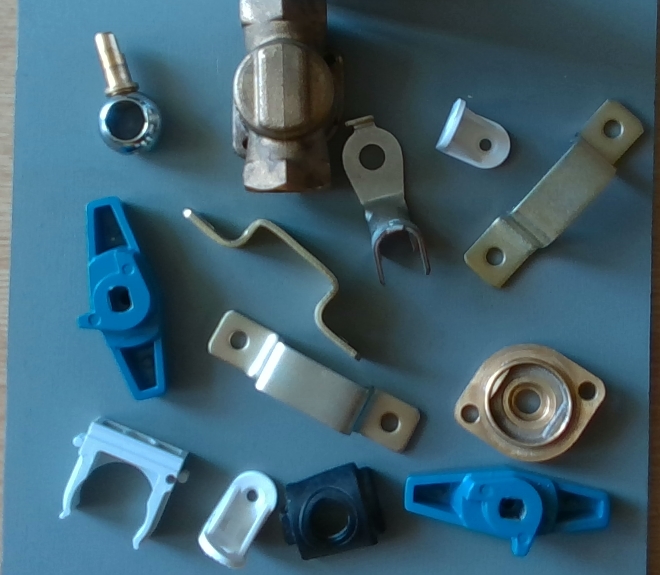}}\hspace{-0.5em}%
  \subfigure[\cite{horvath2022object}]{\includegraphics[width=0.245\textwidth]{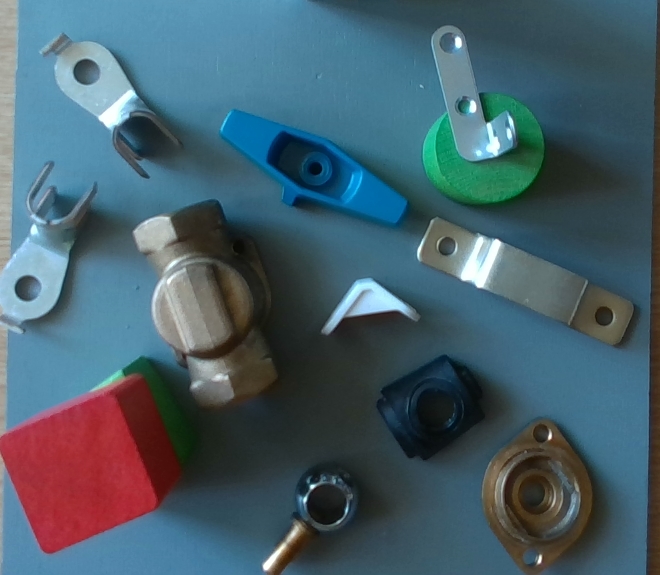}}\hspace{-0.5em}%
  \subfigure[\cite{horvath2022object}]{\includegraphics[width=0.245\textwidth]{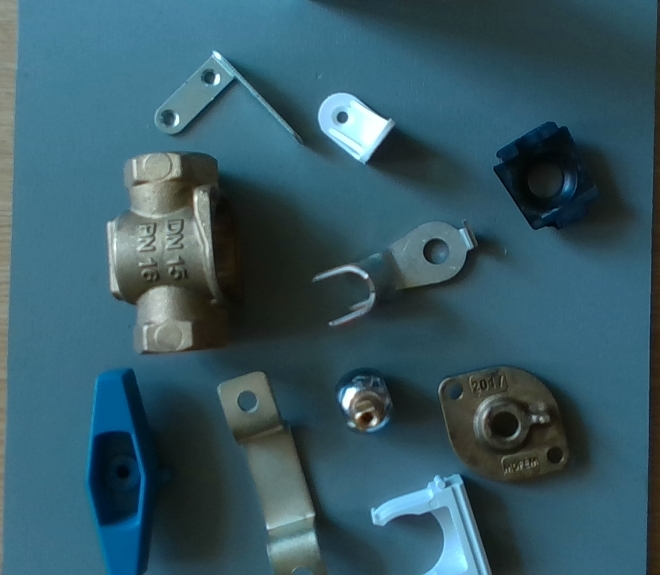}}
  \caption{Sample images from the robotic dataset. (a-d) Synthetic images generated by our pipeline. (e-h) Real images provided by
Horvath et al. \cite{horvath2022object}}
  \label{fig:dataset_rb}
  \vspace{-0.03\textwidth}
\end{figure*}

\begin{figure*}[t]
  \centering
  \subfigure[]{\includegraphics[width=0.245\textwidth]{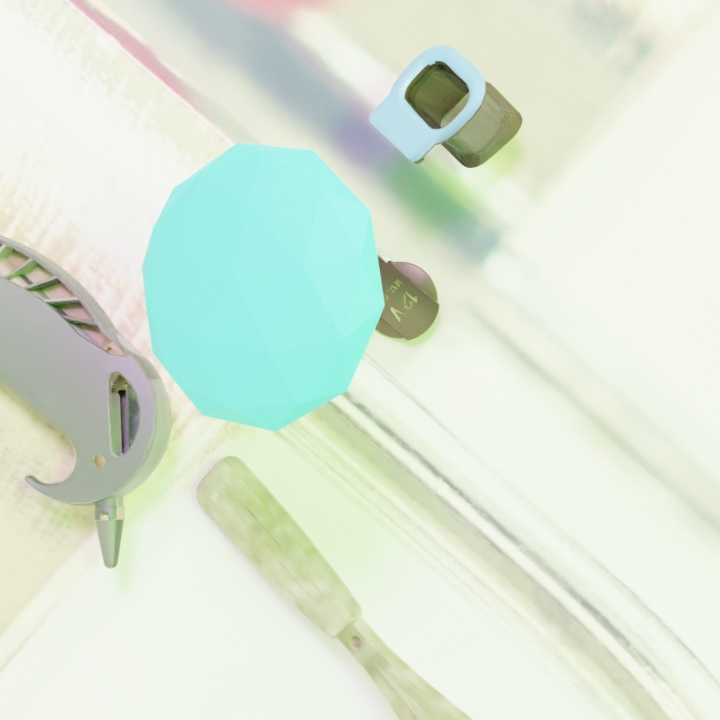}}\hspace{-0.5em}%
  \subfigure[]{\includegraphics[width=0.245\textwidth]{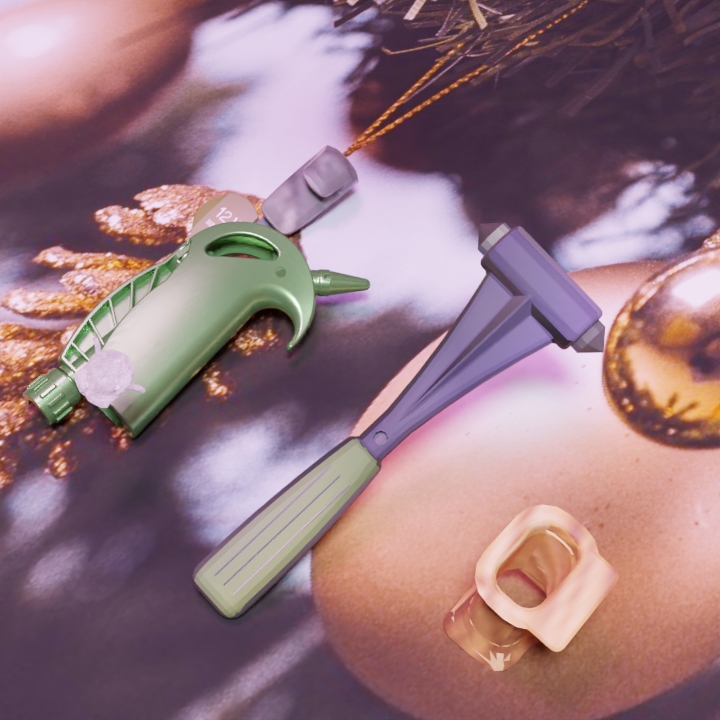}}\hspace{-0.5em}%
  \subfigure[]{\includegraphics[width=0.245\textwidth]{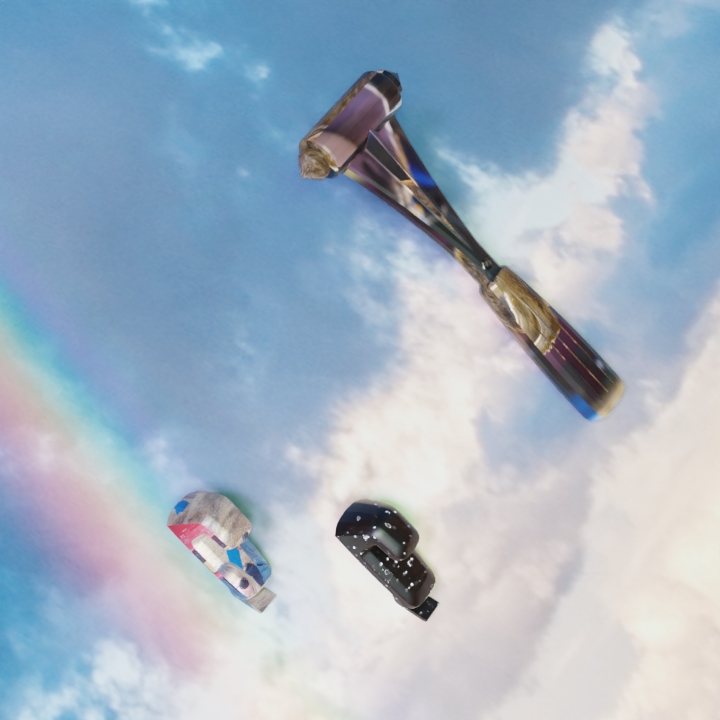}}\hspace{-0.5em}%
  \subfigure[]{\includegraphics[width=0.245\textwidth]{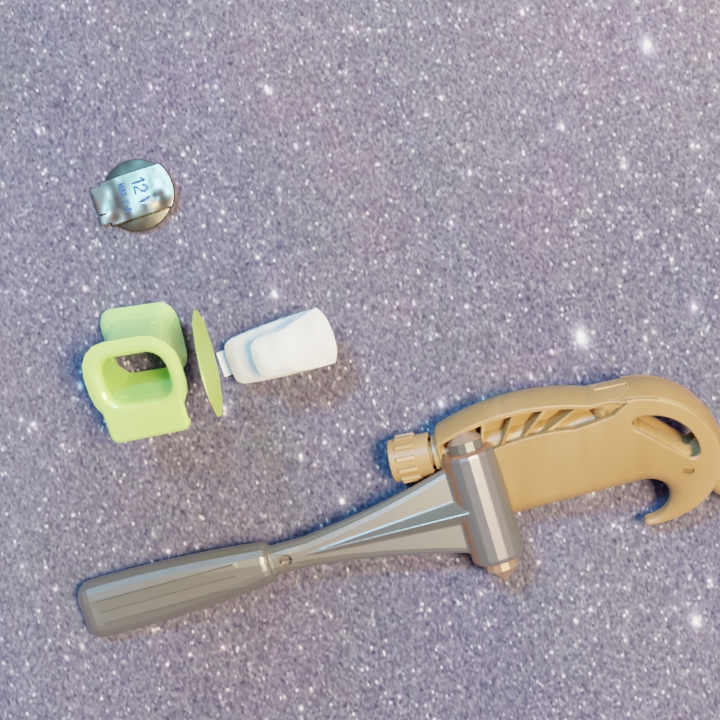}}\\[-0.2em]
  \subfigure[]{\includegraphics[width=0.245\textwidth]{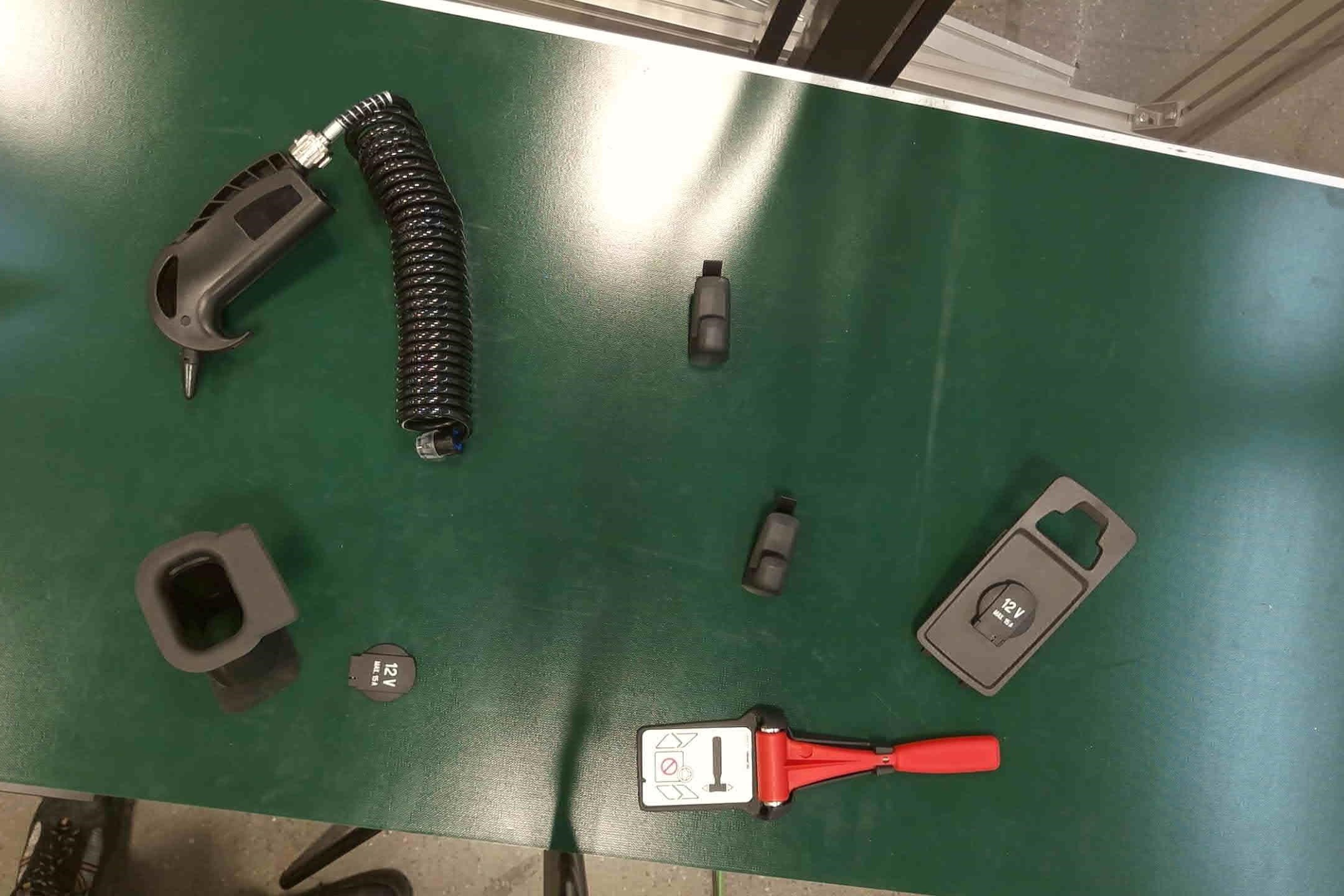}}\hspace{-0.5em}%
  \subfigure[]{\includegraphics[width=0.245\textwidth]{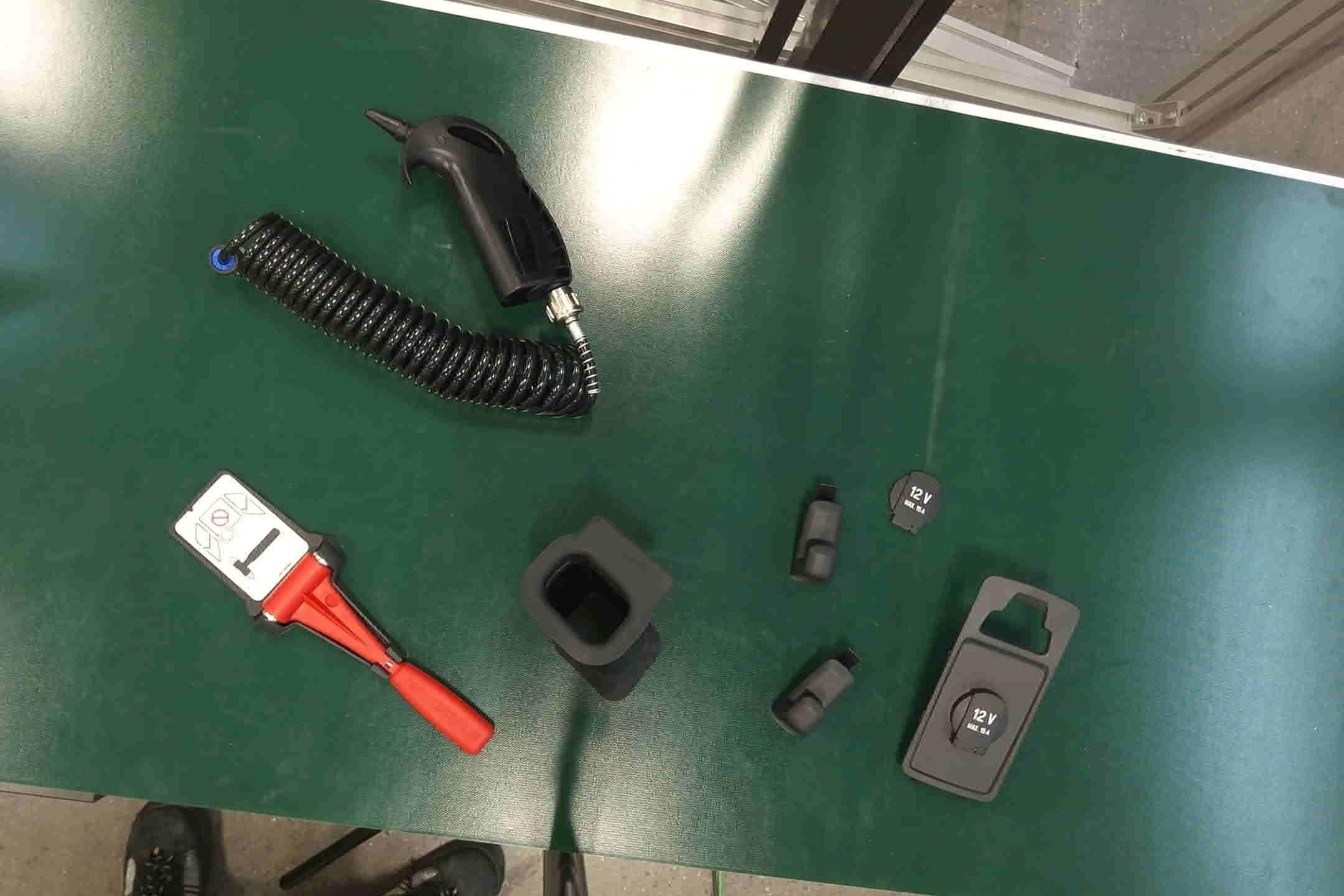}}\hspace{-0.5em}%
  \subfigure[]{\includegraphics[width=0.245\textwidth]{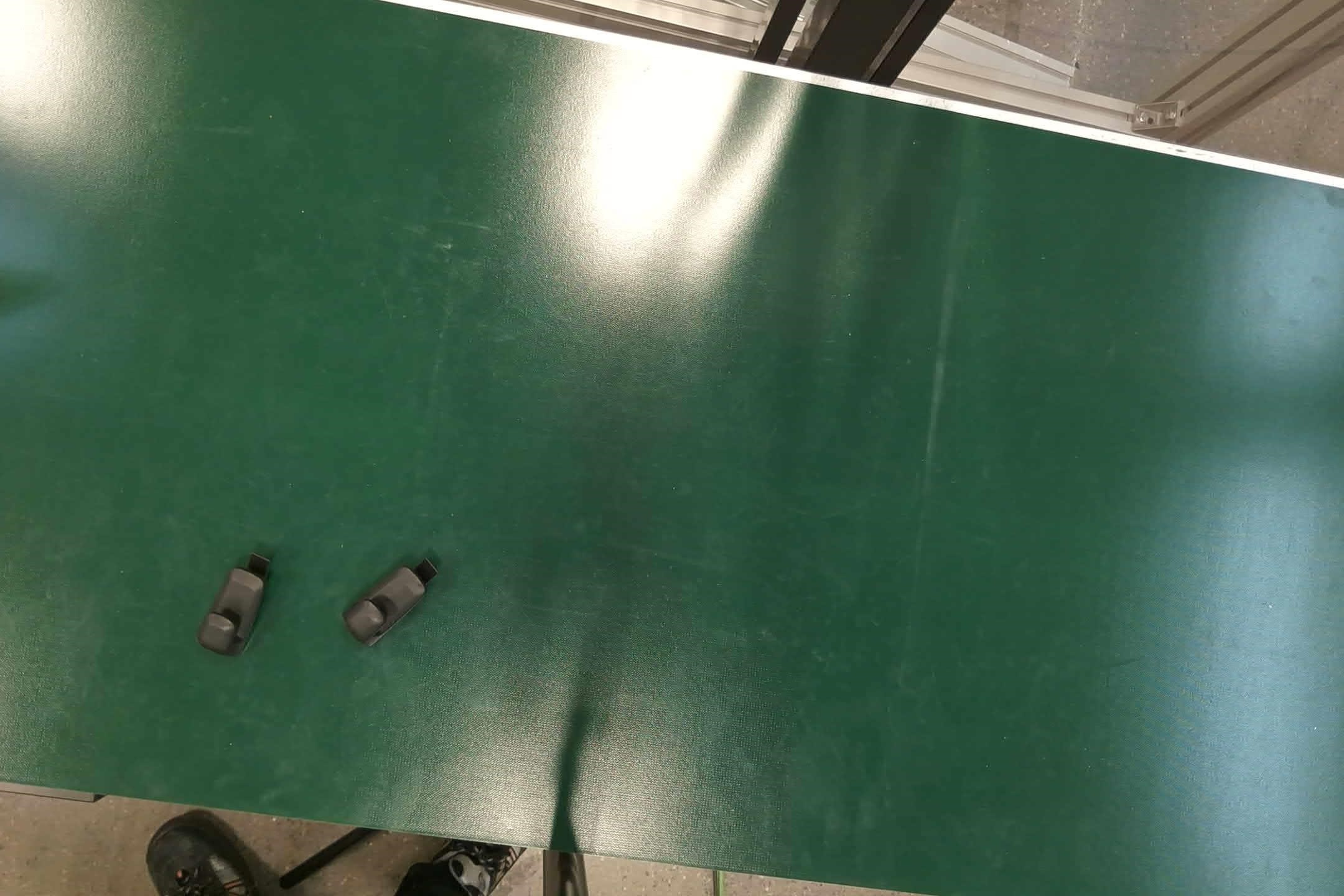}}\hspace{-0.5em}%
  \subfigure[]{\includegraphics[width=0.245\textwidth]{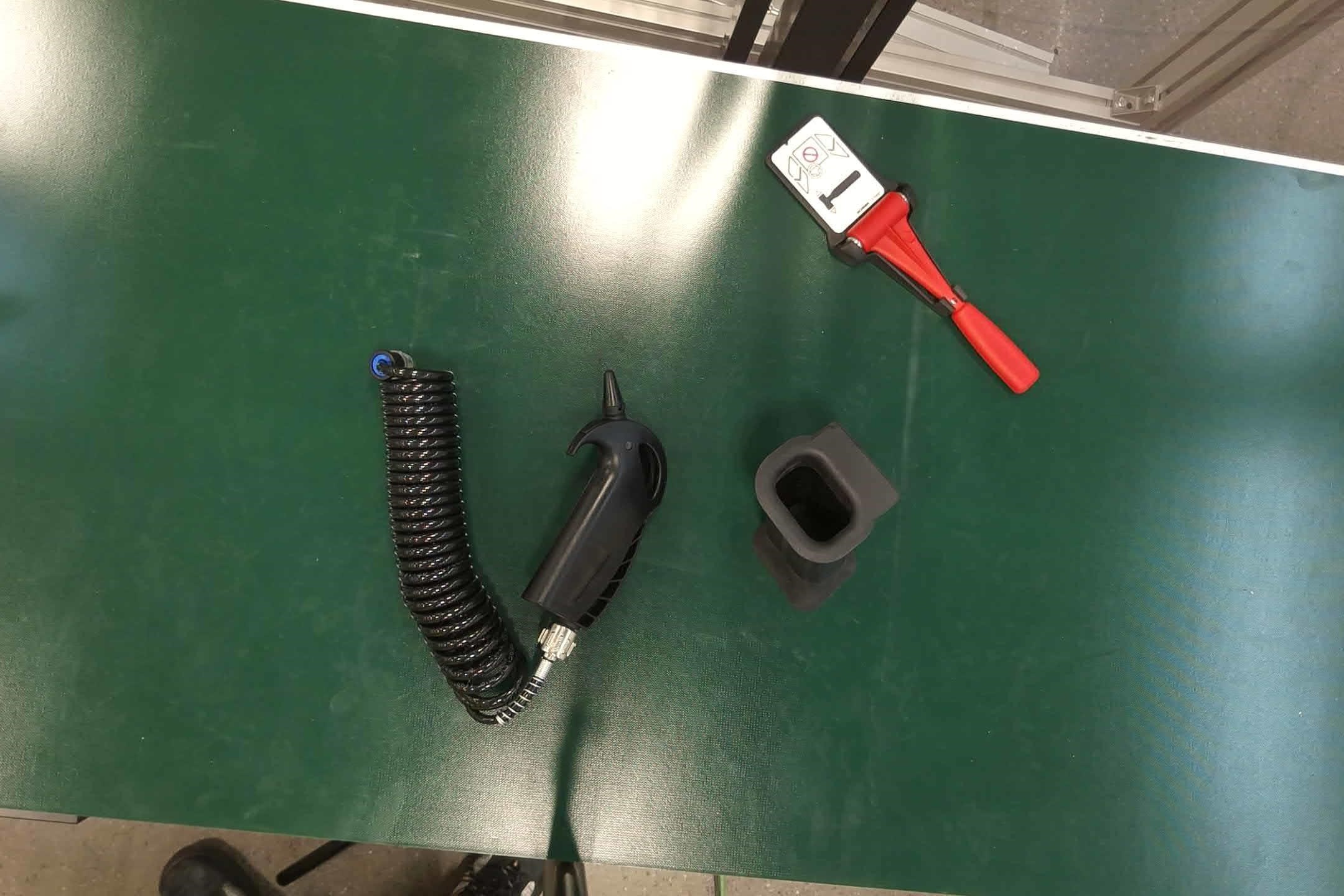}}\\[-0.2em]
  \subfigure[]{\includegraphics[width=0.245\textwidth]{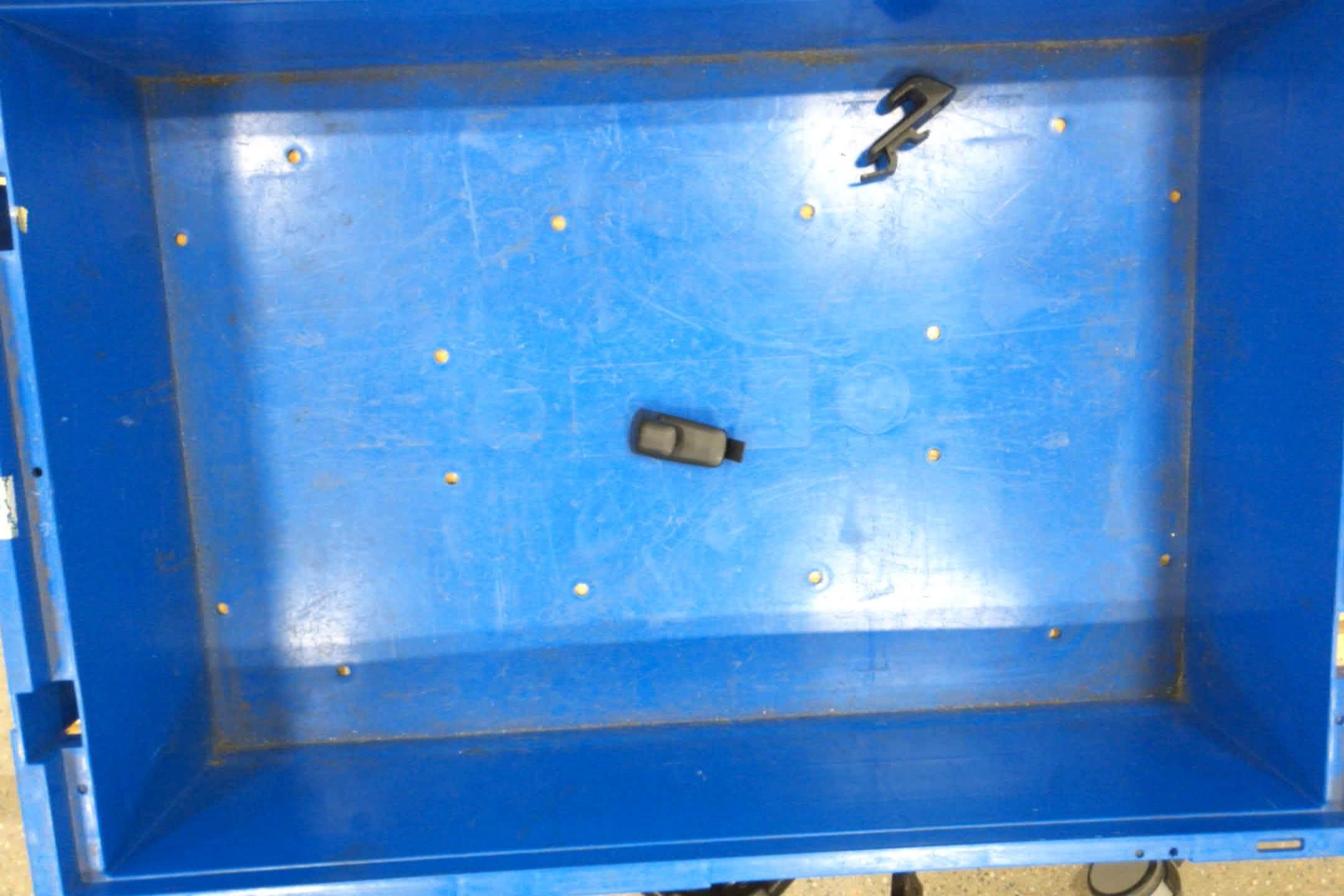}}\hspace{-0.5em}%
  \subfigure[]{\includegraphics[width=0.245\textwidth]{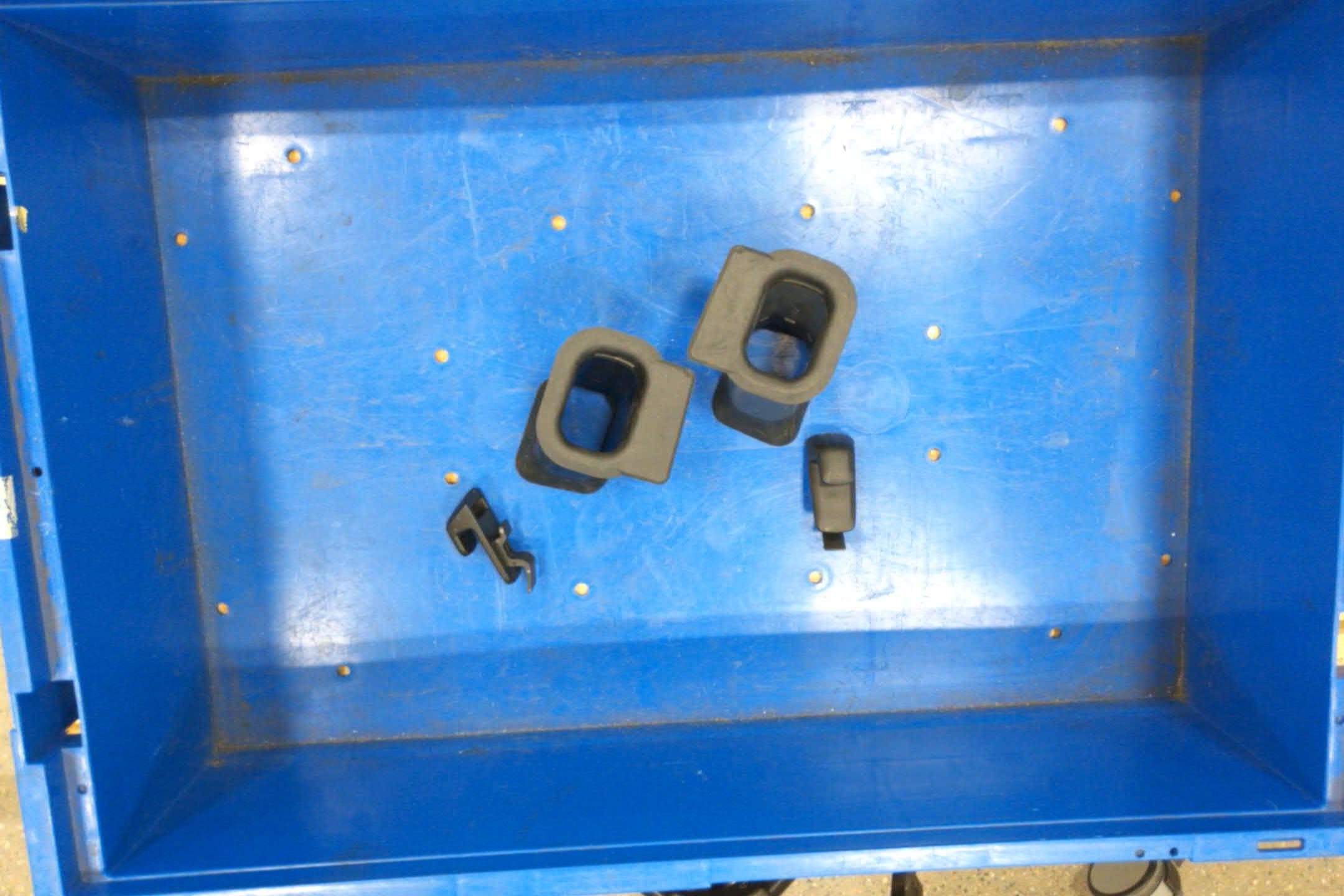}}\hspace{-0.5em}%
  \subfigure[]{\includegraphics[width=0.245\textwidth]{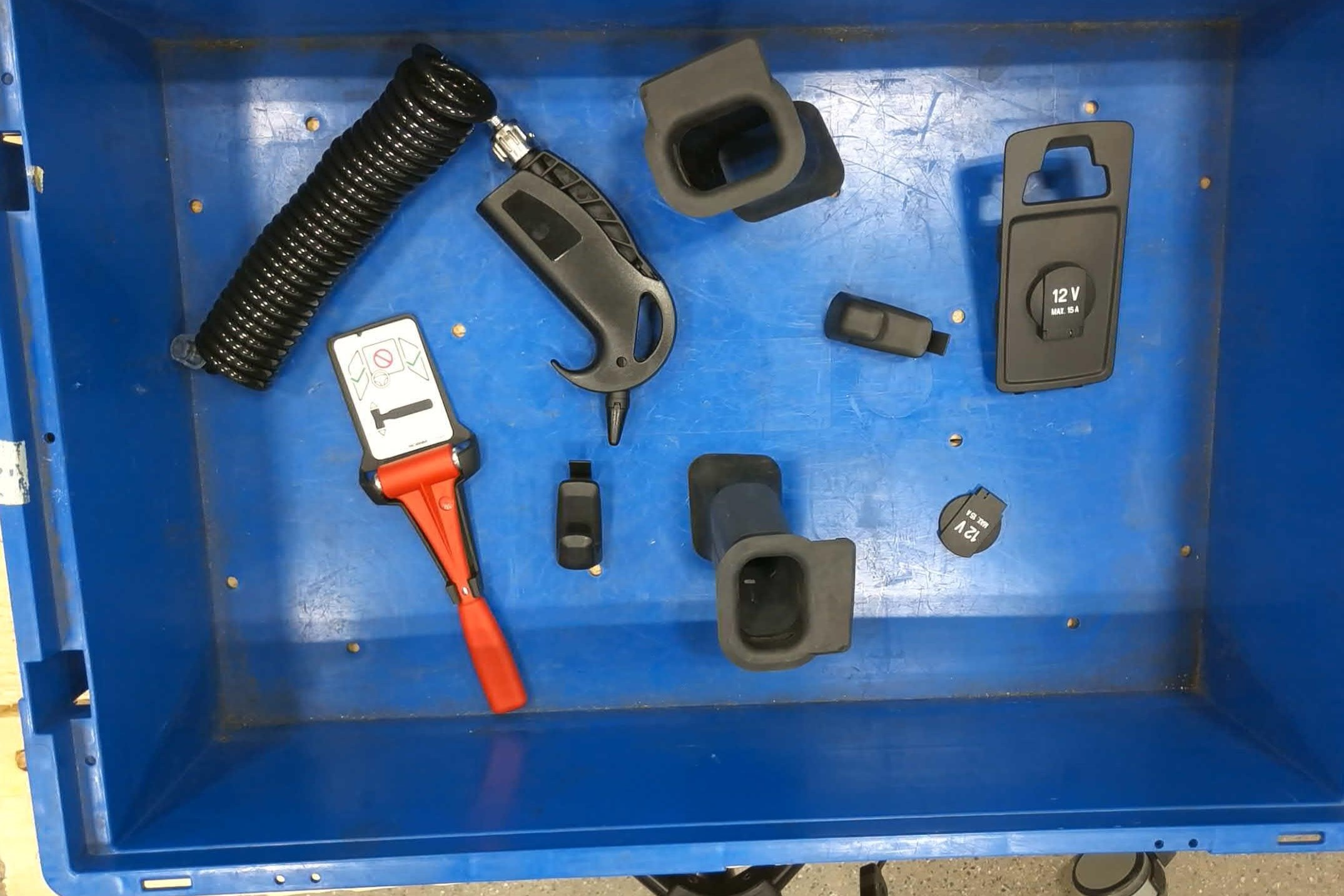}}\hspace{-0.5em}%
  \subfigure[]{\includegraphics[width=0.245\textwidth]{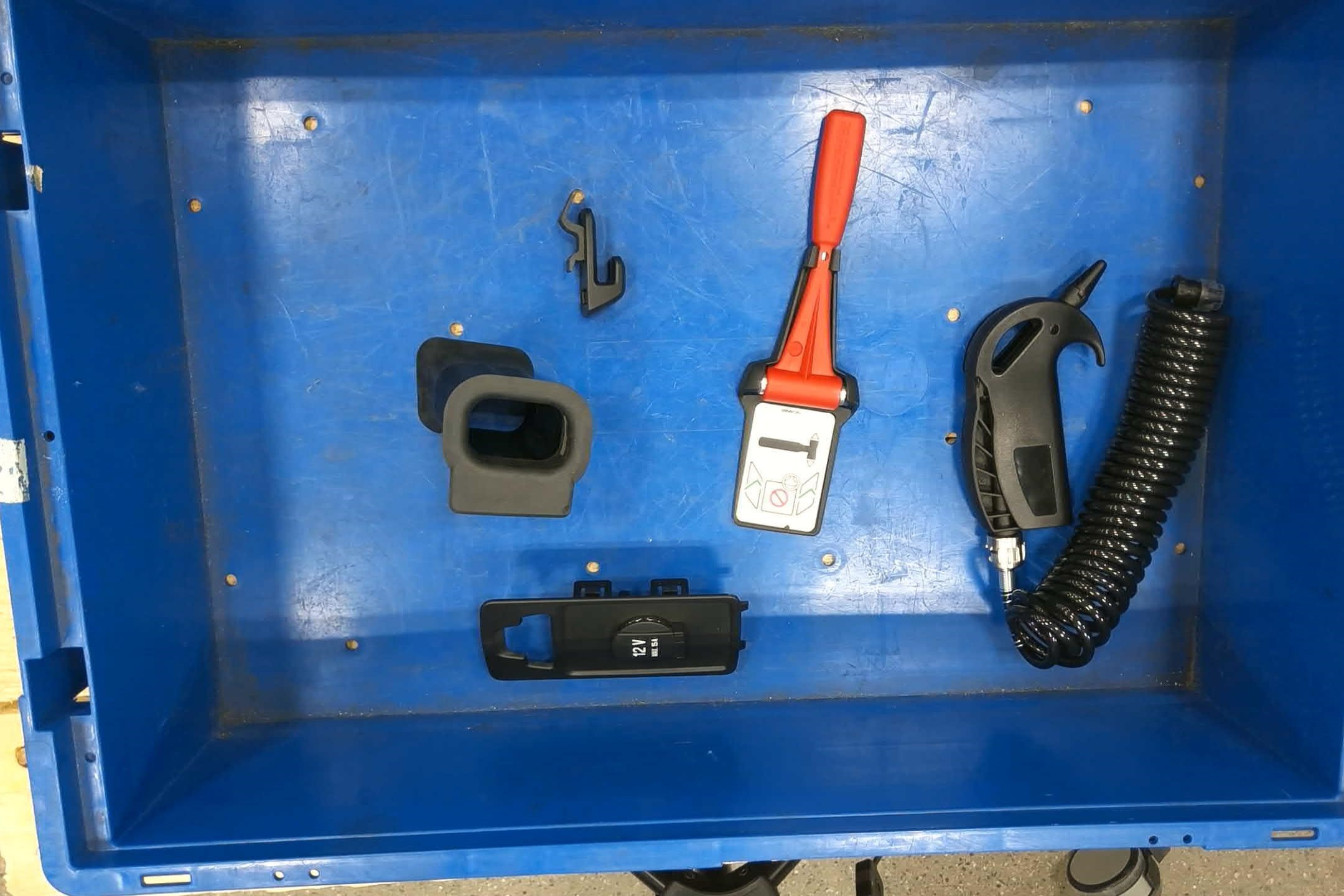}}\\[-0.2em]
  \caption{Samples images from the SIP15-OD dataset, U1. (a-d) Synthetic images. (e-h) Real images from S1. (i-l) Real images from S2. Note: real images are cropped for layout, maintaining original aspect ratios.}
  \label{fig:dataset_us1}
  \vspace{-0.03\textwidth}
\end{figure*}

\begin{figure*}[t]
  \centering
  \subfigure[]{\includegraphics[width=0.245\textwidth]{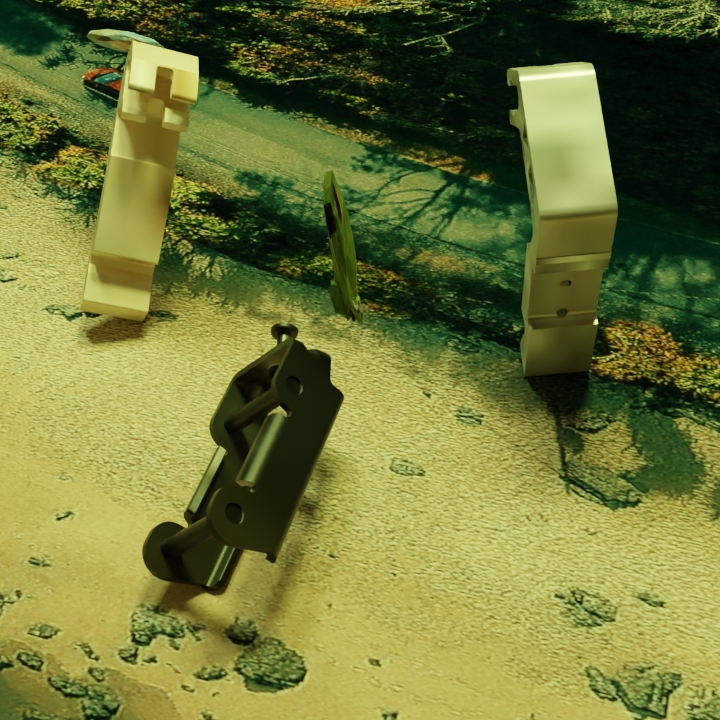}}\hspace{-0.5em}%
  \subfigure[]{\includegraphics[width=0.245\textwidth]{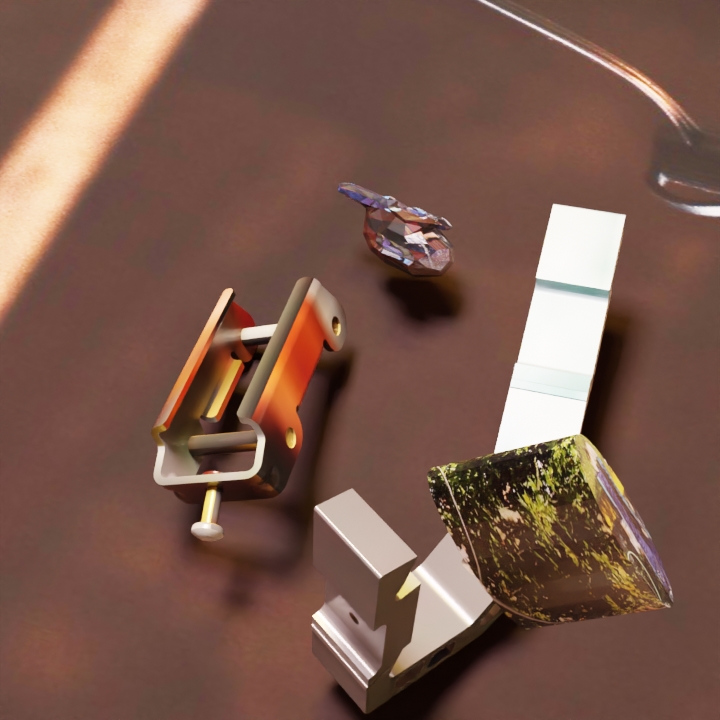}}\hspace{-0.5em}%
  \subfigure[]{\includegraphics[width=0.245\textwidth]{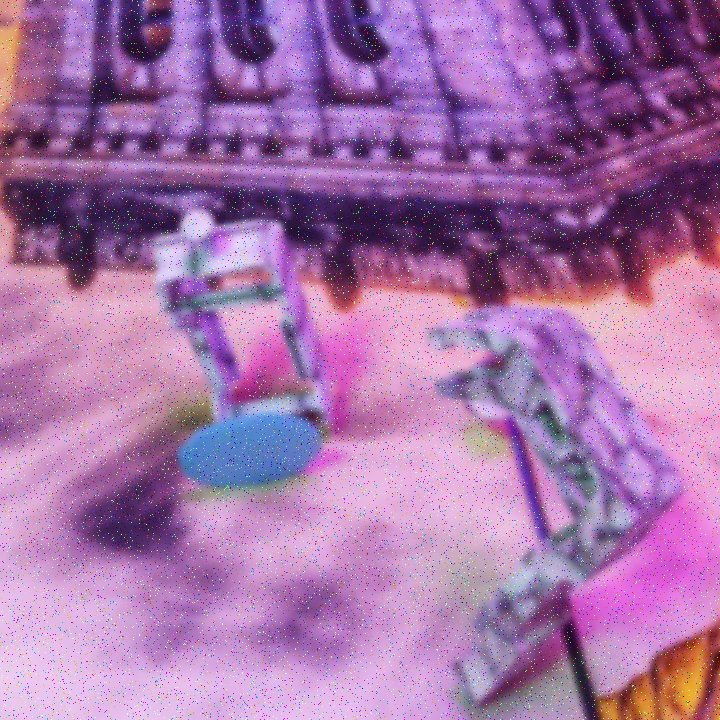}}\hspace{-0.5em}%
  \subfigure[]{\includegraphics[width=0.245\textwidth]{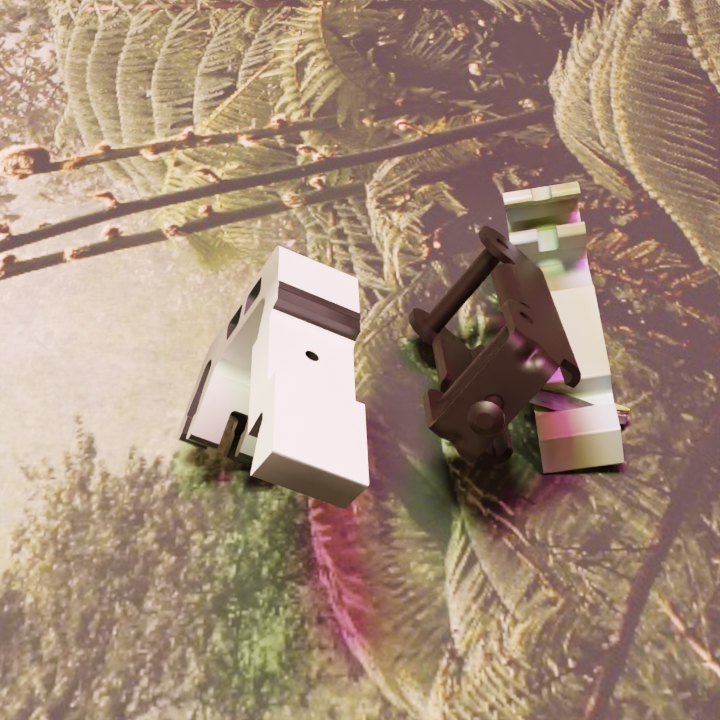}}\\[-0.2em]
  \subfigure[]{\includegraphics[width=0.245\textwidth]{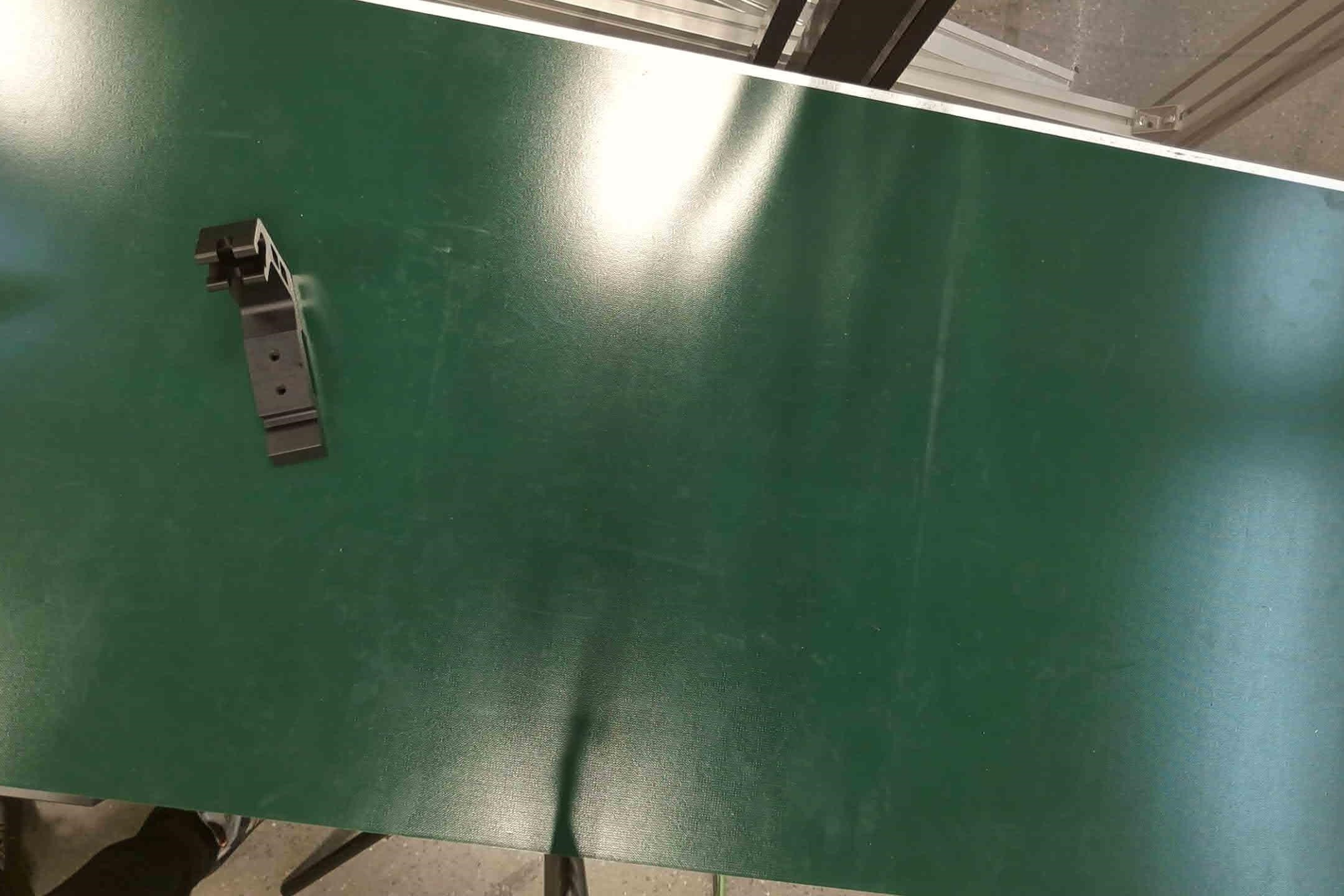}}\hspace{-0.5em}%
  \subfigure[]{\includegraphics[width=0.245\textwidth]{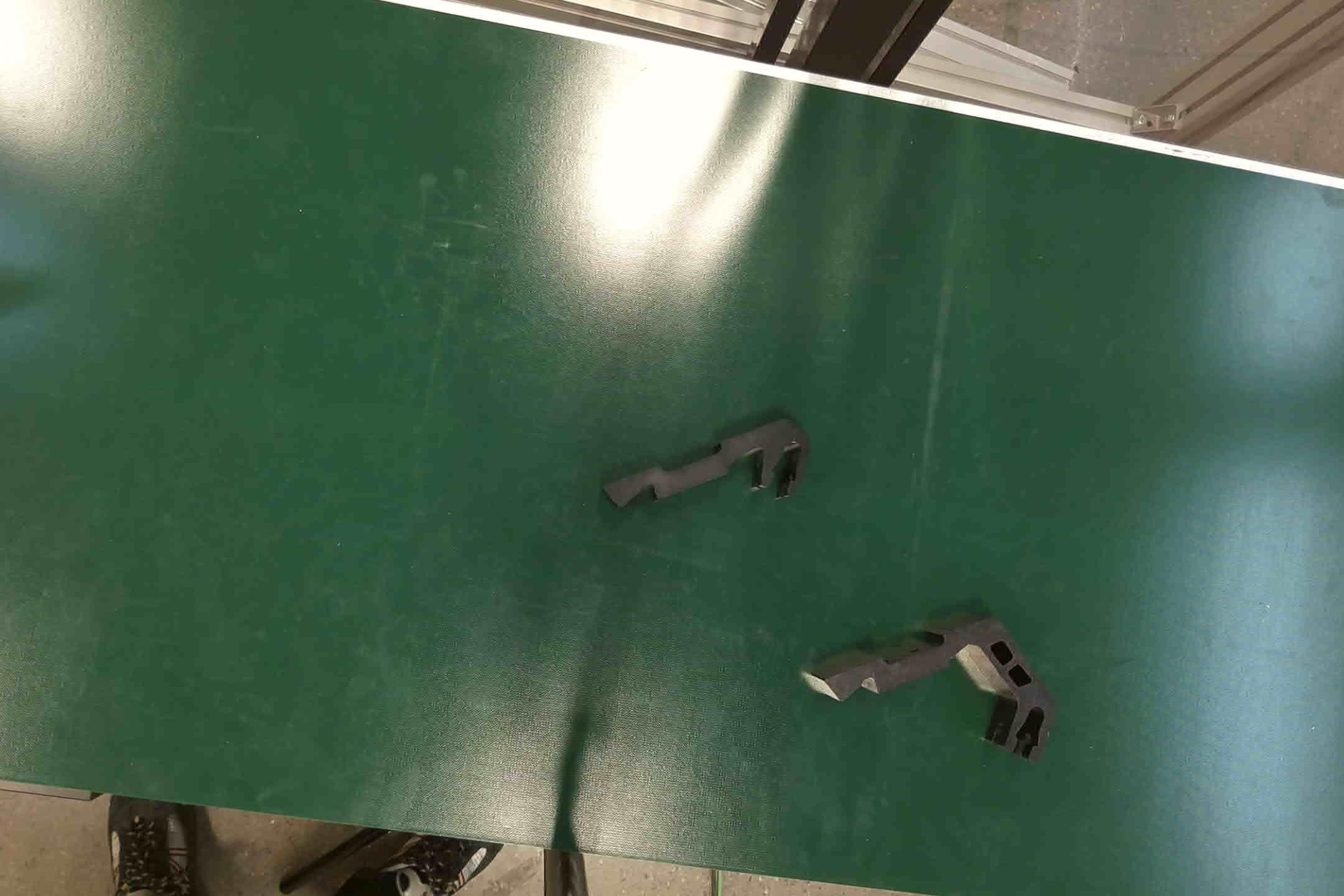}}\hspace{-0.5em}%
  \subfigure[]{\includegraphics[width=0.245\textwidth]{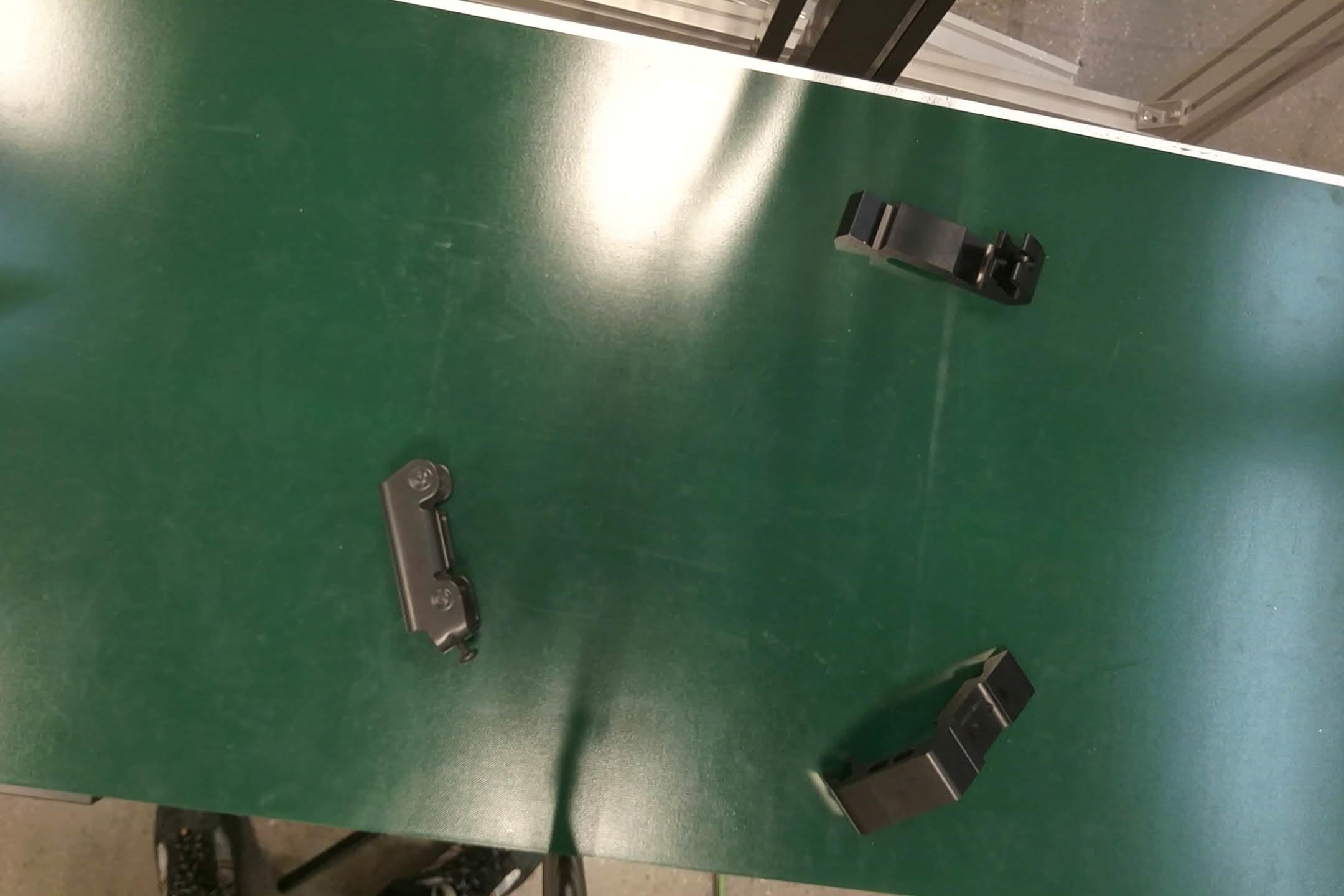}}\hspace{-0.5em}%
  \subfigure[]{\includegraphics[width=0.245\textwidth]{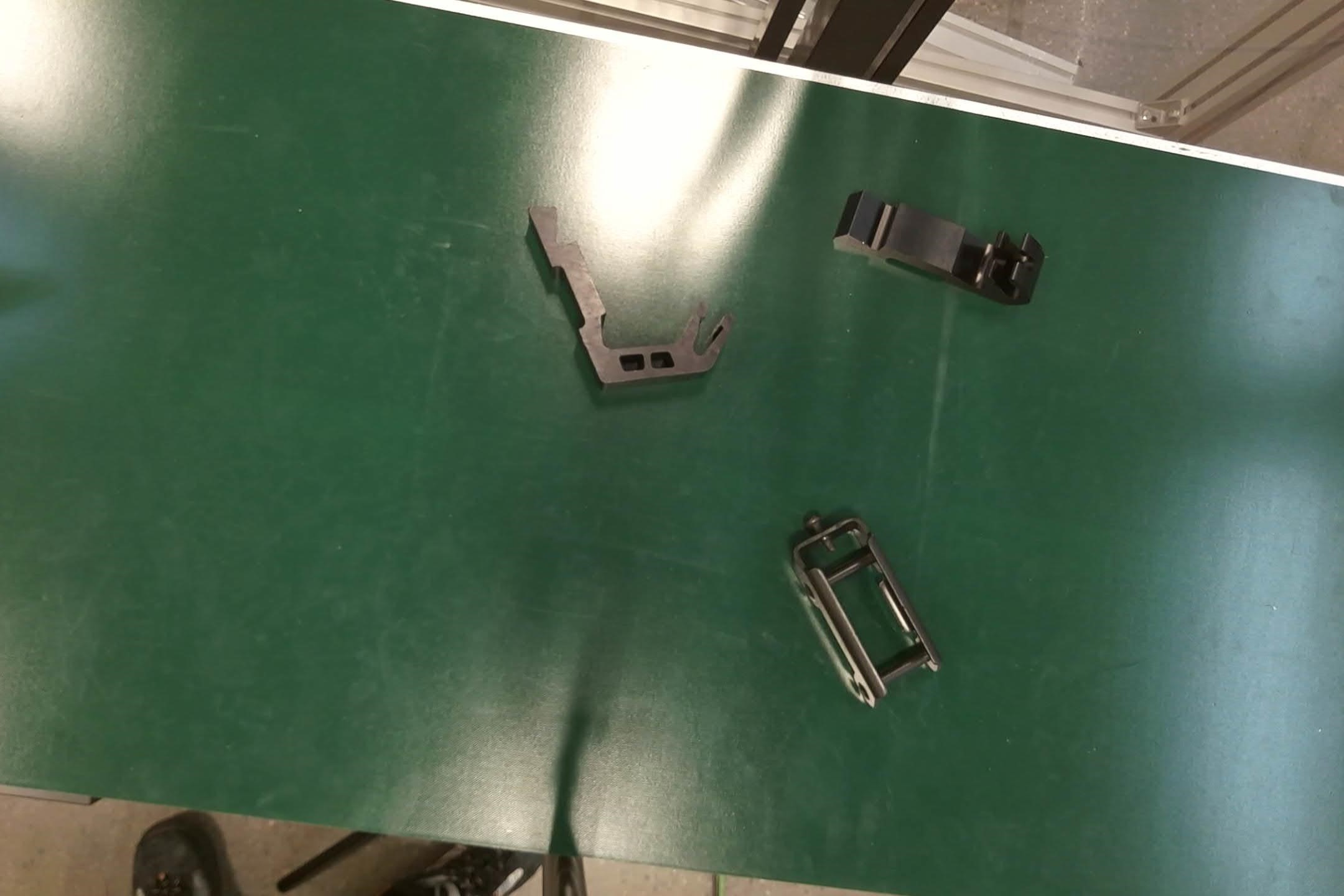}}\\[-0.2em]
  \subfigure[]{\includegraphics[width=0.245\textwidth]{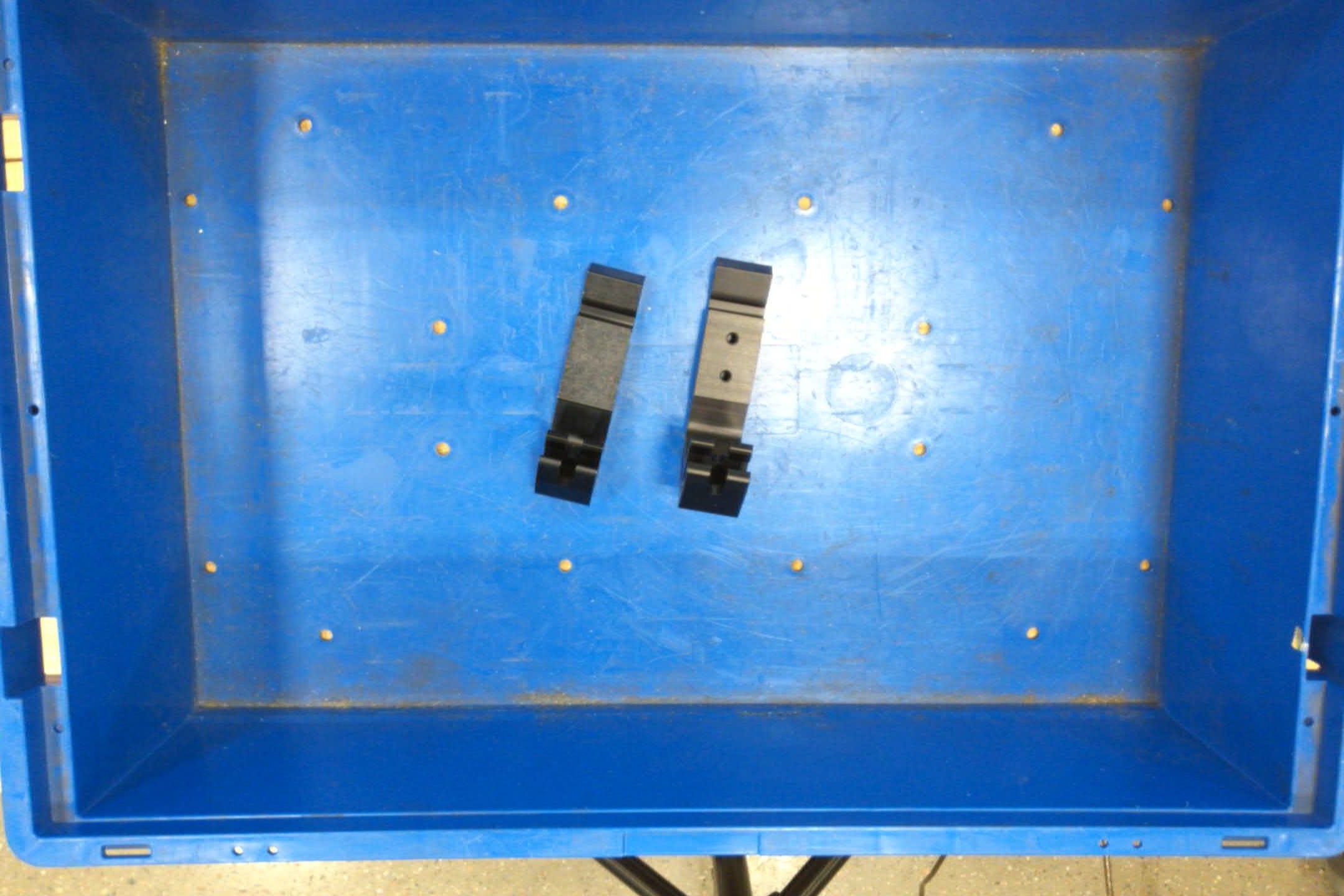}}\hspace{-0.5em}%
  \subfigure[]{\includegraphics[width=0.245\textwidth]{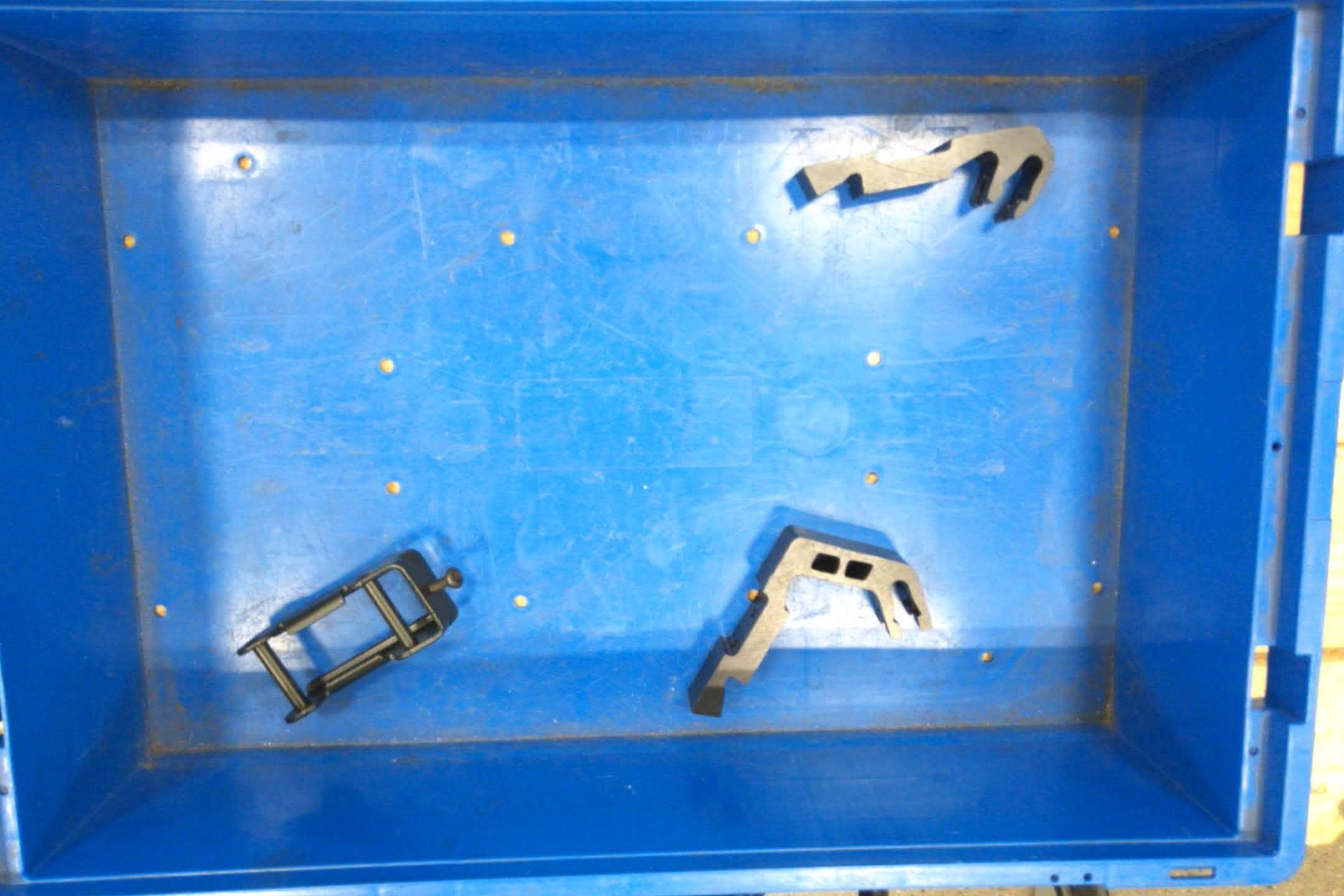}}\hspace{-0.5em}%
  \subfigure[]{\includegraphics[width=0.245\textwidth]{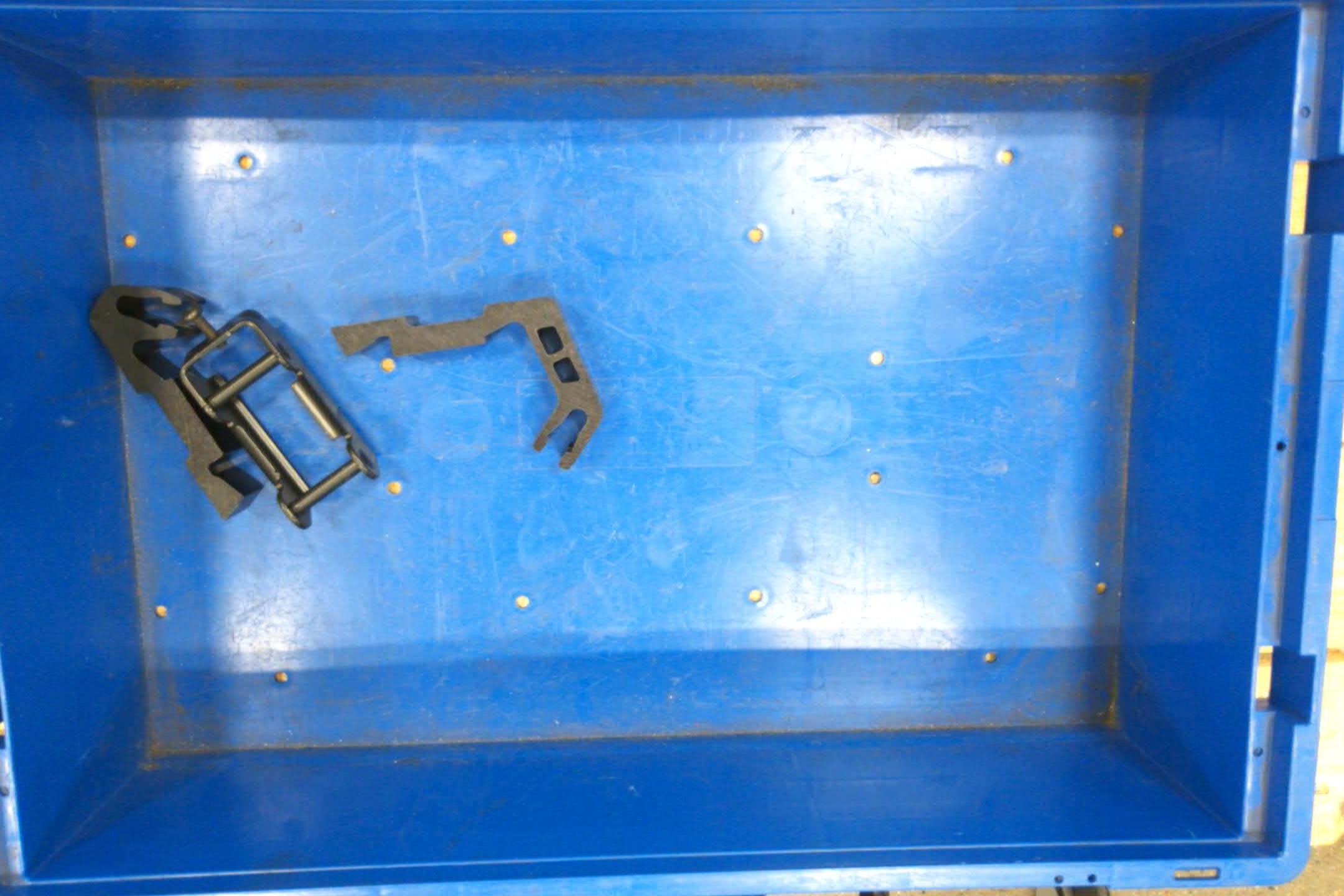}}\hspace{-0.5em}%
  \subfigure[]{\includegraphics[width=0.245\textwidth]{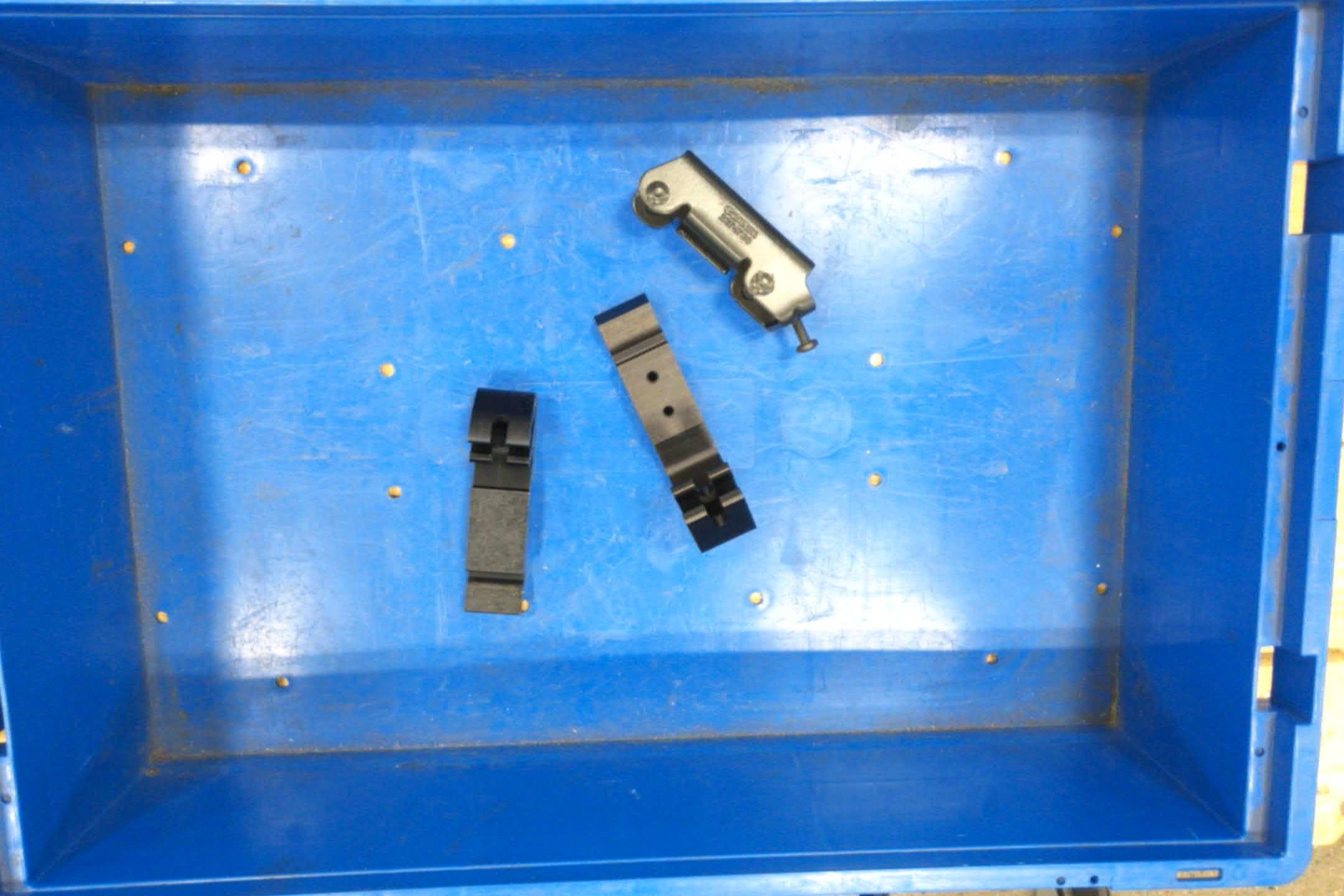}}
  \caption{Sample images from the SIP15-OD dataset, U2. (a-d) Synthetic images. (e-h) Real images from S1. (i-l) Real images from S2. Note: real images are cropped for layout, maintaining original aspect ratios.}
  \label{fig:dataset_us2}
  \vspace{-0.03\textwidth}
\end{figure*}

\begin{figure*}[t]
  \centering
  \subfigure[]{\includegraphics[width=0.245\textwidth]{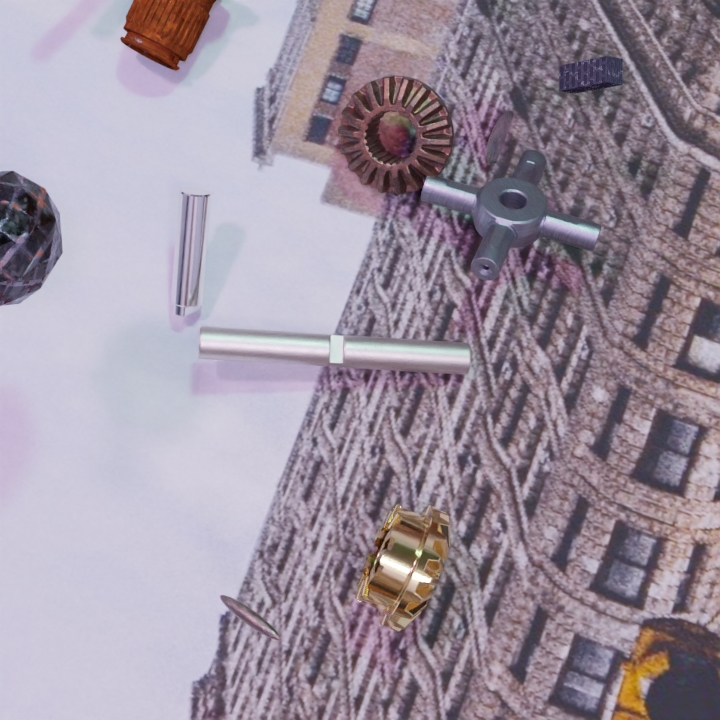}}\hspace{-0.5em}%
  \subfigure[]{\includegraphics[width=0.245\textwidth]{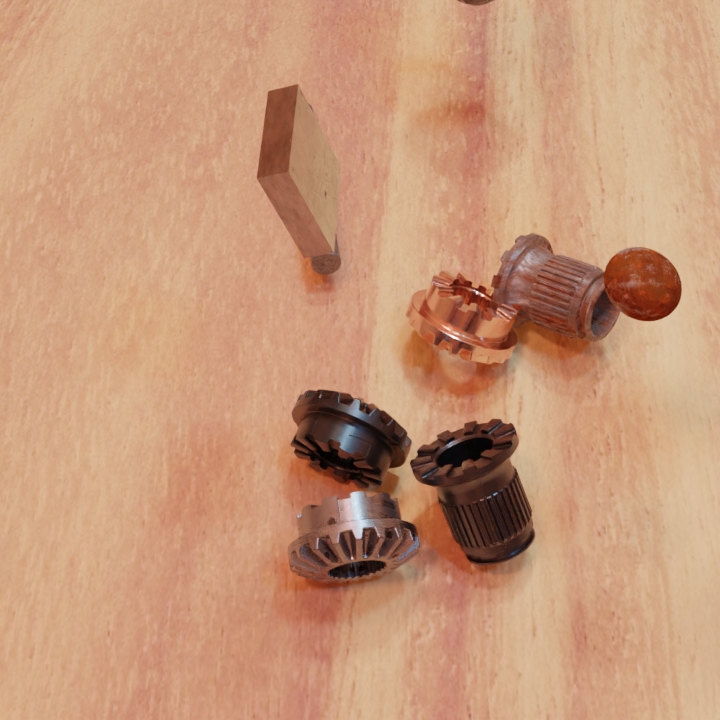}}\hspace{-0.5em}%
  \subfigure[]{\includegraphics[width=0.245\textwidth]{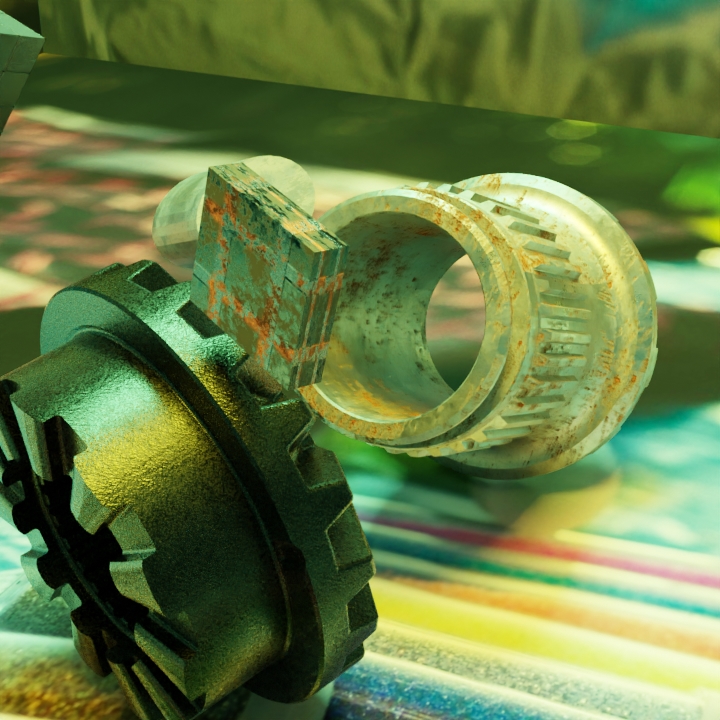}}\hspace{-0.5em}%
  \subfigure[]{\includegraphics[width=0.245\textwidth]{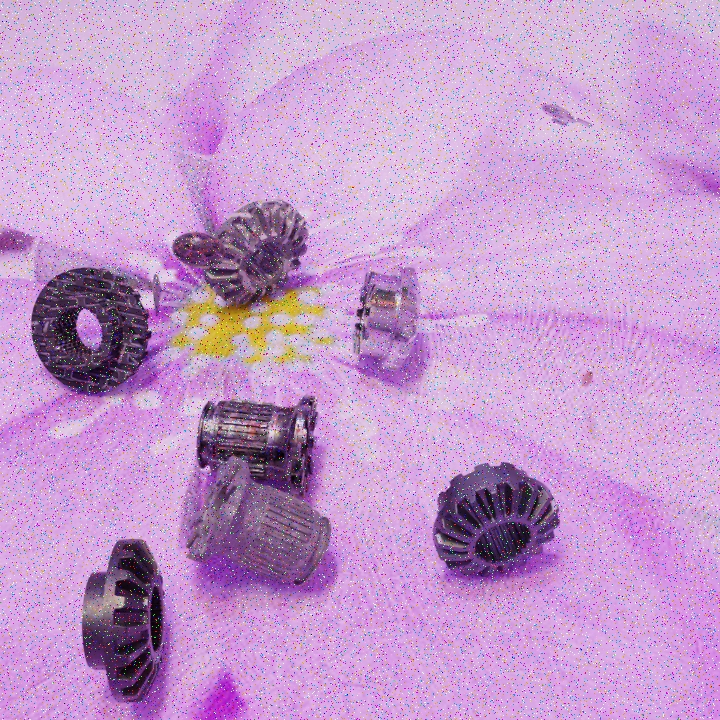}}\\[-0.2em]
  \subfigure[]{\includegraphics[width=0.245\textwidth]{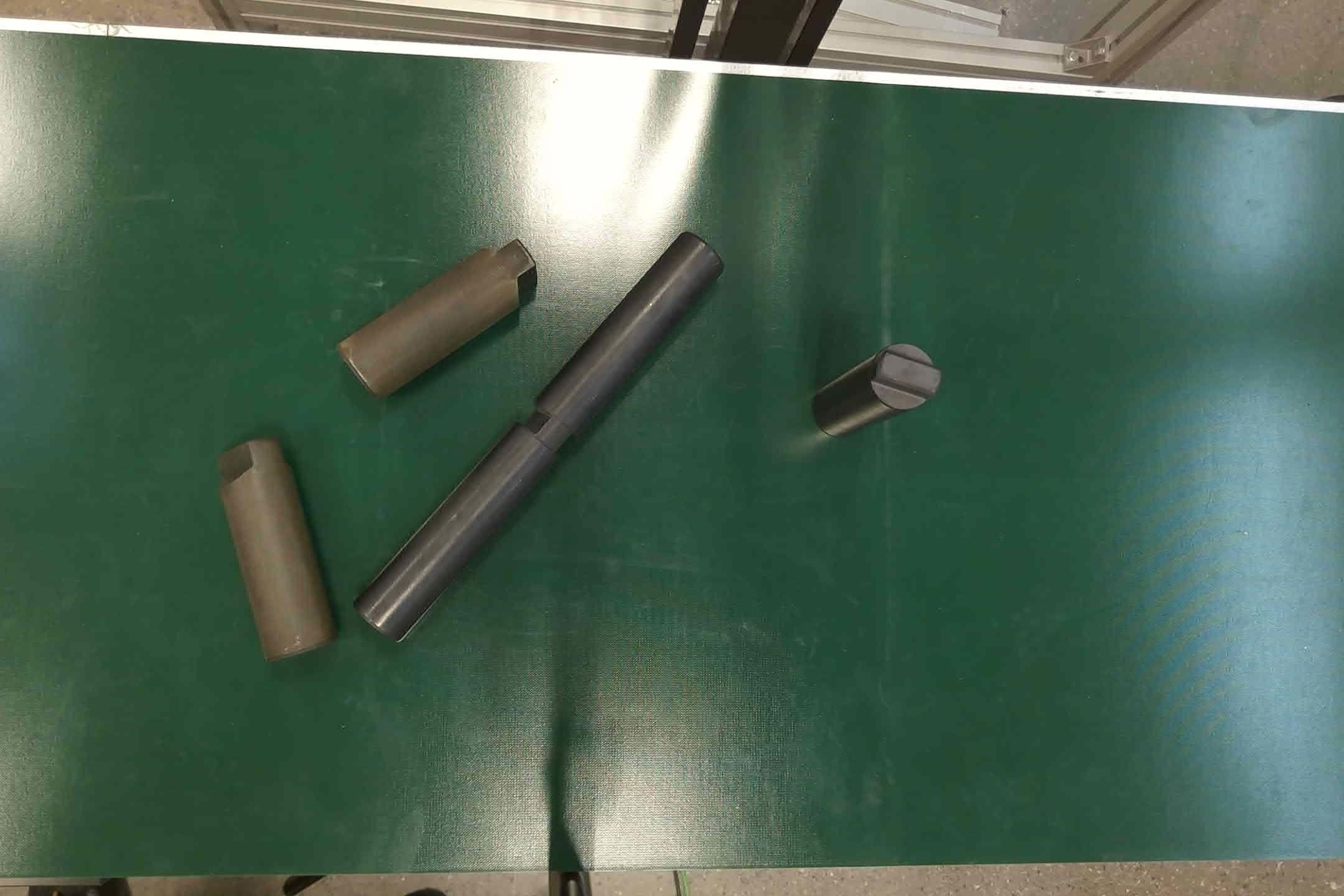}}\hspace{-0.5em}%
  \subfigure[]{\includegraphics[width=0.245\textwidth]{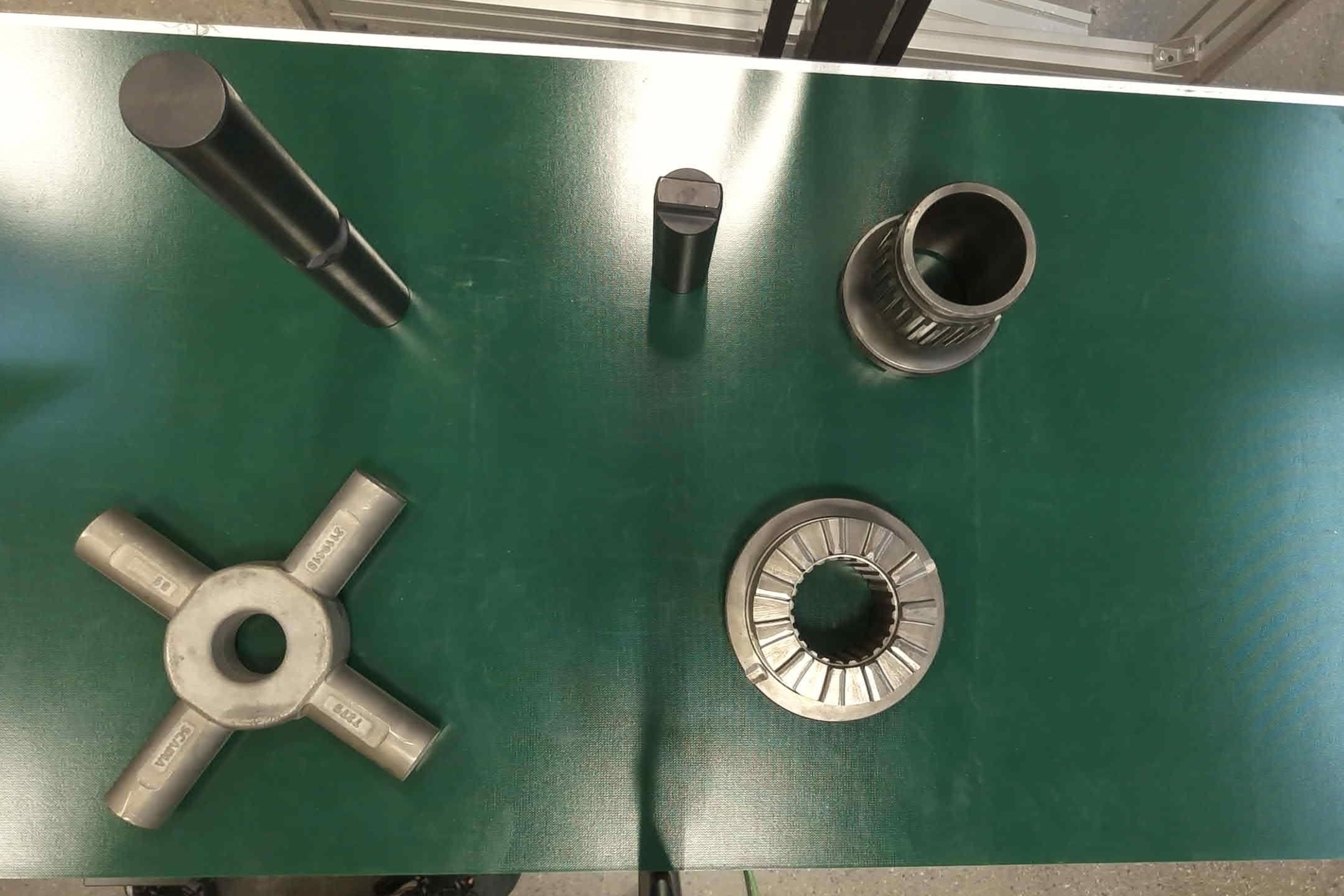}}\hspace{-0.5em}%
  \subfigure[]{\includegraphics[width=0.245\textwidth]{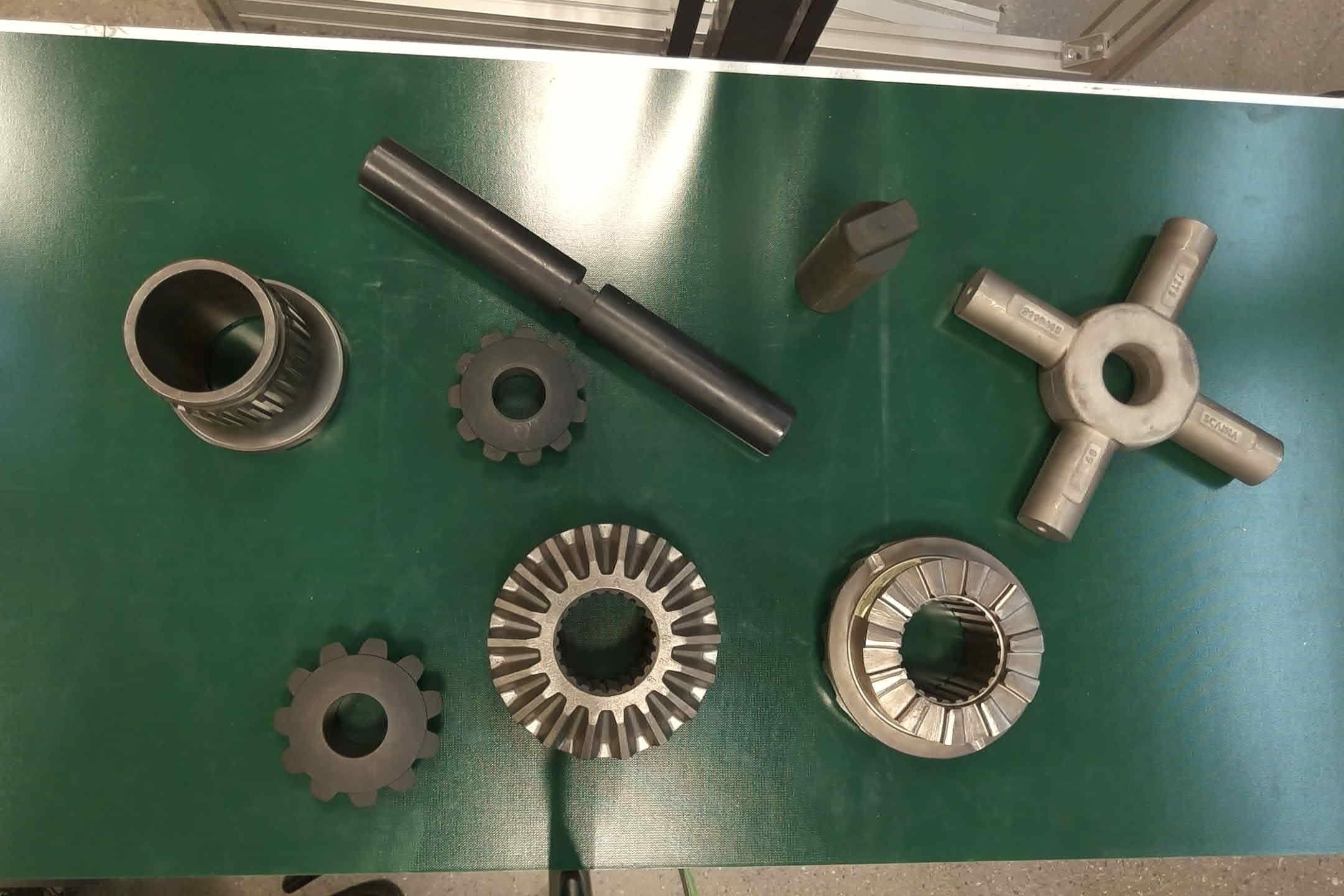}}\hspace{-0.5em}%
  \subfigure[]{\includegraphics[width=0.245\textwidth]{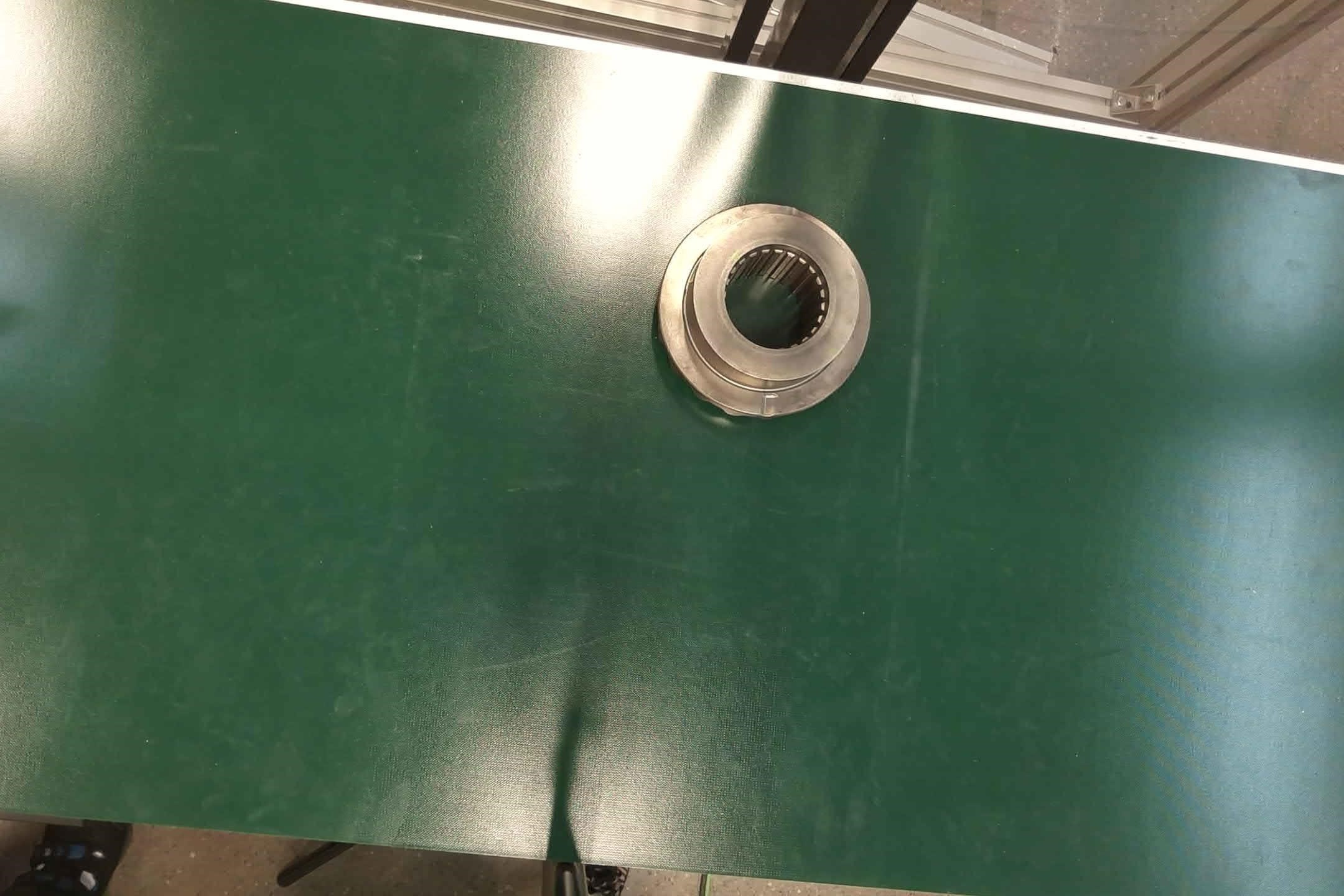}}\\[-0.2em]
  \subfigure[]{\includegraphics[width=0.245\textwidth]{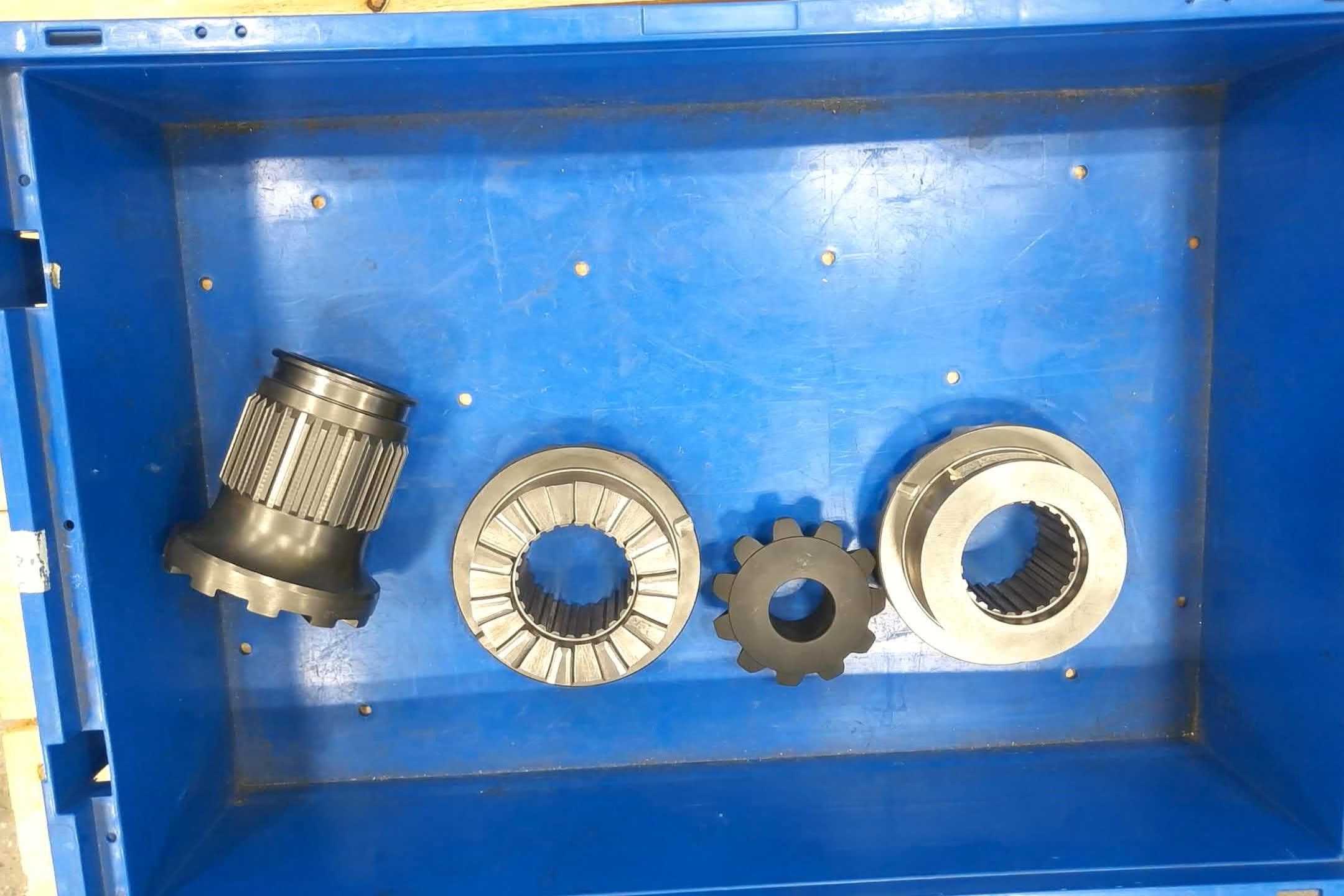}}\hspace{-0.5em}%
  \subfigure[]{\includegraphics[width=0.245\textwidth]{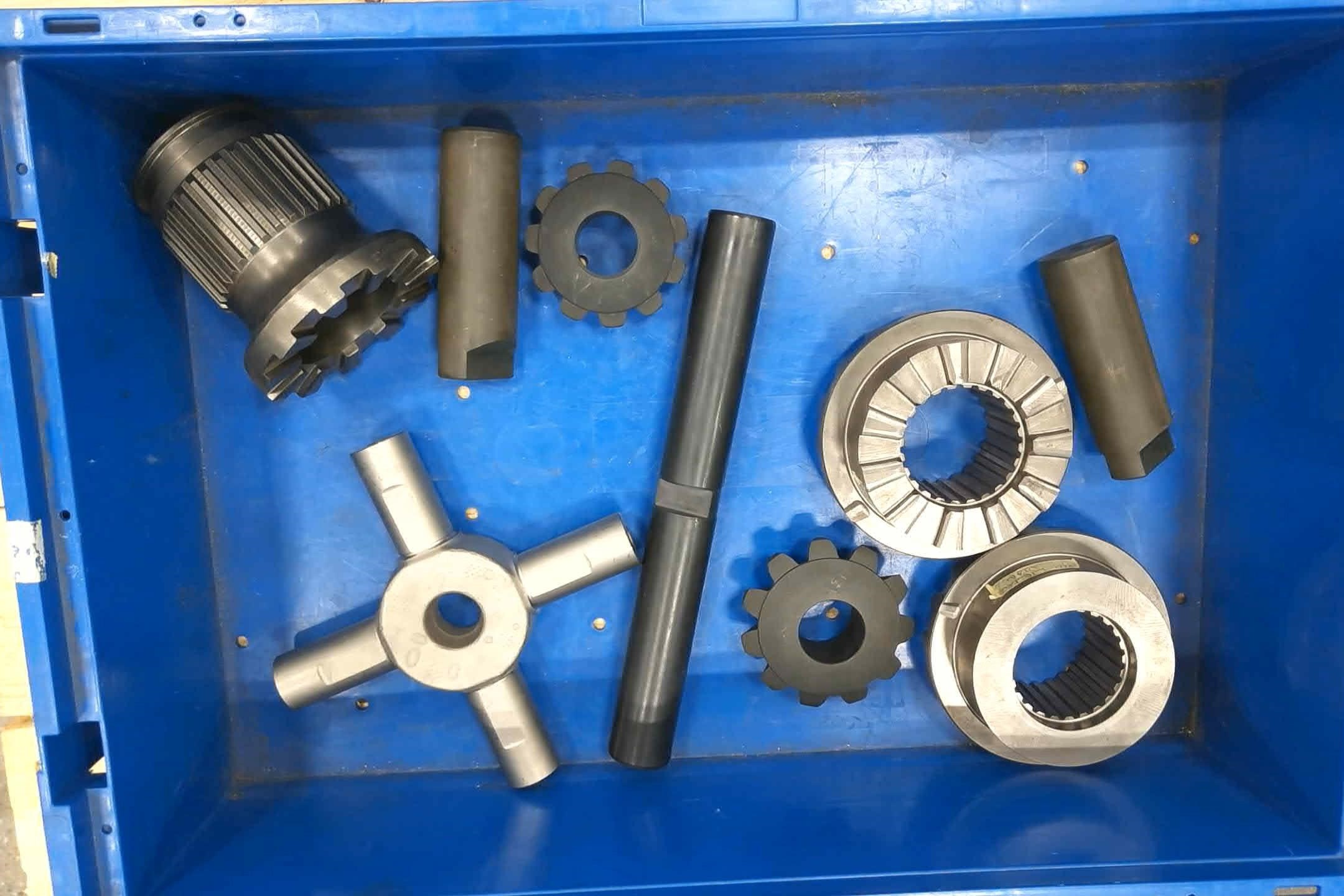}}\hspace{-0.5em}%
  \subfigure[]{\includegraphics[width=0.245\textwidth]{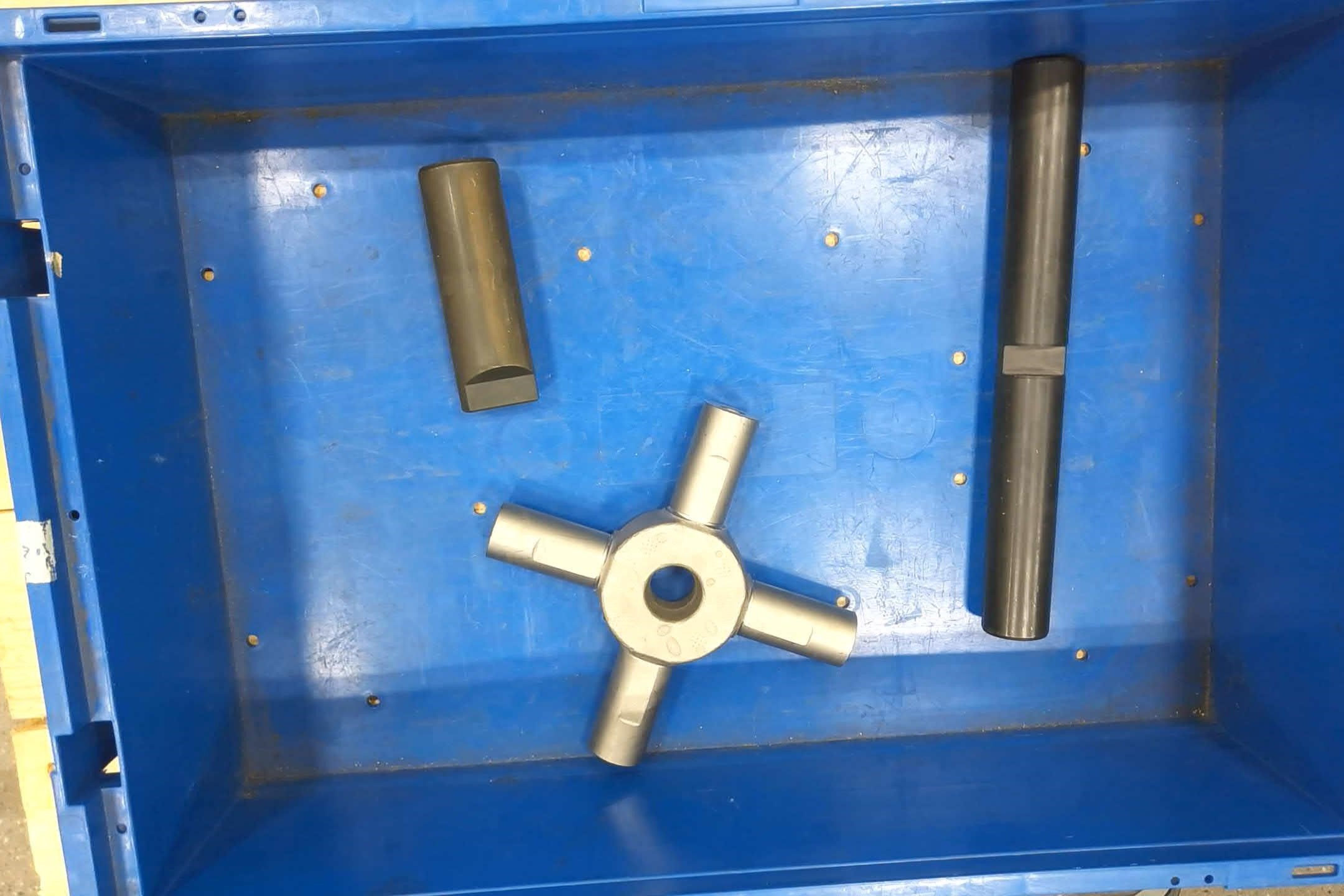}}\hspace{-0.5em}%
  \subfigure[]{\includegraphics[width=0.245\textwidth]{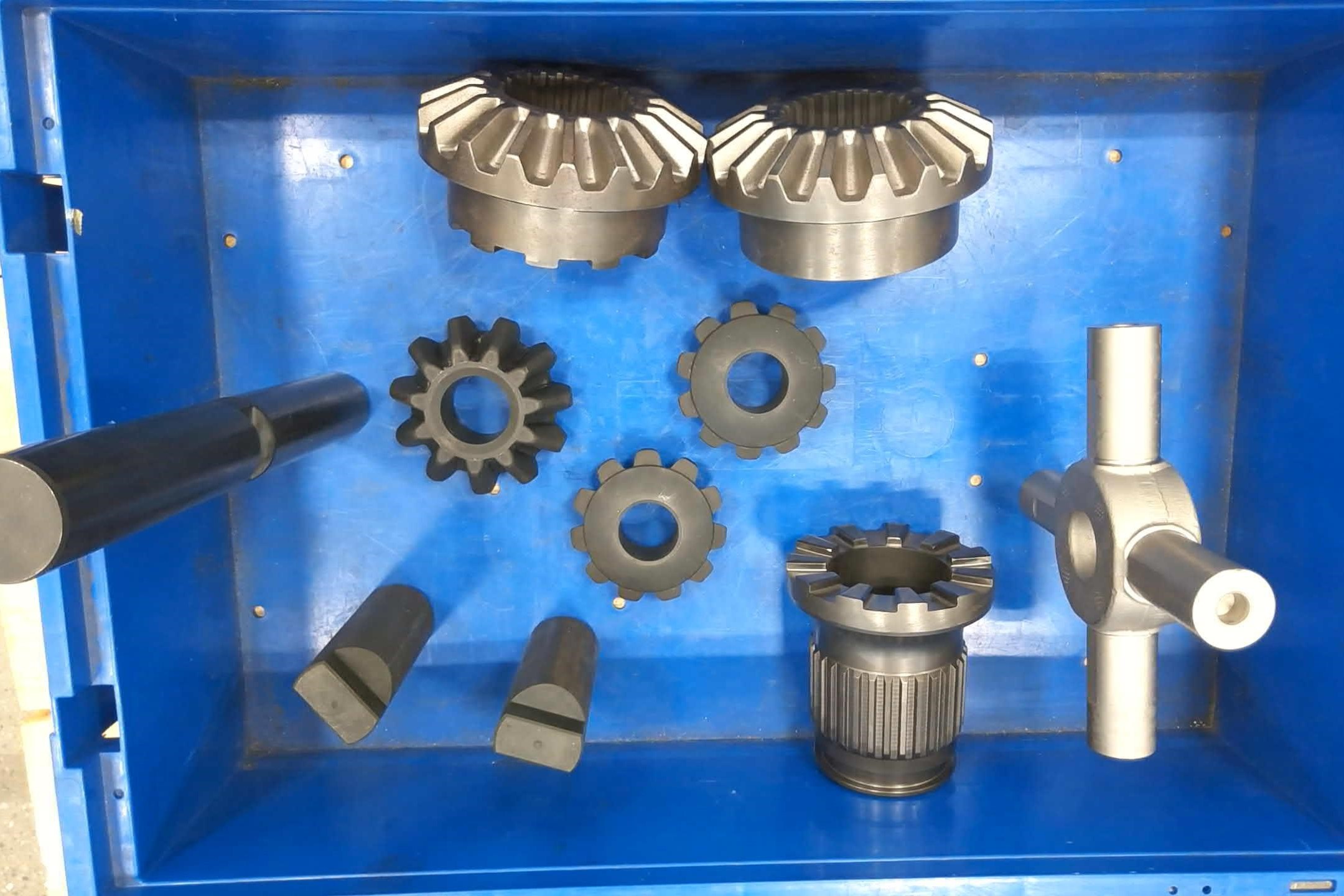}}\\[-0.2em]
  \subfigure[]{\includegraphics[width=0.245\textwidth]{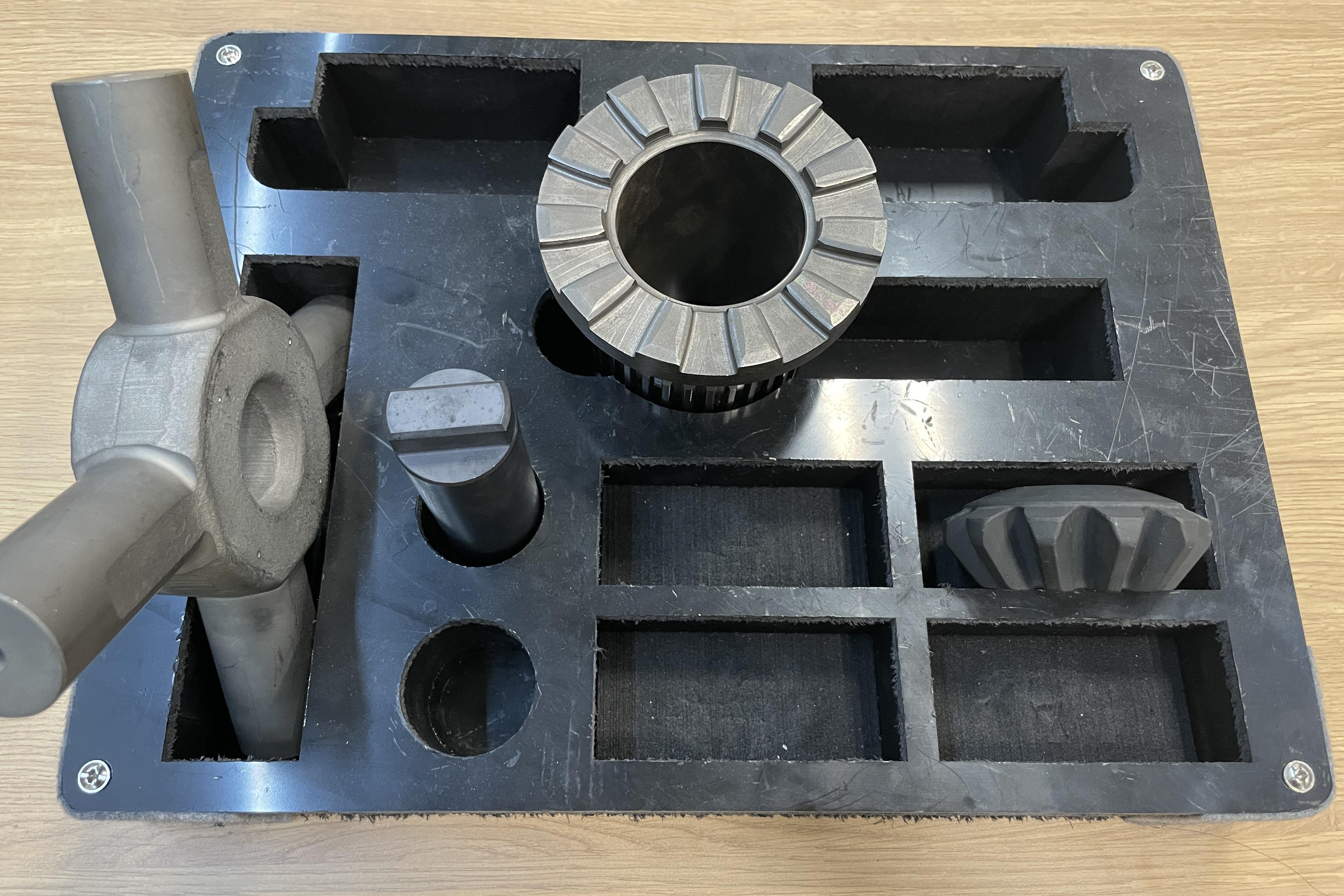}}\hspace{-0.5em}%
  \subfigure[]{\includegraphics[width=0.245\textwidth]{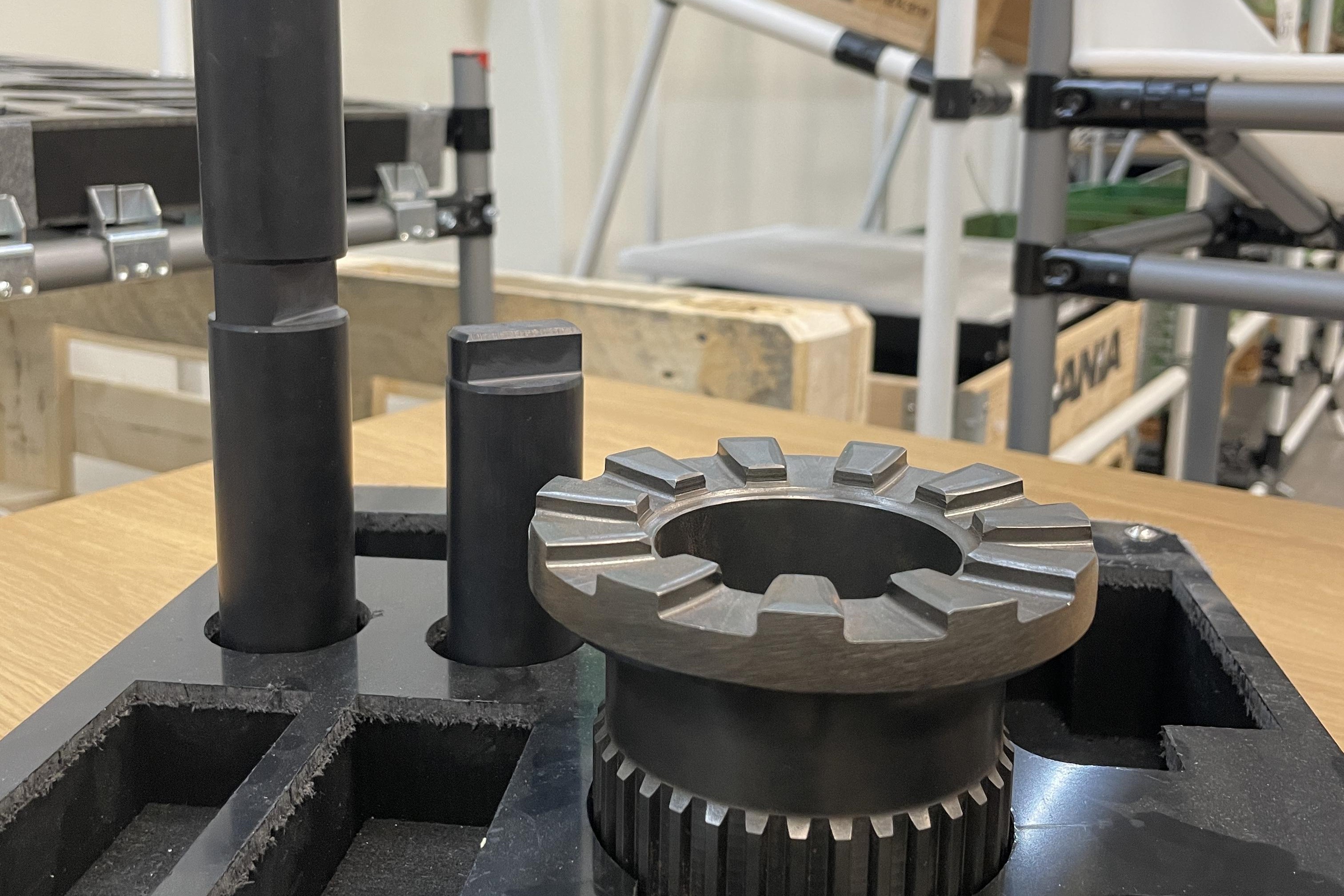}}\hspace{-0.5em}%
  \subfigure[]{\includegraphics[width=0.245\textwidth]{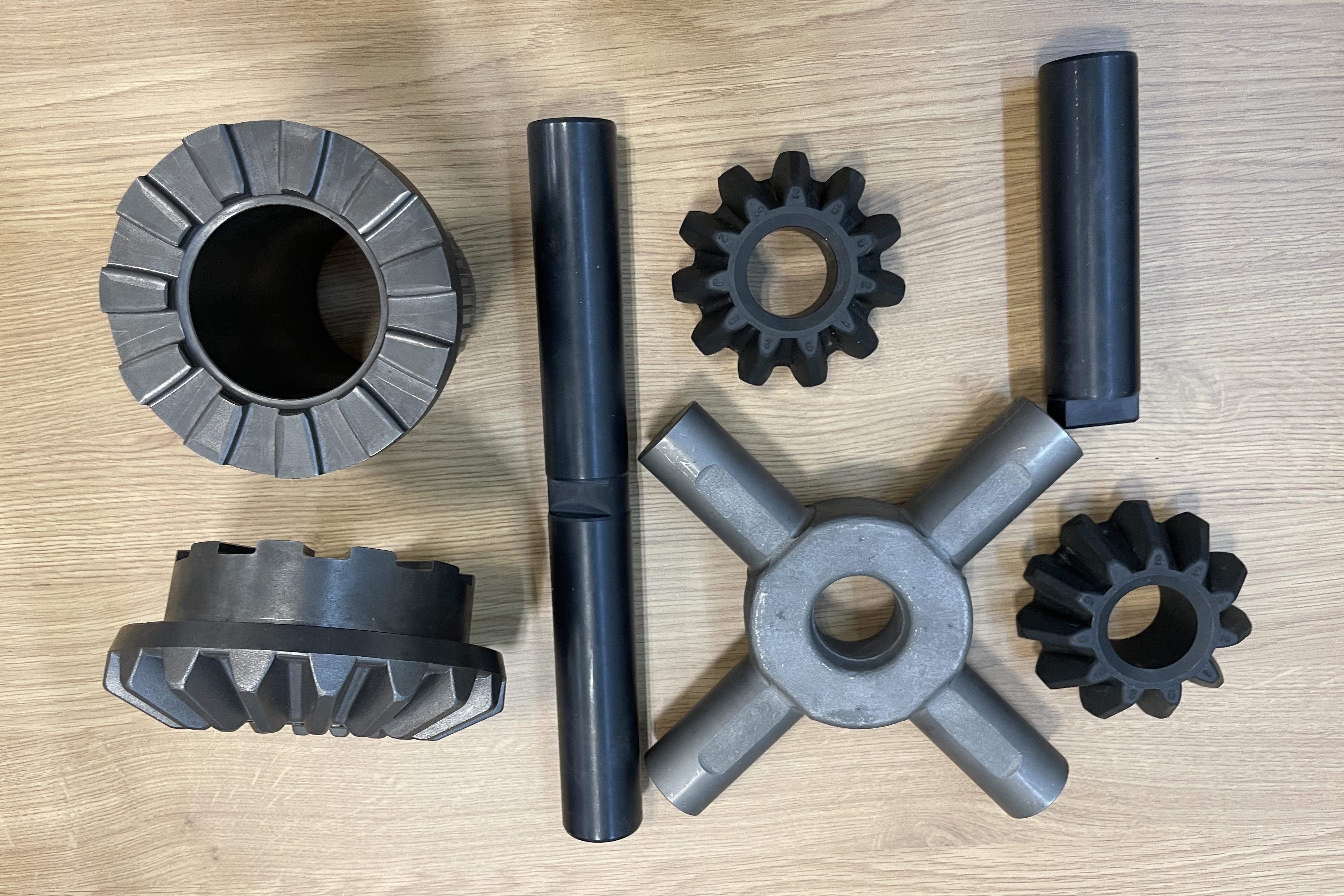}}\hspace{-0.5em}%
  \subfigure[]{\includegraphics[width=0.245\textwidth]{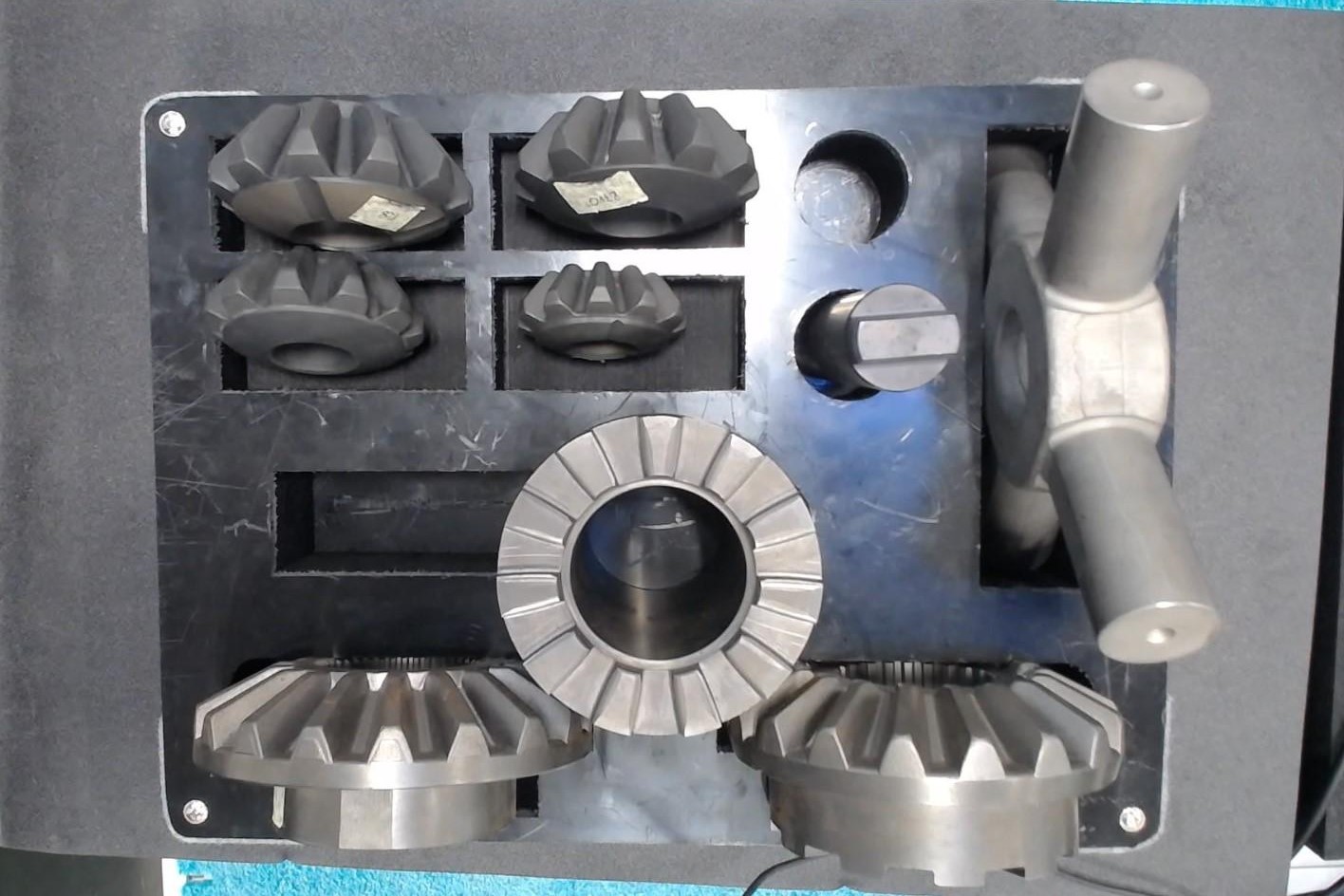}}
  \caption{Sample images from the SIP15-OD dataset, U3. (a-d) Synthetic images. (e-h) Real images from S1. (i-l) Real images from S2. (m-p) Real images from S3. Note: real images are cropped for layout, maintaining original aspect ratios.}
  \label{fig:dataset_us3}
  \vspace{-0.03\textwidth}
\end{figure*}

\paragraph{Visual Results} 
We include a set of visualizations to illustrate our pipeline and model performance. Example synthetic images generated with various DR settings are shown in \cref{fig:DR_pipeline}. Our pipeline was evaluated on two datasets: (1) the public robotic dataset, and (2) the SIP15-OD dataset, which covers three use cases (U1, U2, U3) under three scenarios (S1, S2, S3). Sample synthetic and real images for each use case are provided in \cref{fig:dataset_rb,fig:dataset_us1,fig:dataset_us2,fig:dataset_us3}. All synthetic images were produced using our pipeline. Real images for the robotic dataset were obtained from Horvath et al.\ \cite{horvath2022object}, while SIP15-OD images were collected and annotated by us.

\begin{figure}[t]
  \centering
  \subfigure[Correct prediction.]{%
    \includegraphics[width=0.235\textwidth]{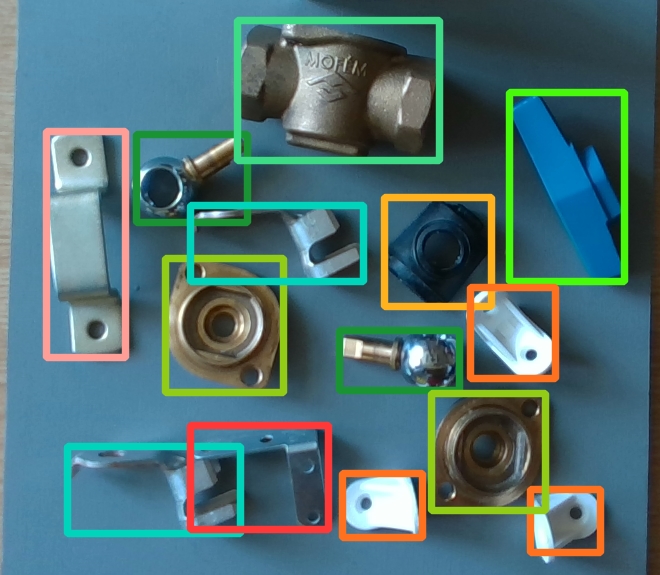}%
  }\hspace{-0.5em}%
  \subfigure[Correct prediction.]{%
    \includegraphics[width=0.235\textwidth]{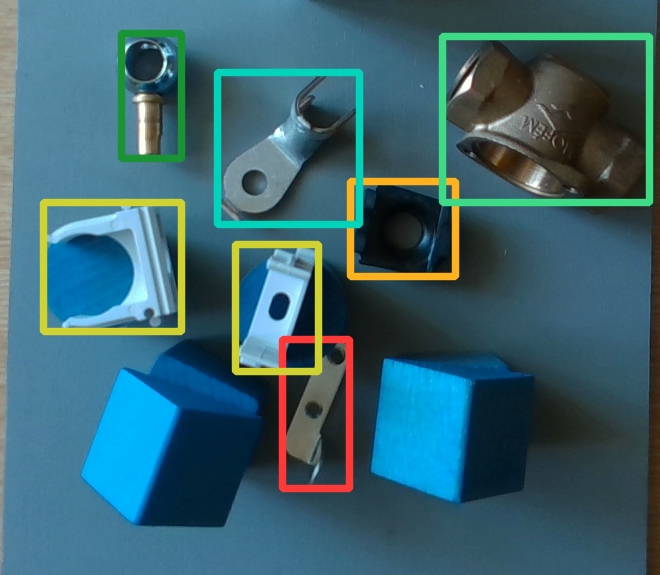}%
  }\hspace{-0.5em}%
  \subfigure[Incorrect prediction.]{%
    \includegraphics[width=0.235\textwidth]{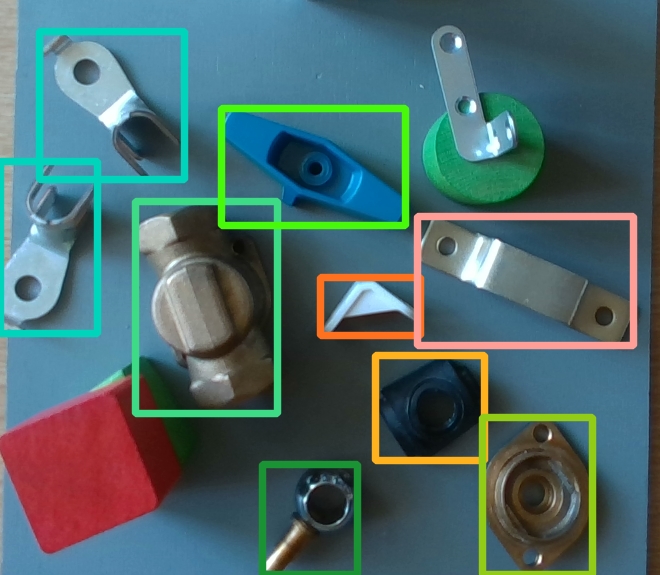}%
  }\hspace{-0.5em}%
  \subfigure[Incorrect prediction.]{%
    \includegraphics[width=0.235\textwidth]{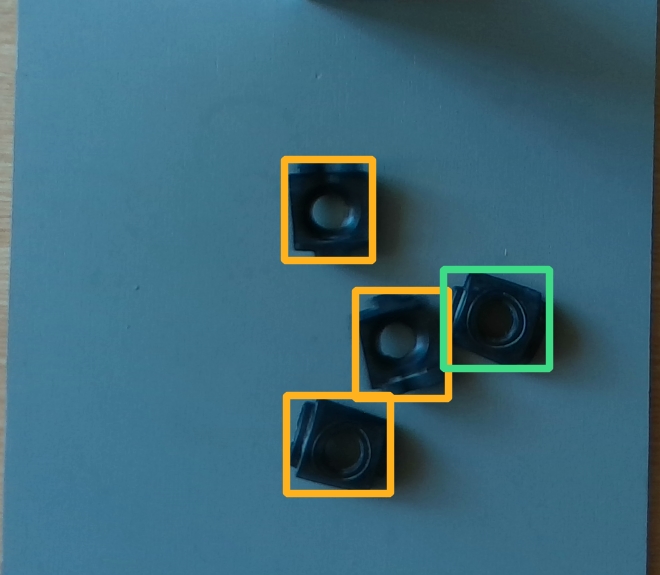}%
  }
  \vspace{-0.5em}
  \caption{Samples of the predictions for the robotic dataset. (a, b) Correct predictions. (c, d) Incorrect predictions. (c) is a missed detection where an \textit{L-bracket} was missed, and (d) is a wrong detection where a \textit{seat} was wrongly detected as a \textit{body}.}
  \label{fig:predict_rb}
  \vspace{-0.02\textwidth}
\end{figure}

\begin{figure}[t]
  \centering
  \subfigure[S1 correct prediction. ]{%
    \includegraphics[width=0.235\textwidth]{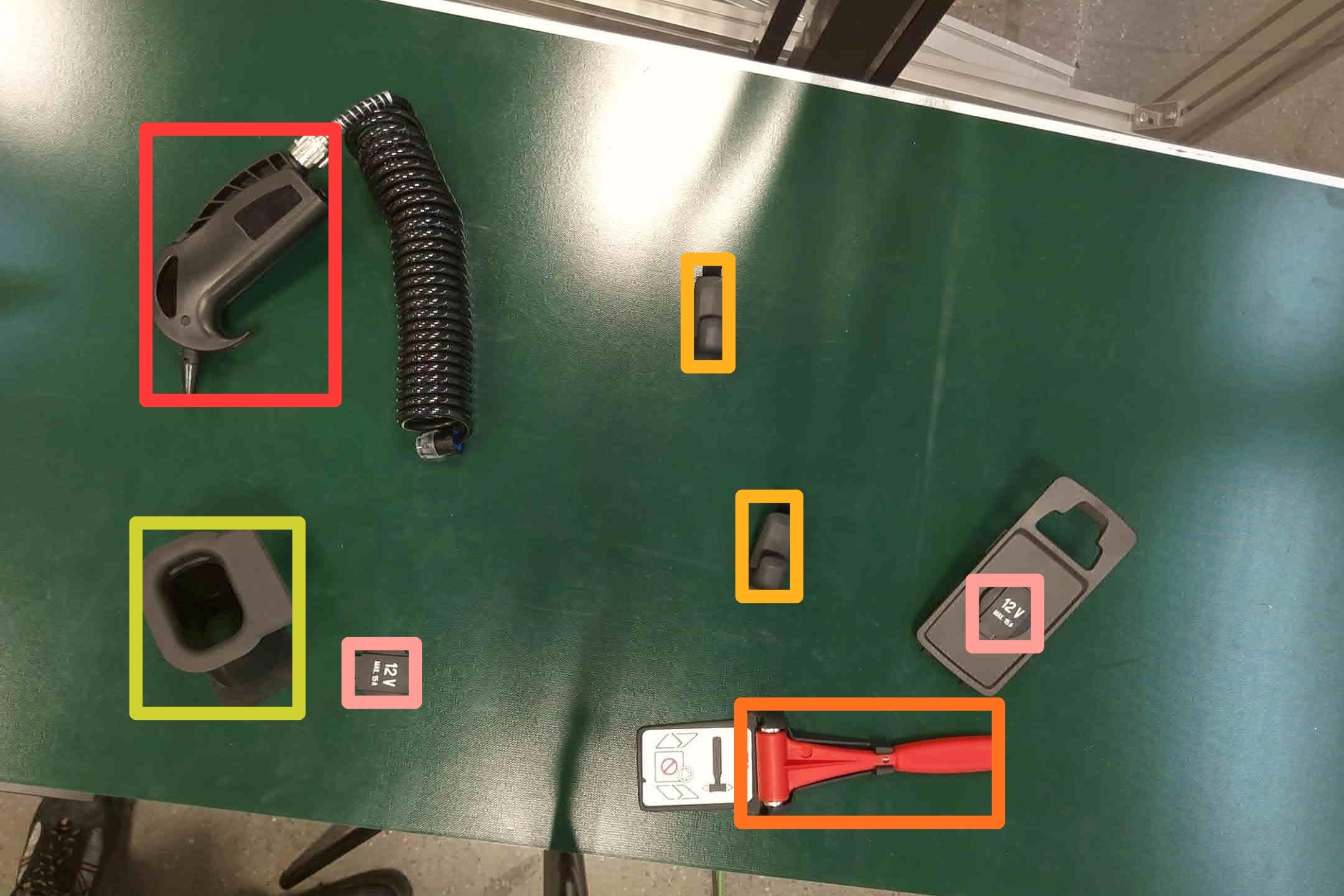}%
  }\hspace{-0.5em}%
  \subfigure[S1 incorrect prediction. ]{%
    \includegraphics[width=0.235\textwidth]{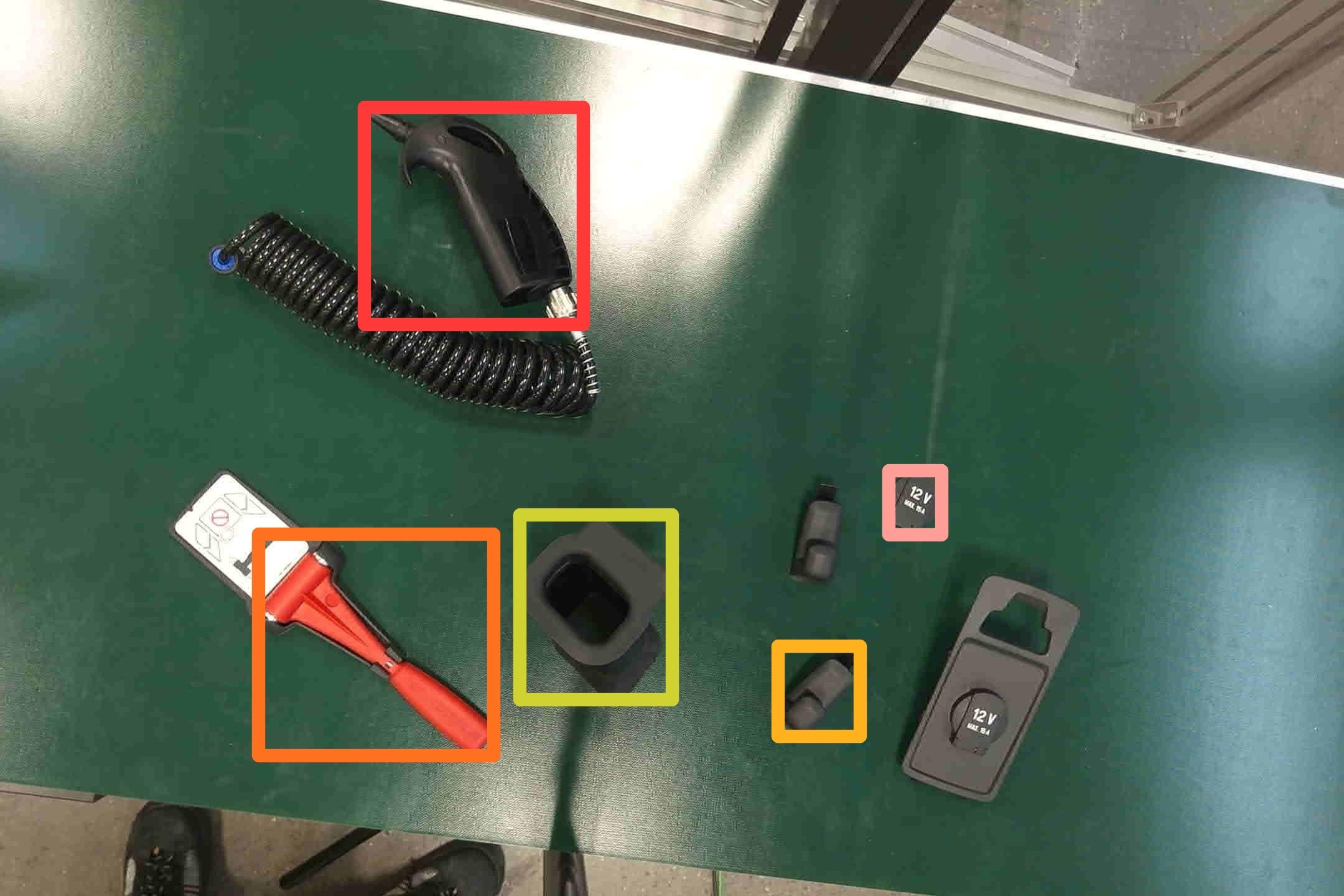}%
  }\hspace{-0.5em}%
  \subfigure[S2 correct prediction.]{%
    \includegraphics[width=0.235\textwidth]{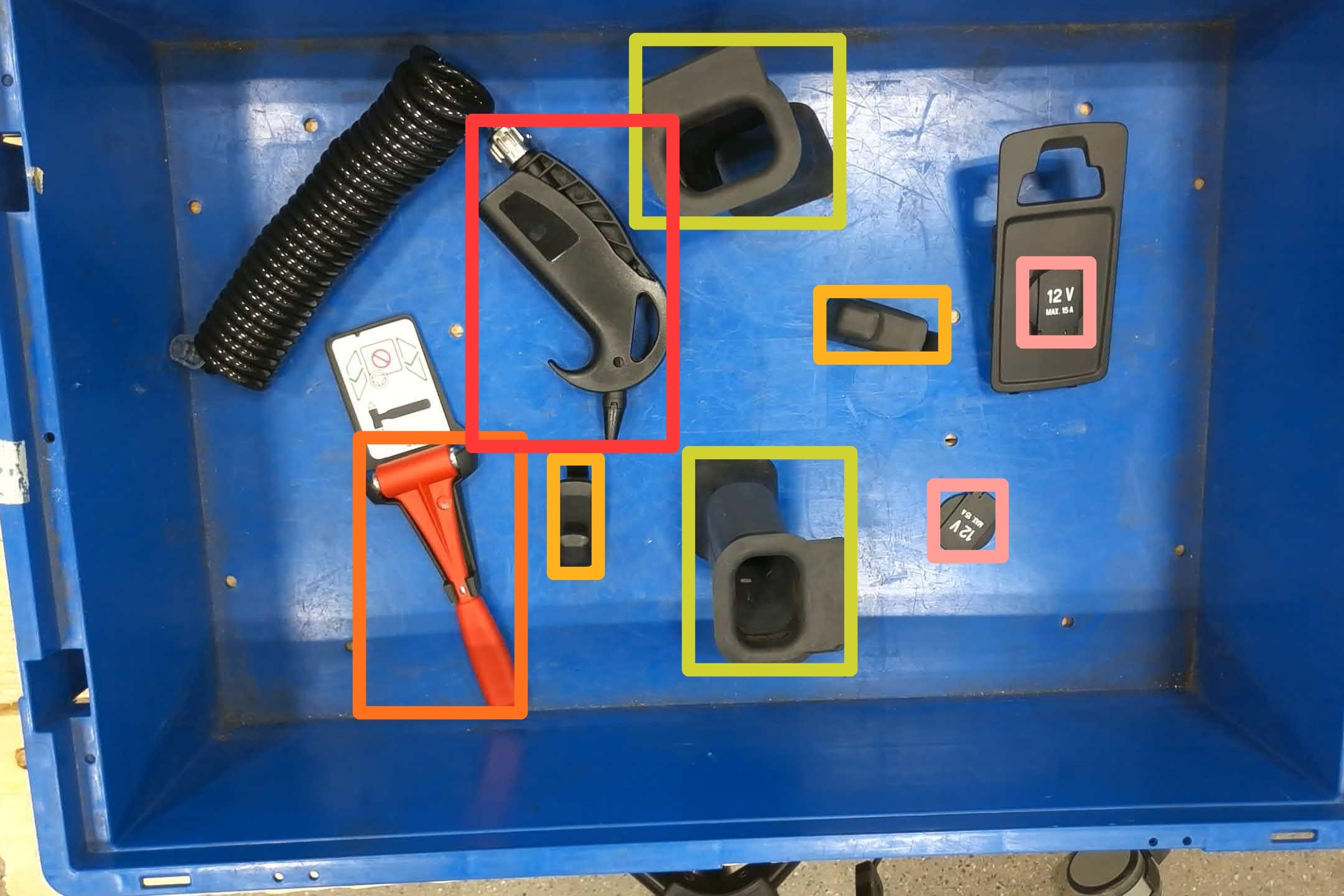}%
  }\hspace{-0.5em}%
  \subfigure[S2 incorrect prediction.]{%
    \includegraphics[width=0.235\textwidth]{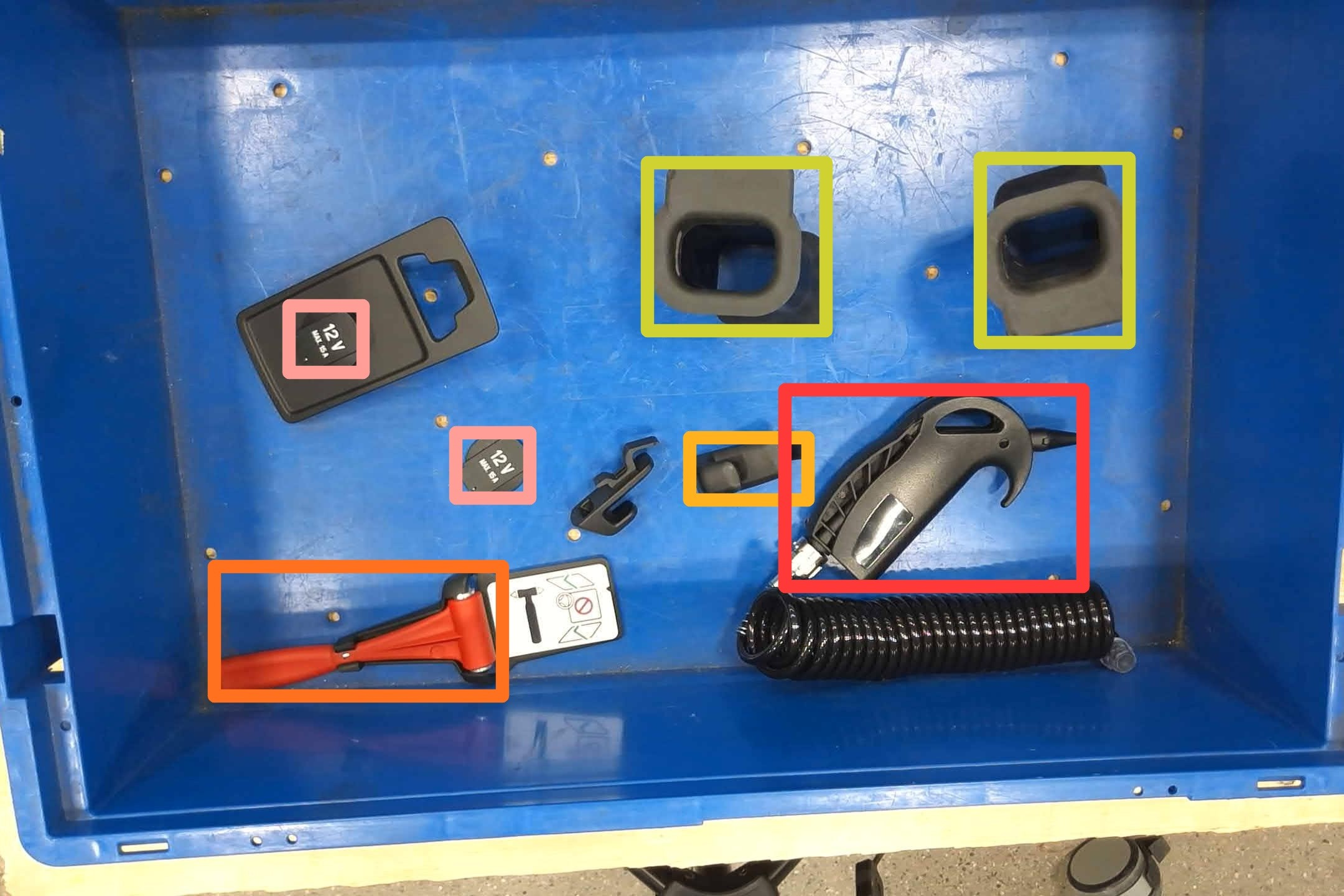}%
  }\\[-0.3em]
  \vspace{-0.5em}
  \caption{Samples of the predictions for the SIP15-OD dataset, U1. (a,c) are correct predictions, (b) is a missed detection where a \textit{hook} and an \textit{eletricity12v} were missed, and (d) is a missed detection where a \textit{hook} was missed. Note: images are cropped for layout, maintaining original aspect ratios.}
  \label{fig:predict_us1}
  \vspace{-0.02\textwidth}
\end{figure}

\begin{figure}[t]
  \centering
  \subfigure[S1 correct prediction. ]{%
    \includegraphics[width=0.235\textwidth]{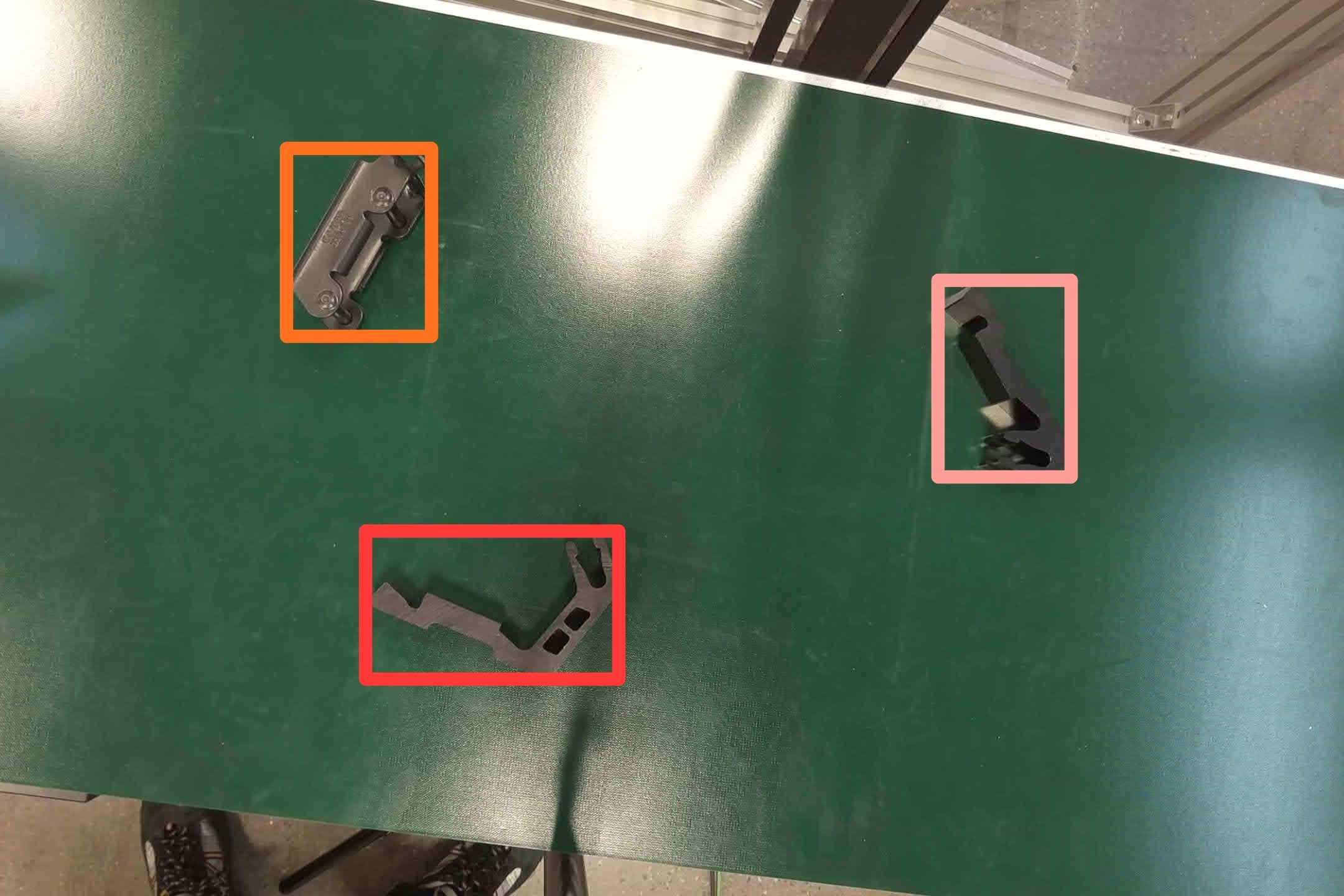}%
  }\hspace{-0.5em}%
  \subfigure[S1 correct prediction. ]{%
    \includegraphics[width=0.235\textwidth]{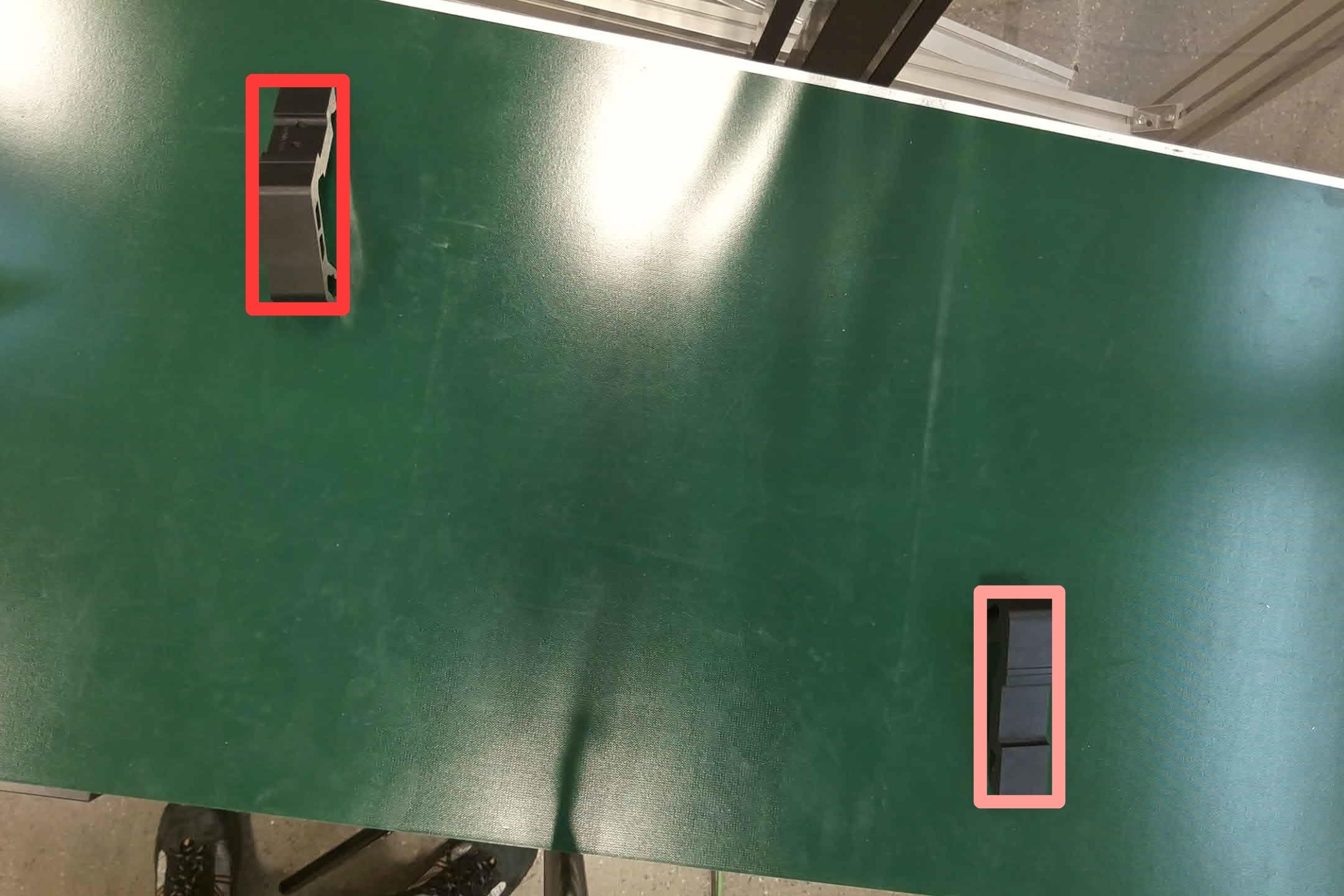}%
  }\hspace{-0.5em}%
  \subfigure[S2 correct prediction. ]{%
    \includegraphics[width=0.235\textwidth]{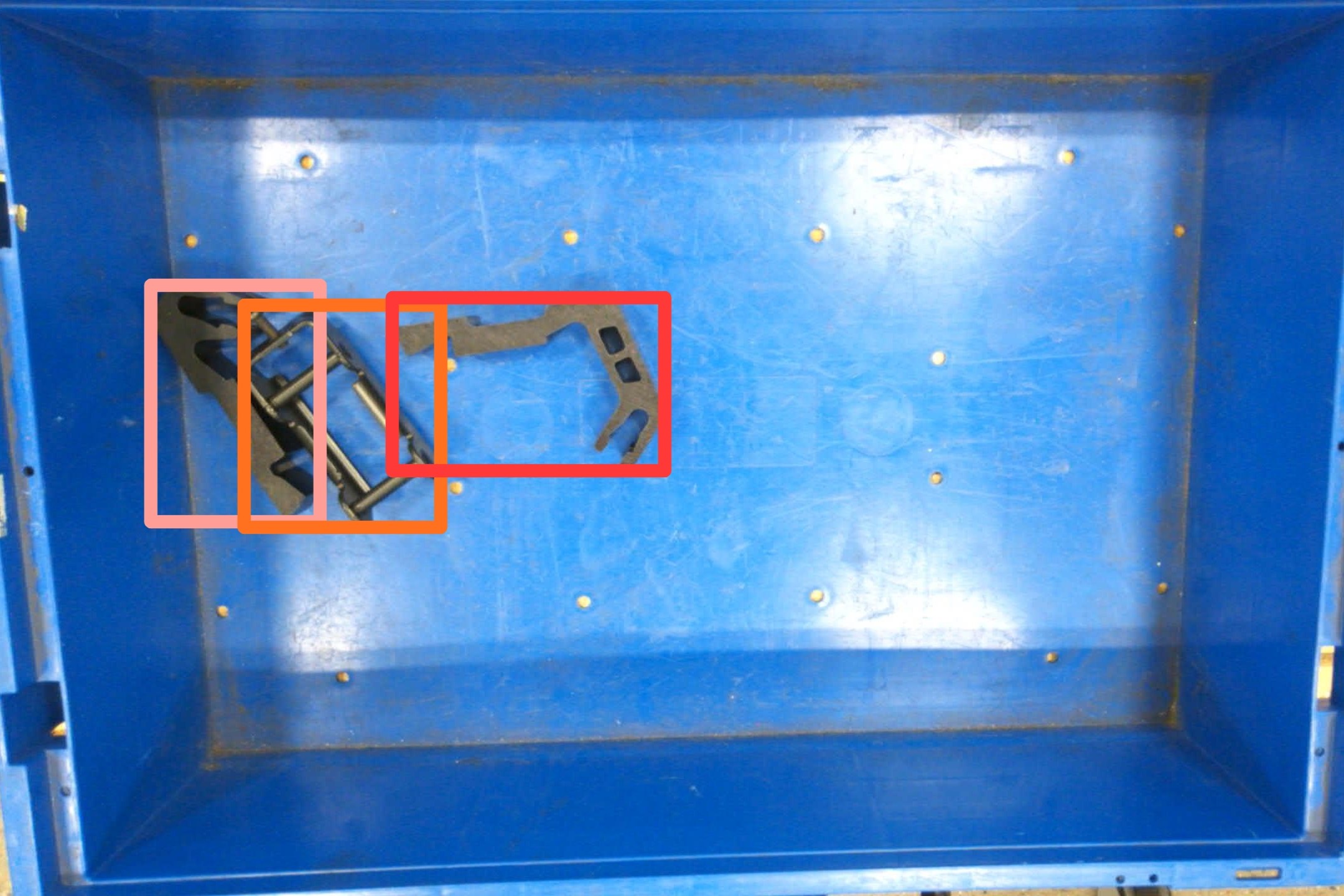}%
  }\hspace{-0.5em}%
  \subfigure[S2 incorrect prediction. ]{%
    \includegraphics[width=0.235\textwidth]{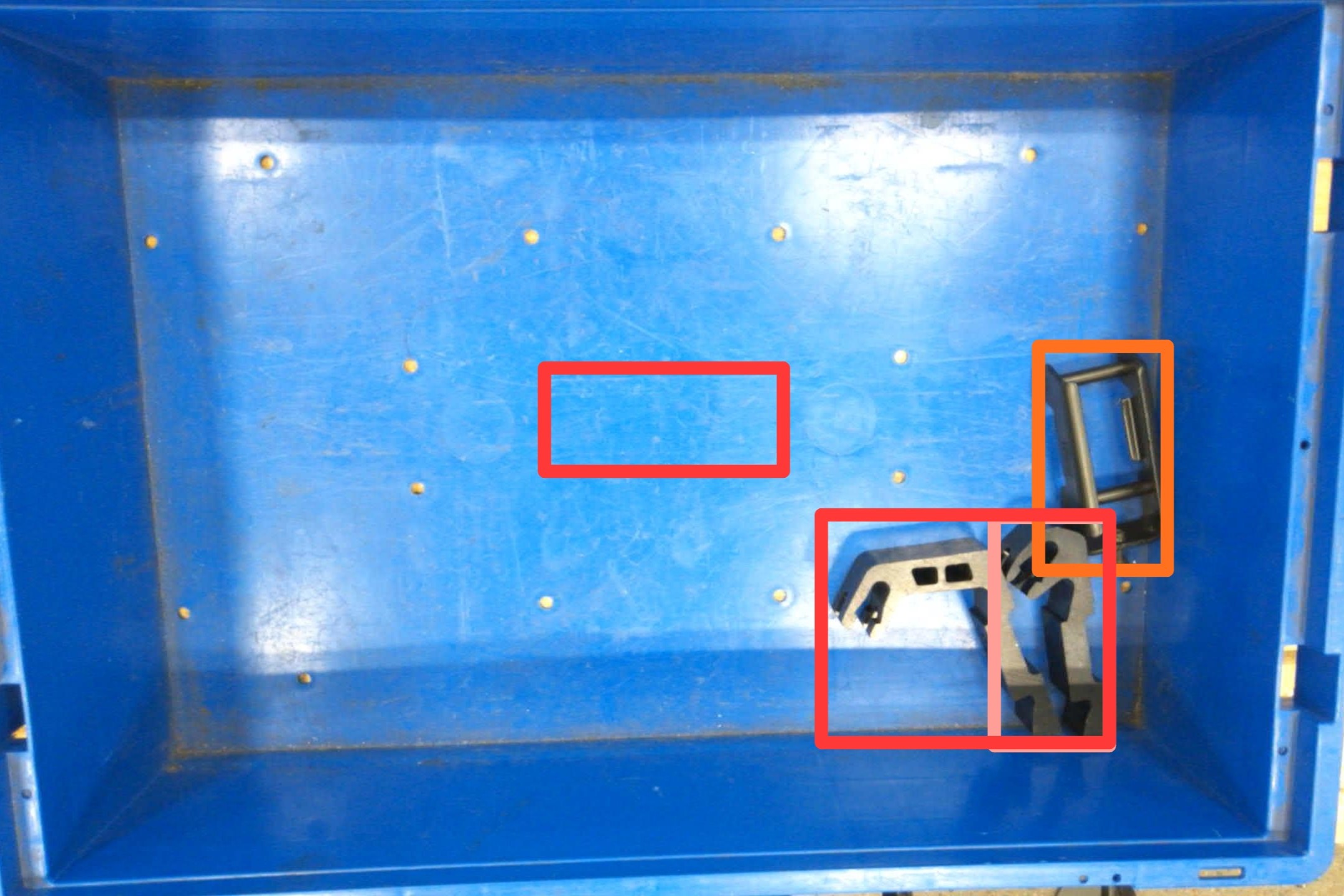}%
  }
  \vspace{-0.5em}
  \caption{Samples of the predictions for the SIP15-OD dataset, U2. (a, b, c) are correct predictions. Due to the good performance, there is only one incorrect prediction in this use case: (d) the background was wrongly detected as a \textit{fork1}. The model may have confused a rectangular mark, which has a similar color to the blue box, for a \textit{fork1}. Note: images are cropped for layout, maintaining original aspect ratios.}
  \label{fig:predict_us2}
  \vspace{-0.02\textwidth}
\end{figure}

\begin{figure}[t]
  \centering
  \subfigure[S1 correct prediction.]{%
    \includegraphics[width=0.235\textwidth]{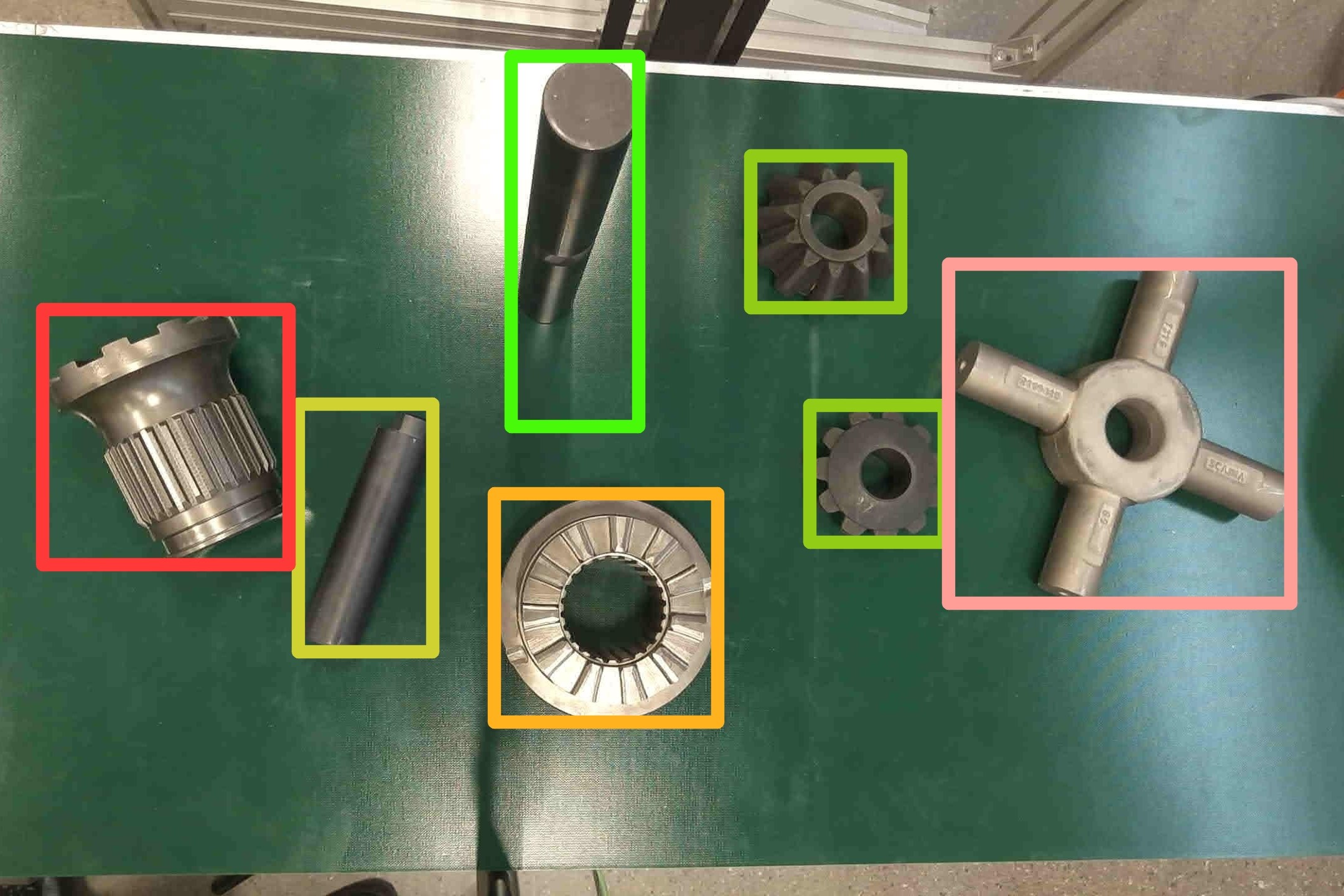}%
  }\hspace{-0.5em}%
  \subfigure[S1 incorrect prediction.]{%
    \includegraphics[width=0.235\textwidth]{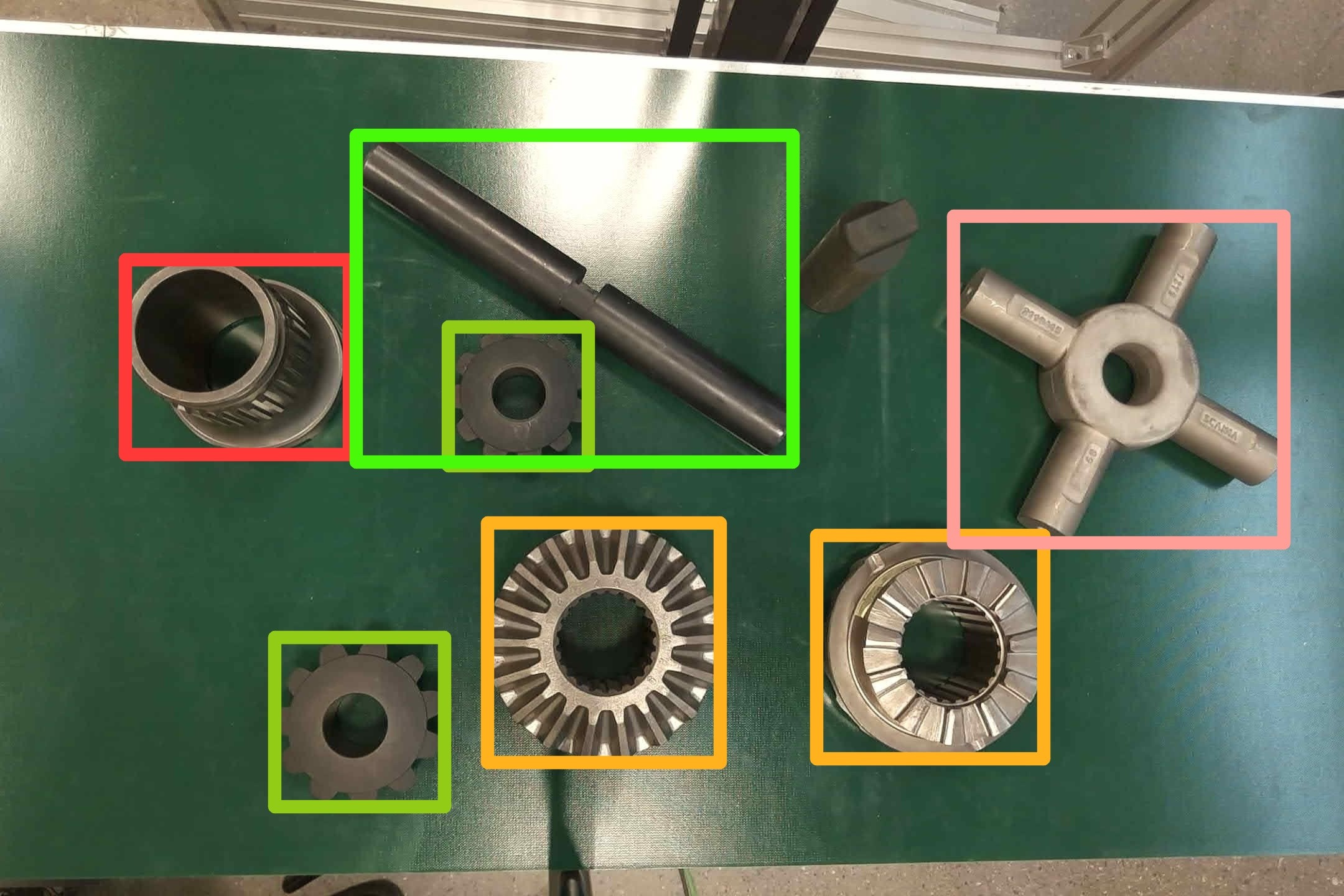}%
  }\hspace{-0.5em}%
  \subfigure[S2 correct prediction.]{%
    \includegraphics[width=0.235\textwidth]{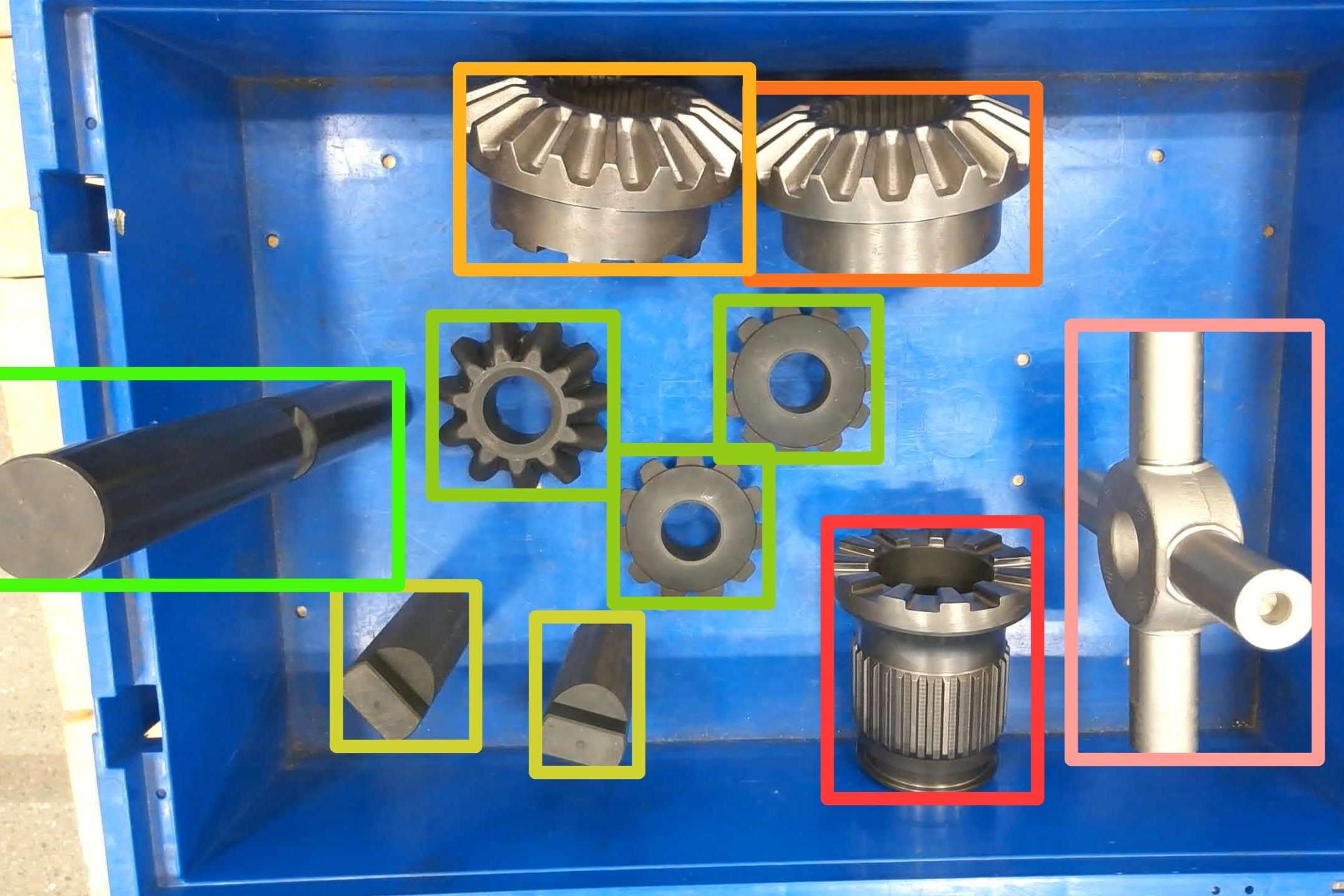}%
  }\hspace{-0.5em}%
  \subfigure[S2 incorrect prediction.]{%
    \includegraphics[width=0.235\textwidth]{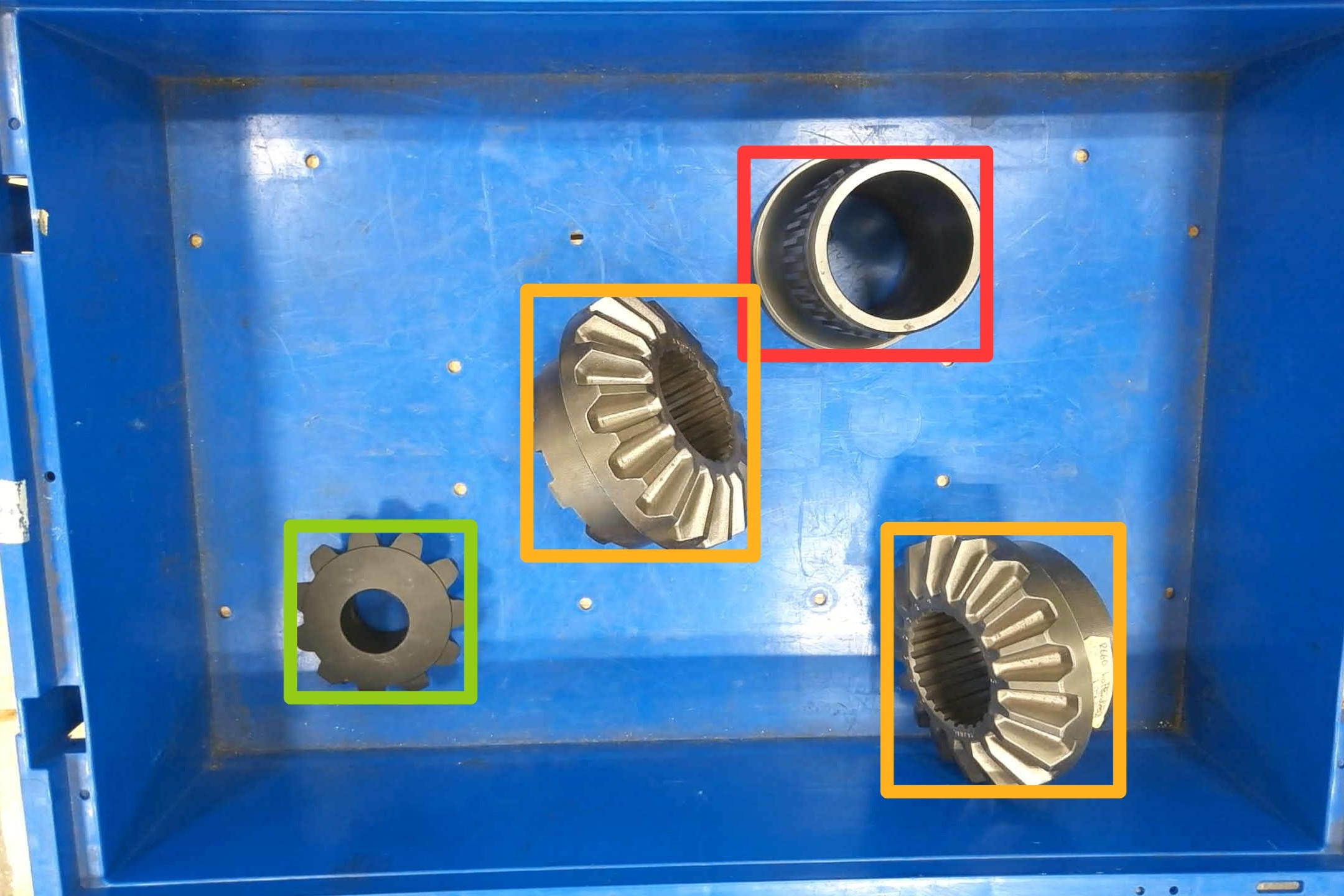}%
  }\\[-0.3em]
  \subfigure[S3 correct prediction.]{%
    \includegraphics[width=0.235\textwidth]{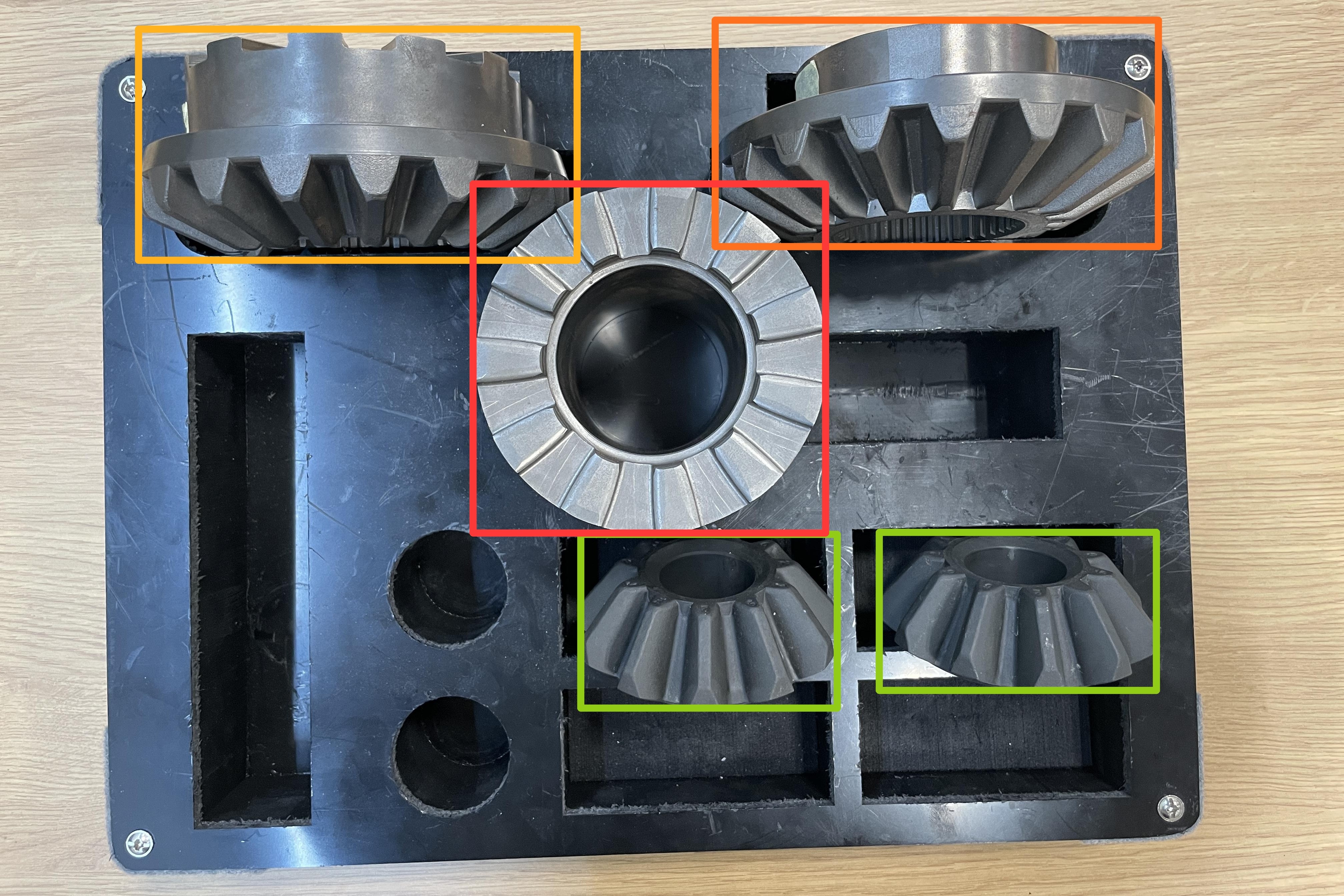}%
  }\hspace{-0.5em}%
  \subfigure[S3 incorrect prediction.]{%
    \includegraphics[width=0.235\textwidth]{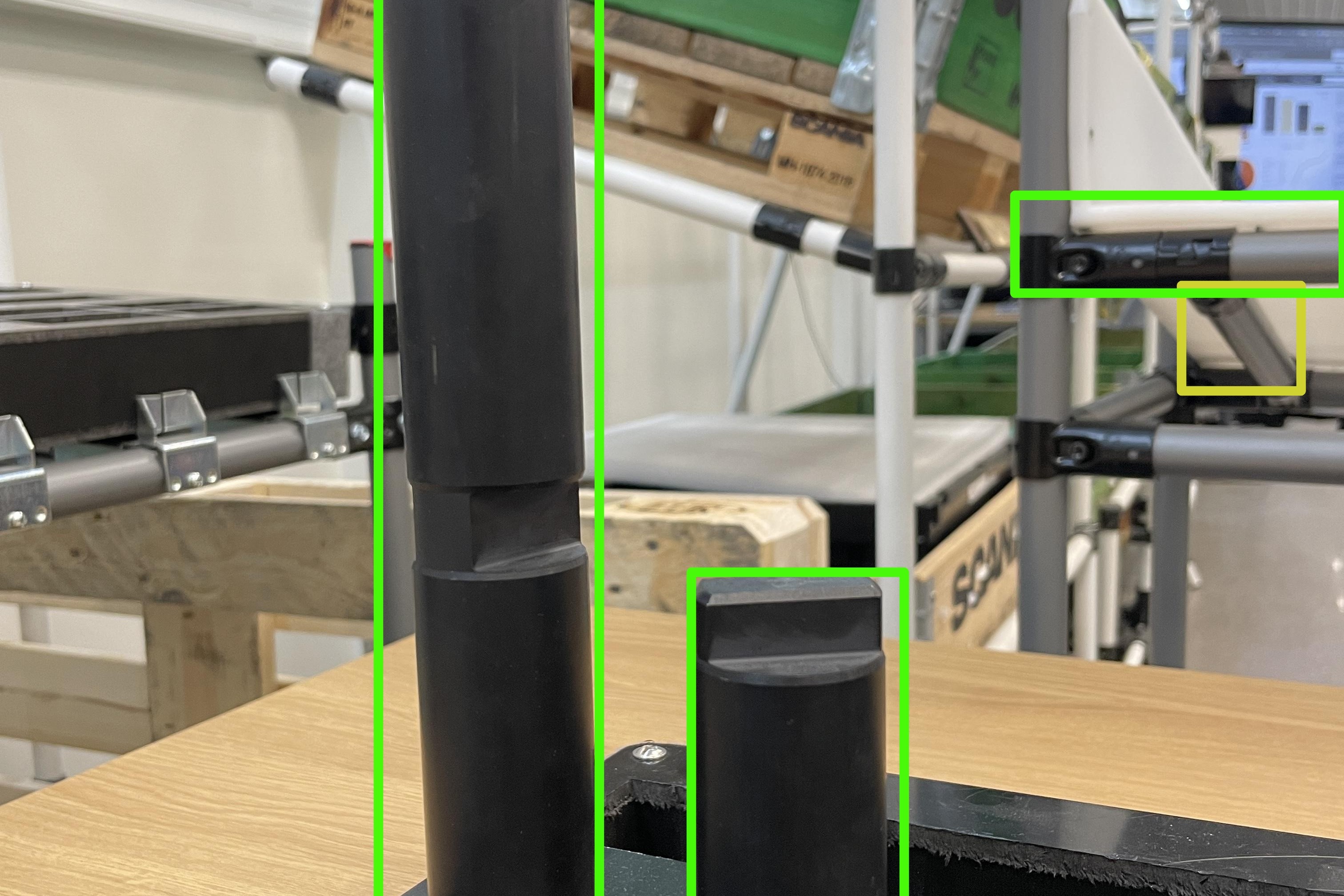}%
  }
  \vspace{-0.5em}
  \caption{Samples of the predictions for the SIP15-OD dataset, U3. (a, c, e) are correct predictions. (b) is a missed detection where a \textit{pin1} was missed; (d) is a wrong detection where a \textit{gear1} was wrongly detected as a \textit{gear2}; and (f) is a wrong detection where a \textit{pin1} was wrongly detected as a \textit{pin2} and two pipes in the background were wrongly detected as \textit{pin1} and \textit{pin2}. Note: images are cropped for layout, maintaining original aspect ratios.}
  \label{fig:predict_us3}
  \vspace{-0.02\textwidth}
\end{figure}

Additionally, representative predictions, both correct and incorrect, are shown in \cref{fig:predict_rb,fig:predict_us1,fig:predict_us2,fig:predict_us3}. These visualizations demonstrate model behavior when trained solely on synthetic data and evaluated on real-world images.

\begin{table*}[t]
  \caption{Comparison of DR techniques in various studies focus on manufacturing applications. Tobin et al. \cite{tobin2017domain} use eight shapes as objects, while others use industrial objects. In our study, we use 10 industrial objects from Horvath et al. \cite{horvath2022object} and 15 objects from our dataset. FOV stands for Field of View; PBR stands for Physically Based Rendering. *Rasterization is not explicitly mentioned in the paper, but appears to be used based on the images and rendering tools.}
  \label{appendix:tab_dr_comparison}
  \centering
  \begin{tabular}{@{}>{\centering\arraybackslash}m{1.2cm} 
                  >{\centering\arraybackslash}m{1.0cm} 
                  >{\centering\arraybackslash}m{2.8cm} 
                  >{\centering\arraybackslash}m{1.2cm} 
                  >{\centering\arraybackslash}m{1.6cm} 
                  >{\centering\arraybackslash}m{1.8cm} 
                  >{\centering\arraybackslash}m{1.8cm} 
                  >{\centering\arraybackslash}m{1.6cm} 
                  >{\centering\arraybackslash}m{1.2cm}@{}}
    \toprule
    \textbf{Work} & \textbf{Categories} & \textbf{Object} & \textbf{Background} & \textbf{Camera} & \textbf{Illumination} & \textbf{Post-processing} & \textbf{Render} & \textbf{Model} \\
    \midrule
    Tobin \cite{tobin2017domain}         & 8   & Position, texture (RGB), gravity                                   & RGB, distractors                              & Pose, FOV                                    & Amount                                  & Noise                                      & *Rasterization      & VGG-16          \\
    Cohen \cite{cohen2020cad}            & 7   & Texture (RGB)                                                      & Few images                                    & Pose                                         & –                                       & Rotation, brightness, blur, shadows       & *Rasterization      & YOLOv3         \\
    Sampaio \cite{sampaio2021novel}      & 2   & Pose, texture (RGB), gravity                                        & Few scenes, distractors                                   & Pose, FOV                                    & Position                                & –                                          & *Rasterization      & Faster-RCNN    \\
    Eversberg \cite{eversberg2021generating} & 1 & Pose, texture (RGB, images, PBR materials)                                   & Images, distractors                           & Pose                                         & Point and HDRI light (amount, position, energy, color)                                    & –                                          & Path tracing        & Faster-RCNN    \\
    Horvath \cite{horvath2022object}     & 10  & Pose, texture (RGB, material images), gravity                            & RGB or images, distractors                              & Pose, FOV                                    & –                                       & Blur, cutout, noise                       & *Rasterization      & YOLOv4         \\
    Tang \cite{tang2024two}              & 10  & Texture (RGB)                                                      & –                                             & Pose                                         & –                                       & Part position, clutter                                   & *Rasterization      & YOLOv5         \\
    \textbf{Ours}                        & 10+15  & Pose, texture (RGB, image, PBR material), gravity                            & Images, distractors                                   & Pose, FOV                                    & Area lights (amount, position, energy, color)        & Noise, blur                               & Path tracing        & YOLOv8         \\
    \bottomrule
  \end{tabular}
  \vspace{-0.03\textwidth}
\end{table*}

\clearpage
\bibliographystyle{IEEEtran}
\bibliography{references}

\end{document}